\documentclass{article}

     \PassOptionsToPackage{numbers, compress}{natbib}

\usepackage[final]{neurips_2023}

\usepackage[utf8]{inputenc} 
\usepackage[T1]{fontenc}    

\usepackage{url}            
\usepackage{booktabs}       
\usepackage{amsfonts}       
\usepackage{nicefrac}       
\usepackage{microtype}      
\usepackage{xcolor}         
\usepackage{graphicx}       
\usepackage{subfig} 
\usepackage{comment} 
\usepackage{longtable} 
\usepackage{tabularx} 
\usepackage{amsmath} 
\usepackage{todonotes}
\usepackage{lipsum}
\usepackage{wrapfig}
\usepackage{placeins} 

\makeatletter
\AtBeginDocument{%
  \renewcommand{\sectionautorefname}{\S\@gobble}

  \renewcommand{\subsectionautorefname}{\S\@gobble}  
}
\makeatother

\newcommand\rurl[1]{%
  \href{https://#1}{\nolinkurl{#1}}%
}
\definecolor{mydarkblue}{rgb}{0,0.08,0.45}
\usepackage[colorlinks,citecolor=mydarkblue,urlcolor=mydarkblue,linkcolor=mydarkblue]{hyperref}
\usepackage{cleveref}
\DeclareMathOperator*{\argmin}{argmin}
\usepackage{minitoc} 

\noptcrule

\title{Scaling Data-Constrained Language Models}

\author{
\vspace{2mm} 
        Niklas Muennighoff~$^1$\hspace{9mm}
        Alexander M. Rush~$^1$ \hspace{9mm}
        Boaz Barak~$^2$ \hspace{9mm}
        Teven Le Scao~$^1$
    \\
    \textbf{
        Aleksandra Piktus~$^1$ \hspace{1.5mm}
        Nouamane Tazi~$^1$ \hspace{1.5mm}
        Sampo Pyysalo~$^3$ \hspace{1.5mm}
        Thomas Wolf~$^1$ \hspace{1.5mm}
        Colin Raffel~$^1$
    }
    \vspace{2mm}
    \\
    \hspace{-3mm}
    \textsuperscript{1} Hugging Face  
    \hspace{4mm}  
    \textsuperscript{2} Harvard University 
    \hspace{4mm}  
    \textsuperscript{3} University of Turku
    \vspace{2mm}
    \\
    \hspace{-8mm}
    {\tt \href{mailto:n.muennighoff@gmail.com}{n.muennighoff@gmail.com}}
}
    \vspace{4mm}

\begin{document}

\doparttoc
\faketableofcontents

\maketitle

\begin{abstract}

  The current trend of scaling language models involves increasing both parameter count and training dataset size. Extrapolating this trend suggests that training dataset size may soon be limited by the amount of text data available on the internet. Motivated by this limit, we investigate scaling language models in data-constrained regimes. Specifically, we run a large set of experiments varying the extent of data repetition and compute budget, ranging up to 900 billion training tokens and 9 billion parameter models. We find that with constrained data for a fixed compute budget, training with up to 4 epochs of repeated data yields negligible changes to loss compared to having unique data. However, with more repetition, the value of adding compute eventually decays to zero. We propose and empirically validate a scaling law for compute optimality that accounts for the decreasing value of repeated tokens and excess parameters. Finally, we experiment with approaches mitigating data scarcity, including augmenting the training dataset with code data or removing commonly used filters. Models and datasets from our 400 training runs are freely available at \url{https://github.com/huggingface/datablations}.

\begin{figure*}[htbp]
    \centering
    \begin{center}
        \includegraphics[width=\textwidth]{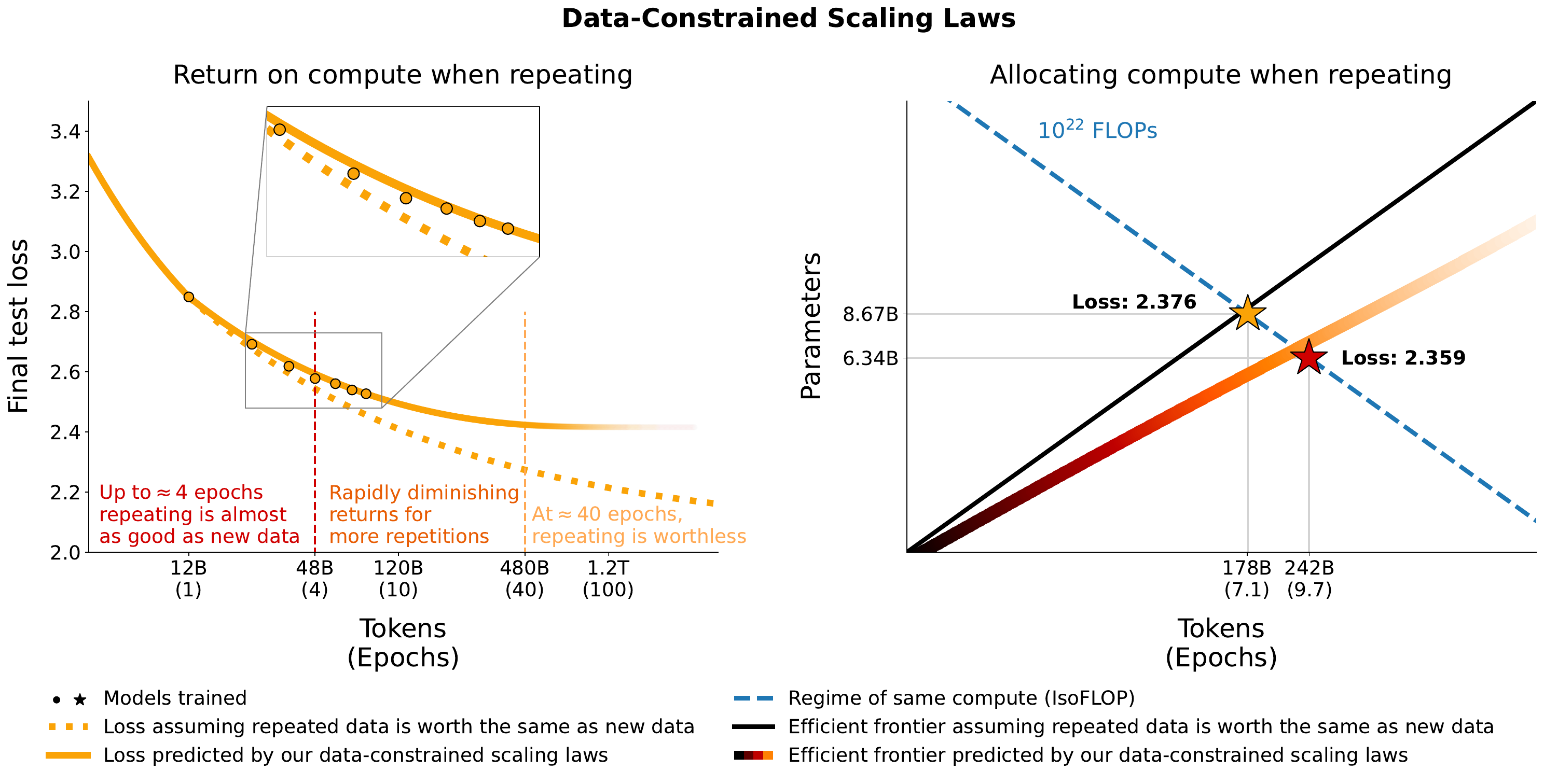}
        \caption{\textbf{\emph{Return} and \emph{Allocation} when repeating data.} \emph{(Left):} Loss of LLMs (4.2B parameters) scaled on repeated data decays predictably (\autoref{sec:fixc}). \emph{(Right):} To maximize performance when repeating, our data-constrained scaling laws and empirical data suggest training smaller models for more epochs in contrast to what assuming Chinchilla scaling laws~\cite{hoffmann2022training} hold for repeated data would predict (\autoref{sec:fixu}).}
        \label{fig:returnalloc}
    \end{center}
\end{figure*}

\end{abstract}

\section{Introduction}

Recent work on compute-optimal language models~\cite{hoffmann2022training} shows that many previously trained large language models (LLMs, which we define as having more than one billion parameters) could have attained better performance for a given compute budget by training a smaller model on more data. Notably, the 70-billion parameter Chinchilla model~\cite{hoffmann2022training} outperforms the 280-billion parameter Gopher model~\cite{rae2021scaling} while using a similar compute budget by being trained on four times more data. Extrapolating these laws for compute allocation (hereafter "Chinchilla scaling laws") to a 530 billion parameter model, such as the under-trained MT-NLG model~\cite{smith2022using}, would require training on a massive 11 trillion tokens, corresponding to more than 30 terabytes of text data. For most languages, available data is several orders of magnitude smaller, meaning that LLMs in those languages are already data-constrained. \citet{villalobos2022will} estimate that even high-quality English language data will be exhausted by the year 2024 given the Chinchilla scaling laws and the trend of training ever-larger models.
This motivates the question~\citep{villalobos2022will,nostalgebraist}: what should we do when we run out of data?

In this work we investigate scaling large language models in a data-constrained regime, and whether training an LLM with multiple epochs of repeated data impacts scaling. Using multiple epochs is, of course, standard in machine learning generally; however, most prior large language models have been trained for a single epoch~\citep{komatsuzaki2019one,brown2020language} and some work explicitly advocates against reusing data~\citep{hernandez2022scaling}. An exception is the recent Galactica models~\cite{taylor2022galactica} that were trained for 4.25 epochs and exhibit continually decreasing validation loss and improving downstream performance throughout training. However, the experiments of Galactica do not compare this setup to an alternative non-data-constrained model trained for one epoch on unique data. Without this comparison, it is difficult to quantify the trade-off between additional compute versus additional data collection. 

Our main focus is to quantify the impact of multiple epochs in LLM training such that practitioners can decide how to allocate compute when scaling models. Toward this end, we assembled a battery of empirical training runs of varying data and compute constraints. Specifically, we train more than 400 models ranging from 10 million to 9 billion parameters for up to 1500 epochs and record final test loss. We use these results to fit a new \emph{data-constrained scaling law} that generalizes the Chinchilla scaling law~\citep{hoffmann2022training} to the repeated data regime and yields a better prediction of loss in this setting. \autoref{fig:returnalloc} summarizes our main results targeting the value of repeated data (\textit{Return}) and optimal allocation of resources in that regime (\textit{Allocation}). We find that, while models trained for a single epoch consistently have the best validation loss per compute, differences tend to be \emph{insignificant} among models trained for up to 4 epochs and do not lead to differences in downstream task performance. Additional epochs continue to be beneficial, but returns eventually diminish to zero. We find that, in the data-constrained regime, allocating new compute to both more parameters and epochs is necessary, and that epochs should be scaled slightly faster. These findings suggest a simple way to continue scaling total training compute budgets further ahead in the future than the previously anticipated limits.


Finally, given the challenges imposed by data constraints, we consider methods complementary to repeating for improving downstream accuracy without adding new natural language data. Experiments consider incorporating code tokens and relaxing data filtering. For code, English LLMs, such as PaLM~\cite{chowdhery2022palm} or Gopher~\cite{rae2021scaling}, are trained on a small amount of code data alongside natural language data, though no benchmarking was reported to justify that decision. We investigate training LLMs on a mix of language data and Python data at 10 different mixing rates and find that mixing in code is able to provide a 2$\times$ increase in effective tokens even when evaluating only natural language tasks. For filtering, we revisit perplexity and deduplication filtering strategies on both noisy and clean datasets and find that data filtering is primarily effective for noisy datasets.

\section{Background}
\label{sec:background}

Predicting the scaling behavior of large models is critical when deciding on training resources. Specifically, two questions are of interest: \textit{(Allocation)} What is the optimal balance of resources? \textit{(Return)} What is the expected value of additional resources?  For scaling LLMs, the resource is compute (measured in FLOPs), and it can be allocated to training a larger model or training for more steps.\footnote{In this work we use  \cite{kaplan2020scaling}'s approximation for the compute cost: $\text{FLOPs}(N, D) \approx 6 N D$, where N denotes the number of model parameters and D denotes the number of tokens processed.}  The metric used to quantify progress is the model's loss on held-out data, i.e.\ the ability to predict the underlying data as measured in the model's cross-entropy~\cite{alabdulmohsin2022revisiting,hoffmann2022training}. We aim to minimize the loss ($L$) subject to a compute resource constraint ($C$) via optimal allocation to $N$ and $D$ as:

\begin{equation}\label{eq:model}
    \argmin_{N, D} L(N, D) \text{ s.t. } \text{FLOPs}(N, D) = C
\end{equation}

Currently, there are established best practices for scaling LLMs. \textit{Return} follows a power-law: loss scales as a power-law with the amount of compute used for training~\cite{henighan2020scaling,kaplan2020scaling,bahri2021explaining,ghorbani2021scaling,bansal2022data,hernandez2021scaling}. \textit{Allocation} is balanced: resources are divided roughly equally between scaling of parameters and data~\cite{hoffmann2022training}. These scaling laws were established empirically by training LLMs and carefully extrapolating behavior. 

Chinchilla~\cite{hoffmann2022training} uses three methods for making scaling predictions: 
\begin{itemize}
    \item 
(\textit{Fixed Parameters}) Train with a fixed model size but on varying amounts of data. 
\item (\textit{Fixed FLOPs}) Train with fixed computation while parameters and training tokens vary. 
\item (\textit{Parametric Fit}) Derive and fit a formula for the loss.
\end{itemize}

For the parametric fit, the loss ($L$) is a function of parameters ($N$) and training tokens ($D$):

\begin{equation}\label{eq:ccbase}
L(N,D) = \frac{A}{N^\alpha} + \frac{B}{D^\beta} + E
\end{equation}

Where $\{A, \alpha, B, \beta, E\}$ are learned variables fit using the training runs from the first two approaches~\cite{hoffmann2022training}. Using these learned variables, they propose calculating the optimal allocation of compute ($C$) to $N$ and $D$ as follows:

\begin{equation}\label{eq:ccopt}
\begin{aligned}
&N_{opt}(C) = G {\left({C}/{6}\right)}^{a}  \quad
D_{opt}(C) = G^{-1} {\left({C}/{6}\right)}^{b}\\
\quad \text{ where } &\quad G = {\left(\frac{\alpha A}{\beta B} \right)}^{\frac{1}{\alpha + \beta}} \quad
a = \frac{\beta}{\alpha+\beta}\quad b = \frac{\alpha}{\alpha + \beta}
\end{aligned}
\end{equation}

These methods lead to the conclusion that $\alpha\approx \beta$ and hence $N$ and $D$ should be scaled proportionally for compute-optimal training. As loss can be an imperfect proxy for performance on natural language tasks~\cite{xia2022training,shin2022effect,tay2021scale}, they also validate their conclusions on various downstream tasks.

\section{Method: Data-Constrained Scaling Laws}
\label{sec:method}

We are interested in scaling behavior in the data-constrained regime. Specifically, given a limited amount of unique data, what is the best \textit{Allocation} of and \textit{Return} for computational resources. Prior work~\cite{kaplan2020scaling,hoffmann2022training} assumes that the necessary data to support scaling is unlimited. Our aim is therefore to introduce a modified version of \autoref{eq:ccbase} that accounts for data constraints and fit the terms in the modified scaling law to data from a large body of experiments.

The primary method we consider is \textit{repeating} data, i.e.\ allocating FLOPs to multiple epochs on the same data. Given a budget of unique data $D_C$, we split the Chinchilla total data term $D$ into two parts: the number of unique tokens used, $U_D$, and the number of repetitions, $R_D$ (i.e. epochs - 1). Given total training tokens $D$ and data budget $D_C$ these terms are simply computed as $U_D = \min \{D_C,D\}$ and $R_D = (D/U_D)-1$. 
When training for a single epoch like done in prior scaling studies, $R_D=0$. We are thus interested in minimizing \autoref{eq:model} with the additional constraint of a data budget $D_C$:

\begin{equation}\label{eq:modeldc}
    \argmin_{N, D} L(N, D) \text{ s.t. } \text{FLOPs}(N, D) = C, U_D \le D_C
\end{equation}

Symmetrically, for mathematical convenience, we split the parameter term $N$ into two parts: the base number of parameters needed to optimally fit the unique tokens $U_N$, and the number of times to ``repeat'' this initial allocation, $R_N$. We compute $U_N$ by first rearranging \autoref{eq:ccopt} to find the optimal compute budget for the unique tokens used ($U_D$). We input this value into the $N_{opt}$ formula of \autoref{eq:ccopt} to get $U_N= \min \{ N_{opt}, N\}$. $U_N$ thus corresponds to the compute-optimal number of parameters for $U_D$ or less if $N < N_{opt}$. Once we have $U_N$, we compute the repeat value as $R_N = (N/U_N)-1$.


To empirically explore the scaling behavior in a data-limited setting we train LLMs under these constraints. We consider three different experimental protocols in this work:

\begin{itemize}
    \item (\textit{Fixed Unique Data}) In \autoref{sec:fixu} we fix the data constraint $D_C$ and train models varying epochs and parameters. These experiments target \textit{Allocation}, specifically tradeoff of $D$ and $N$. 
    \item (\textit{Fixed FLOPs}) In \autoref{sec:fixc} we fix the computation available and vary $D_C$ (and thus also $U_D$ and $U_N$). These experiments target \textit{Return}, i.e. how well does repeating scale compared to having more unique data. 
    \item  (\textit{Parametric Fit}) We fit a formula introduced in \autoref{sec:parametricfit} on all our training runs and evaluate its predictive capability throughout \autoref{sec:fixu} and \autoref{sec:fixc}.
\end{itemize}

Before discussing experimental results we describe the parametric assumptions.

\subsection{Parametric Fit} \label{sec:parametricfit}

To extrapolate scaling curves, it is necessary to incorporate repetition into the Chinchilla formula (\autoref{eq:ccbase}). We generalize \autoref{eq:ccbase} by replacing $D$ and $N$ with terms corresponding to the \emph{effective data} ($D'$) and \emph{effective model parameters} ($N'$).

\[L(N,D)=\frac{A}{N'^\alpha} + \frac{B}{D'^\beta} + E   \]

Intuitively, $D'$ should be smaller or equal to $D$ where $D$ is the total number of processed tokens since repeated tokens provide less useful information to the model than new ones.
We use an \emph{exponential decay} formulation, where the value of a data token processed loses roughly $(1-1/R^*_D)$ fraction of its value per repetition, where $R^*_D$ is a learned constant. 
After some derivations and approximations (see \autoref{sec:scalinglaws}), this boils down to 
\begin{equation}
D' = U_D + U_D R_D^* (1 - e^{\frac{-R_D}{R_D^*}}) \;\label{eq:repd}
\end{equation}
Note that for $R_D=0$ (no repetitions), $D'=U_D=D$.
For $R_D  \ll R^*_D$, $e^{-R_D/R^*_D}\approx 1- \tfrac{R_D}{R^*_D}$ and so \[
D' \approx U_D + U_DR^*_D(1-1+ R_D/R^*_D) = U_D(1+R_D)=D \]
and hence in this case, repeated data is worth almost the same as fresh data.
(This is also consistent with the predictions of the ``deep bootstrap'' framework~\cite{NakkiranNS21}.)
As $R_D$ grows, the value of repeated tokens tends to zero, and the effective data $D'$ becomes much smaller than $D$.
The formula implies that no matter how many times we repeat the data, we will not get a better loss than could be obtained with a single epoch on  $U_D + U_DR^*_D$ fresh tokens.

Just as processing repeated tokens yields a diminishing return, both intuitively and empirically, models with sizes that vastly outstrip the available data also offer diminishing returns per parameter.
Hence we use a symmetric formula for the number of effective parameters, where again $R^*_N$ is learned,

\begin{equation}\label{eq:repn}
N' = U_N + U_N R_N^* (1 - e^{\frac{-R_N}{R_N^*}}) \;
\end{equation}

The learned constants $R^*_D$, $R^*_N$ roughly correspond to the ``half-life'' of repeated data and excess parameters. For example, at $R_D=R^*_D$, the number of effective tokens $D'$ is 
$U_D+ U_DR_D(1-e^{-1})$ which means that the $U_DR_D$ repeated tokens are worth on average $1-1/e$ fraction of fresh ones.

Using a methodology similar to \cite{hoffmann2022training}, $R_N^*$ and $R_D^*$ can be fit on empirical measurements, which yields data-driven estimates. See \autoref{sec:scalinglaws} for more details on the derivations and the fitting procedure.

\pagebreak
\section{Experimental Setup}
\label{sec:exp}

\begin{wrapfigure}{r}{0.4\textwidth}
    \vspace{-5em}
    \centering
    \begin{center}
        \includegraphics[width=\linewidth]{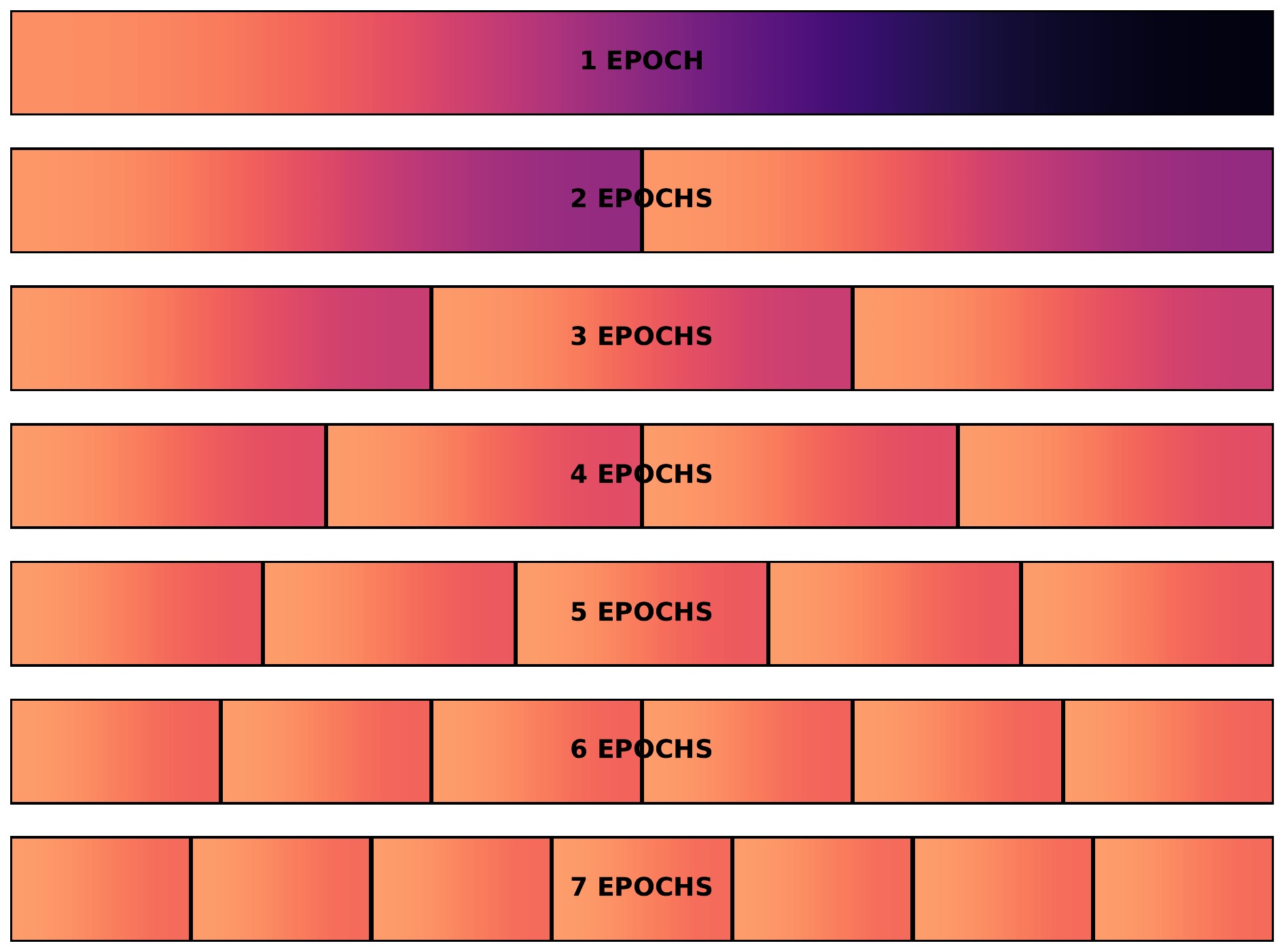}
        \caption{\textbf{Dataset setup.} We ensure that runs using less data (more epochs) always use a subset of the data used in runs with more data (fewer epochs).}
        \label{fig:composition}
    \end{center}
\end{wrapfigure}

For all experiments, we train transformer language models with the GPT-2 architecture and tokenizer~\cite{radford2019language}. Models have up to 8.7 billion parameters and are trained for up to 900 billion total tokens. Following~\cite{hoffmann2022training} we use cosine learning rate schedules that decay 10$\times$ over the course of training for each model (different schedules led to different estimates in \cite{kaplan2020scaling}). Unlike~\cite{kaplan2020scaling}, we do not use early stopping to also explore the extent of overfitting when repeating. Other hyperparameters are based on prior work~\cite{rae2021scaling,hoffmann2022training} and detailed in \autoref{sec:arch}. Models are trained on subsets of C4~\cite{raffel2020exploring}. The data constraints are carefully defined to ensure maximal overlap as shown in \autoref{fig:composition}. Unlike~\cite{hernandez2022scaling}, we always repeat the entire available data rather than subsets of it. Data is shuffled after each epoch. As repeating data can result in extreme overfitting (see \autoref{sec:trainlossc4}), we report loss on a held-out test set unless otherwise specified (see \autoref{sec:eval}). This contrasts training loss used in~\cite{hoffmann2022training}, but should not alter our findings as the held-out data stems from the same underlying dataset.

\section{Results: Resource Allocation for Data-Constrained Scaling}
\label{sec:fixu}

\begin{figure*}
    \centering
    \begin{center}
        \includegraphics[width=\textwidth]{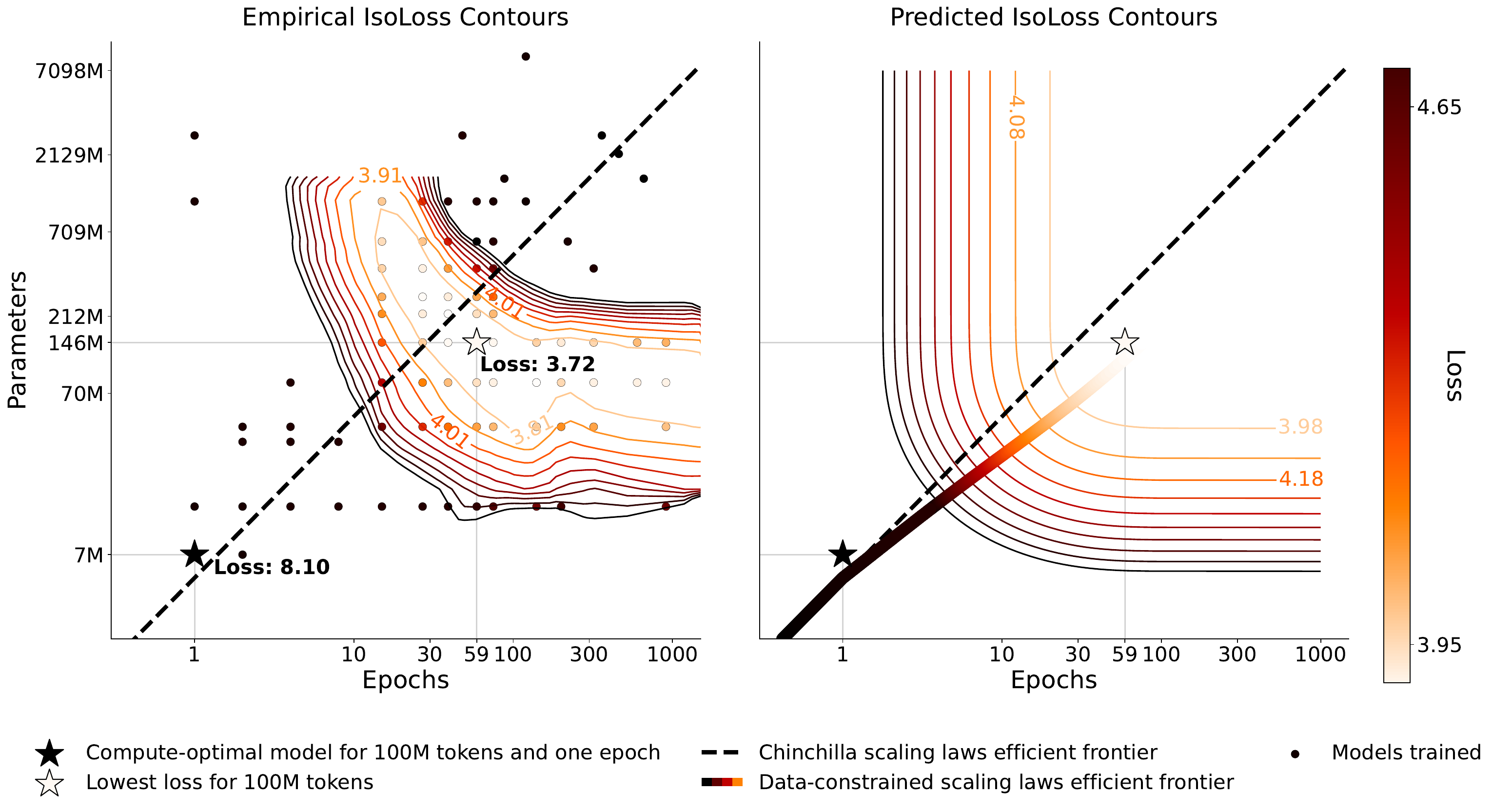}
        \caption{\textbf{IsoLoss contours for 100 million unique tokens.} \textit{(Left):} 93 models trained with varying parameters and epochs on a fixed dataset. Contours show an interpolation of results with the same final test loss. \textit{(Right):} Comparison with the loss predictions from our proposed scaling laws for the same budget of 100 million unique tokens and the predicted efficient frontier. The diminishing returns from training on repeated data can be seen in the increase in distance of the contour curves. 
        }
        \label{fig:100misoloss}
    \end{center}
\end{figure*}

Our first experimental setting considers scaling in a setting where all models have the same data constraint.
For these experiments,  the unique training data budget $D_C$ is fixed at either 100M, 400M or 1.5B tokens. For each data budget, we train a set of language models with increasing amounts of compute that is allocated to either more parameters or more epochs on the unique training data. 

\autoref{fig:100misoloss} (left) shows the main results for scaling with 100M unique tokens\footnote{Although small, for example, this is the order of magnitude of a realistic data constraint reflecting data available after filtering the OSCAR dataset~\citep{OrtizSuarezSagotRomary2019} for Basque, Punjabi, or Slovenian.} (see \autoref{sec:addcon} for 400M and 1.5B tokens). For 100M tokens, the corresponding one-epoch compute-optimal model according to scaling laws from~\cite{hoffmann2022training} has $U_N$ of approximately 7M parameters (see \autoref{sec:c4scaling} for the scaling coefficients we use). 
Results show that more than a 50\% reduction in loss can be attained by training for several epochs ($R_D>0$) and increasing model size beyond what would be compute-optimal for 100M tokens ($R_N>0$). We find the best loss to be at around 20-60$\times$ more parameters and epochs, which corresponds to spending around 7000$\times$ more FLOPs. These results suggest that one-epoch models significantly under-utilize their training data and more signal can be extracted by repeating data and adding parameters at the cost of sub-optimal compute utilization. 

\autoref{fig:100misoloss} (right) shows the predicted contours created by fitting our data-constrained scaling laws on 182 training runs. In the single-epoch case ($R_D=0$) with near compute-optimal parameters ($R_N=0$) our scaling equation (\autoref{sec:parametricfit}) reduces to the Chinchilla equation. In this case, both formulas predict the optimal allocation of compute to parameters and data to be the same, resulting in overlapping efficient frontiers. As data is repeated for more than a single epoch, our fit predicts that excess parameters decay faster in value than repeated data ($R_N^*<R_D^*$). As a result, the data-constrained efficient frontier suggests allocating most additional compute to more epochs rather than more parameters. This contrasts the Chinchilla scaling laws~\cite{hoffmann2022training}, which suggest equally scaling both. However, note that they do not repeat the entire training data and their parametric fit explicitly relies on the assumption that models are trained for a single epoch only. Thus, there is no guarantee that their scaling predictions hold for repeated data.

\begin{figure*}
    \centering
    \begin{center}
        \includegraphics[width=\textwidth]{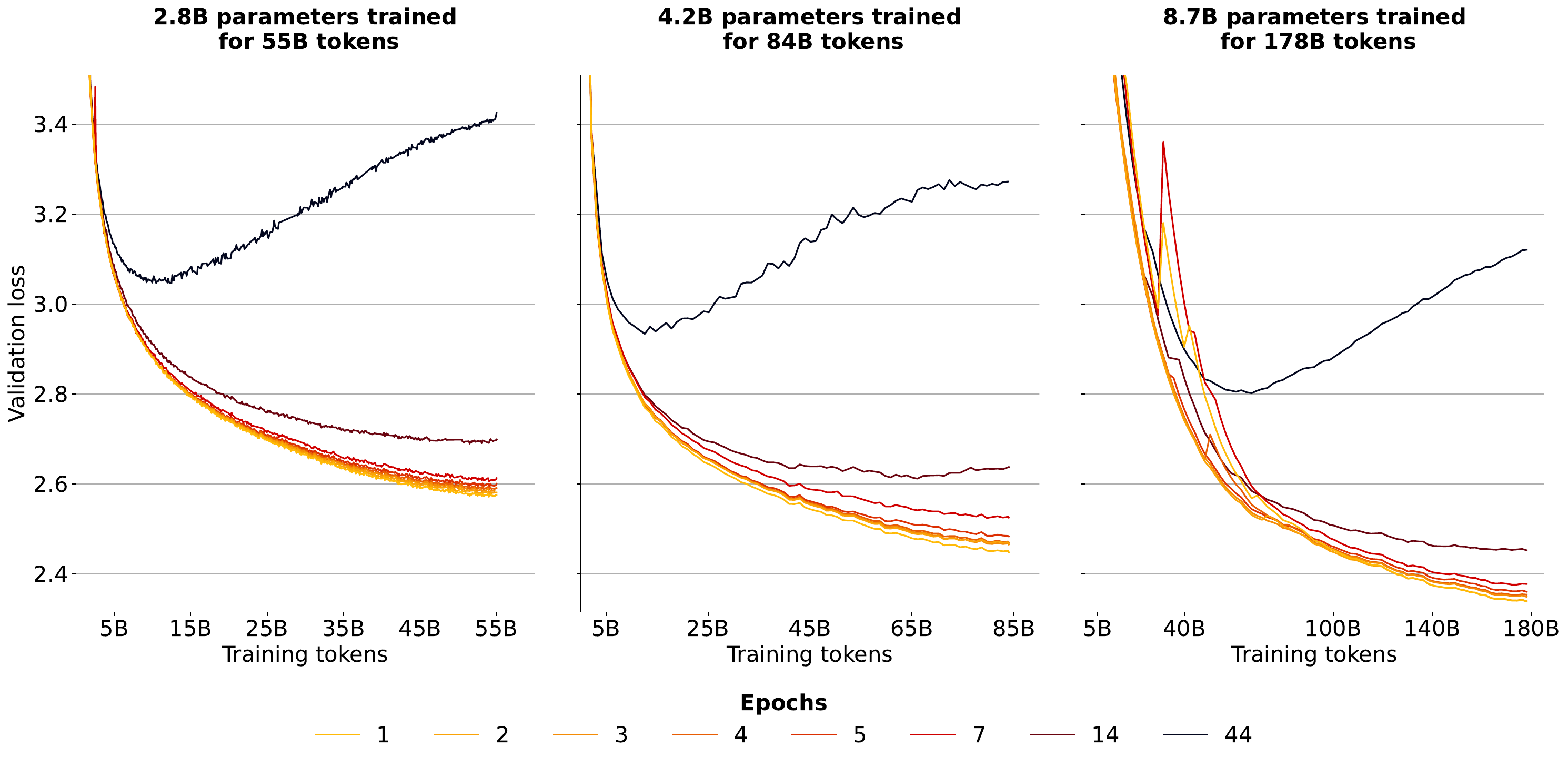}
        \begin{tabular}{c c c c}
\toprule
FLOP budget ($C$) & Parameters ($N$) & Training tokens ($D$) & Data budget ($D_C$) \\
\midrule
$9.3 \times 10^{20}$ & $2.8$B & 55B & $\{\ 55,28,18,14,11,9,4,1.25\}$B\\
$2.1 \times 10^{21}$ & $4.2$B & 84B  & $\{\ 84,42,28,21,17,12,6,1.9\}$B\\
$9.3 \times 10^{21}$ & $8.7$B & 178B & $\{178,88,58,44,35,25,13,4\}$B\\
\bottomrule
\end{tabular}
        \caption{\textbf{Validation Loss for Different Data Constraints (IsoFLOP).} Each curve represents the same number of FLOPs spent on an equal-sized model. Colors represent different numbers of epochs due to repeating because of data constraints. Parameters and training tokens are set to match the single-epoch compute-optimal configurations for the given FLOPs. Models trained on data that is repeated for multiple epochs have consistently worse loss and diverge if too many epochs are used. Only loss curves for 8.7B runs are smoothed with an exponential moving average and weight of 0.85.}
        \label{fig:validation}
    \end{center}
\end{figure*}

For all three data budgets, our results suggest that \emph{Allocation} is optimized by scaling epochs faster than parameters. We confirm this at scale by training the data-constrained compute-optimal model for $9.3 \times 10^{21}$ FLOPs and 25 billion unique tokens as suggested by our efficient frontier. Despite having 27\% fewer parameters, this model achieves better loss and downstream performance than the model suggested by the Chinchilla scaling laws (\autoref{fig:returnalloc} (right) and \autoref{tab:8b7rep}). Similarly, the 120 billion parameter Galactica model trained on repeated data should have been significantly smaller according to data-constrained scaling laws (\autoref{sec:galactica}). An additional benefit of using a smaller model is cheaper inference, though adding parameters can make it easier to parallelize training across GPUs.

Adding parameters and epochs causes the loss to decrease and eventually increase again, suggesting that too much compute can hurt performance. Results from \cite{kaplan2020scaling} also show that loss can increase when too many parameters are used, even with early stopping. However, we expect that appropriate regularization (such as simply removing all excess parameters as an extreme case) could prevent this behavior. Thus, our formula presented in \autoref{sec:method} and its predicted isoLoss contours in \autoref{fig:100misoloss} do not model the possibility that excess epochs or parameters could hurt performance.

\section{Results: Resource Return for Data-Constrained Scaling}
\label{sec:fixc}


\begin{figure*}
    \centering
    \begin{center}
        \includegraphics[width=\textwidth]{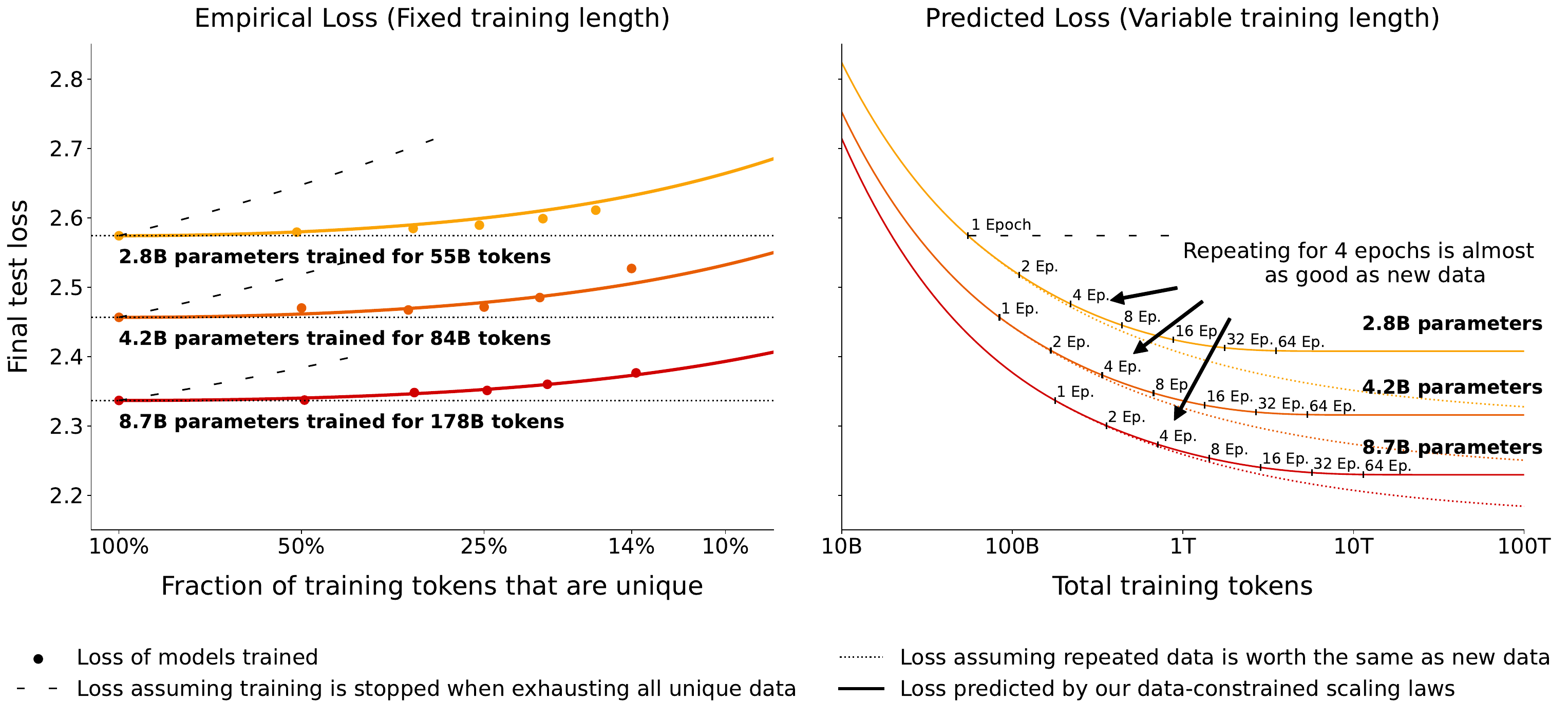}
        \caption{\textbf{Empirical and Extrapolated loss with constrained data.} \textit{(Left):} Loss as a function of repeated tokens for three different training budgets each with fixed number of parameters. Loss curves predicted by our data-constrained scaling laws are shifted to exactly match the loss at 100\% unique data. Return on FLOPs decays with repeated data in a regular pattern. \textit{(Right):} Extrapolating from the proposed data-constrained  scaling law shows that 
        at small numbers epochs are benign, but at large number of epochs loss stops improving. }
        \label{fig:epochs}
    \end{center}
\end{figure*}


Next, consider the question of \textit{Return} on scaling. To quantify this value, we run experiments with three FLOP budgets across eight respective data budgets to compare return on FLOPs. 

\autoref{fig:validation} shows the configurations and validation curves for models trained on the same number of total tokens. Conforming to intuition and prior work on deduplication~\cite{deduplicatinglee2021}, repeated data is worth less, thus models trained on less unique data (and, correspondingly, more epochs) have consistently higher loss. However, the loss difference for a few epochs is negligible. For example, the $N=8.7$ billion parameter model trained for four epochs ($D_C=44$ billion unique tokens) finishes training with only 0.5\% higher validation loss than the single-epoch model ($D_C=178$ billion unique tokens).

In \autoref{fig:epochs} (left), we compare the final test loss of each model to predictions from our parametric fit. The data-constrained scaling laws can accurately measure the decay in the value of repeated data as seen by the proximity of empirical results (dots) and parametric fit (lines). We note however that it significantly underestimates the final test loss of failing models where loss increases midway through training, such as models trained for 44 epochs (not depicted). 

In \autoref{fig:epochs} (right), we extrapolate the three budgets by further scaling compute while keeping the data constraints ($D_C$) at 55B, 84B, and 178B tokens, respectively. The parameter $R_D^*$ introduced in \autoref{sec:method} represents roughly the  ``half-life'' of epochs: specifically the point where repeated tokens have lost  $\tfrac{1}{e}$ of their value. Through our fitting in \autoref{sec:scalinglaws}, we found $R_D^* \approx 15$, corresponding to 15 repetitions (or 16 epochs). Graphically, this can be seen by the stark diminishing returns in the proximity of the 16-epoch marker and the flattening out soon after.

Overall, the \emph{Return} when repeating data is relatively good. Meaningful gains from repeating data can be made up to around 16 epochs ($R_D^*$) beyond which returns diminish extremely fast.

\section{Results: Complementary Strategies for Obtaining Additional Data }
\label{sec:beyond}

\begin{figure*}
    \centering
    {{\includegraphics[width=0.40\textwidth]{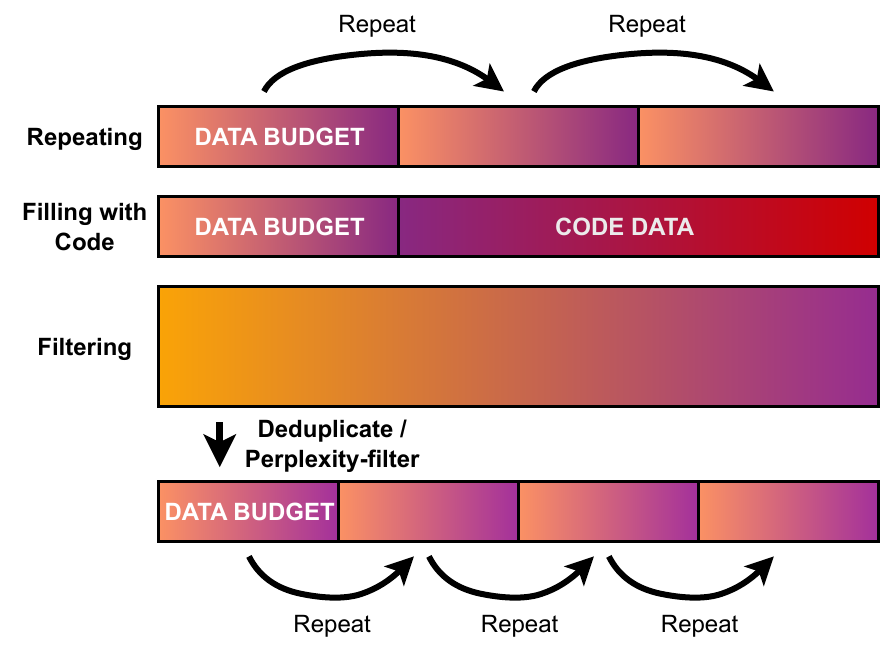}}}
    \qquad
    {{\includegraphics[width=0.48\textwidth]{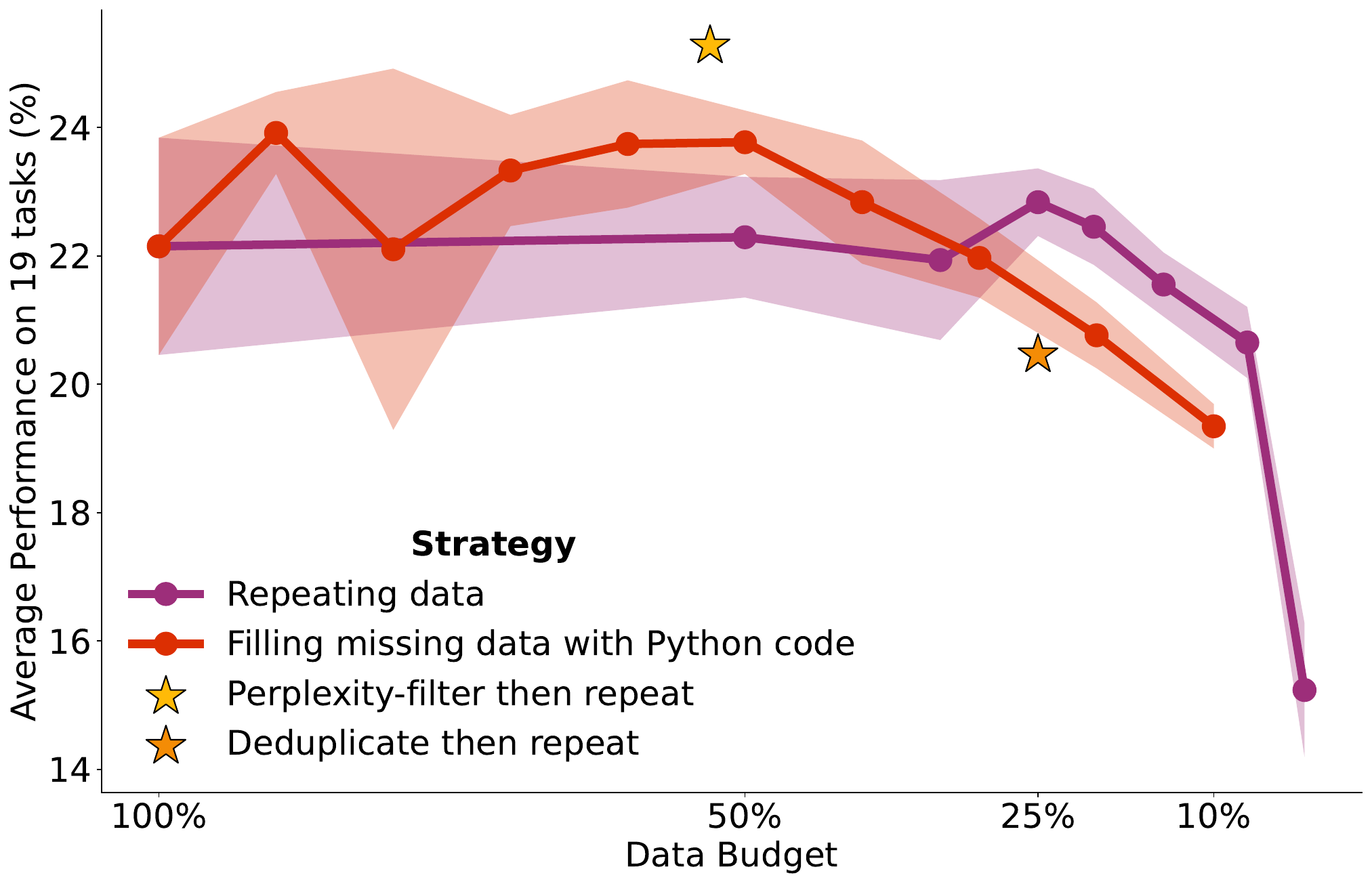}}}
    \caption{\textbf{Strategies for data-constrained settings and their downstream performance.} \emph{(Left):} Schematic showing alternative data use strategies of code filling and filtering. \emph{(Right):} $N=4.2$ billion parameter models trained for a total of $D=84$ billion tokens with varying budgets $D_C$. For repeating and filling with code, five models with different seeds are trained for each dot and the standard deviation is visualized as the shaded area.}
    \label{fig:beyond}
\end{figure*}

While repeating data is effective, it has diminishing returns. We therefore consider strategies for scaling $D$ targeting improved downstream performance as opposed to directly minimizing loss.

\autoref{fig:beyond}~(left) illustrates the strategies: \textbf{(a) Code augmentation:}  We use Python code from The Stack~\cite{kocetkov2022stack} to make up for missing natural language data. The combined dataset consisting of code and natural language samples is shuffled randomly. \textbf{(b) Adapting filtering:} We investigate the performance impact of deduplication and perplexity filtering, two common filtering steps that can severely limit available data. Removing such filtering steps can free up additional training data.

For these experiments, we set a maximum data budget ($D_C$) of 84 billion tokens. For repetition and code filling, only a subset of $D_C$ is available and the rest needs to be compensated for via repeating or adding code. For both filtering methods, we start out with approximately twice the budget (178 billion tokens), as it is easier to gather noisy data and filter it than it is to gather clean data for training. For perplexity filtering, we select the top 25\% samples with the lowest perplexity according to a language model trained on Wikipedia. This results in 44 billion tokens that are repeated for close to two epochs to reach the full data budget. For deduplication filtering, all samples with a 100-char overlap are removed resulting in 21 billion tokens that are repeated for four epochs during training. See \autoref{sec:filtering} for more details on the filtering procedures.

When comparing across data strategies, loss ceases to be a good evaluation metric as the models are trained on different data distributions. We thus evaluate models on 19 natural language tasks with zero to five in-context few-shot exemplars~\cite{brown2020language} producing 114 scores per model. As our evaluation tasks cover different metrics and random baselines, we re-scale all scores to be in the same range to better reflect performance ranges before averaging. Details on the evaluation datasets are in \autoref{sec:eval}.

In \autoref{fig:beyond} (right) we compare the downstream performance of all strategies. For repeating data, differences in downstream performance are insignificant for up to around 4 epochs (25\% budget) and then start dropping, which aligns with our results on test loss in \autoref{sec:fixc}. Filling up to 50\% of data with code (42 billion tokens) also shows no deterioration. Beyond that, performance decreases quickly on natural language tasks. However, adding more code data may benefit non-natural language tasks, which are not considered in the benchmarking. Two of the tasks benchmarked, WebNLG~\cite{castro-ferreira20:bilin-bi-direc-webnl-shared,gehrmann2021gem}, a generation task, and bAbI~\cite{weston2015towards,liang2022holistic}, a reasoning task, see jumps in performance as soon as code is added, possibly due to code enabling models to learn long-range state-tracking capabilities beneficial for these tasks. 

Of the filtering approaches, we find perplexity-filtering to be effective, while deduplication does not help. Prior work found deduplication was able to improve perplexity~\cite{deduplicatinglee2021}; however, it did not evaluate on downstream tasks. Deduplication may have value not captured in our benchmark, such as reducing memorization \cite{kandpal2022deduplicating,hernandez2022scaling,carlini2022quantifying,biderman2023emergent}. We also investigate filtering on a different noisier dataset in \autoref{sec:addfilter}, where we find it to be more effective. Overall, in a data-constrained regime, we recommend reserving filtering for noisy datasets and using both code augmentation and repeating to increase data tokens. For example, first doubling the available data by adding code and then repeating the new dataset for four epochs results in 8$\times$ more training tokens that are expected to be just as good as having had 8$\times$ more unique data from the start.

\section{Related Work}
\noindent

\textbf{Large language models} Scaling up transformer language models~\cite{vaswani2017attention} across parameter count and training data has been shown to result in continuous performance gains~\cite{chowdhery2022palm}. Starting with the 1.4 billion parameter GPT-2 model~\cite{radford2019language}, a variety of scaled-up language models have been trained, commonly referred to as large language models (LLMs). They can be grouped into dense models~\cite{brown2020language,Khrushchev_YaLM_100B_2022,lieber2021jurassic,rae2021scaling,chung2022scaling,black2022gpt,zhang2022opt,thoppilan2022lamda,su2022welm,taylor2022galactica,zeng2022glm,scao2022bloom,li2023starcoder,luukkonen2023fingpt} and sparse models~\cite{fedus2021switch,zeng2021pangu,du2022glam,zoph2022designing} depending on whether each forward pass makes use of all parameters. These models are generally pre-trained to predict the next token in a sequence, which makes them applicable to various language tasks directly after pre-training~\cite{brown2020language,wei2022chain,kojima2022large,muennighoff2022sgpt,srivastava2022beyond} by reformulating said NLP tasks as context continuation tasks (see~\cite{DBLP:journals/corr/abs-1806-08730} for an earlier proposal on this topic). We focus on the most common scenario, where a dense transformer model is trained to do next-token prediction on a large corpus and evaluated directly after pre-training using held-out loss or zero- to few-shot prompting.

\noindent

\textbf{Scaling laws} 
Prior work has estimated an optimal allocation of compute for the training of LLMs. \citet{kaplan2020scaling} suggested a 10$\times$ increase in compute should be allocated to a 5.5$\times$ increase in model size and a 1.8$\times$ increase in training tokens. This first scaling law has led to the creation of very large models trained on relatively little data, such as the 530 billion parameter MT-NLG model trained on 270 billion tokens~\cite{smith2022using}. More recent work~\cite{hoffmann2022training}, however, showed that model size and training data should rather be scaled in equal proportions. These findings called for a renewed focus on the scaling of pre-training data rather than scaling model size via complex parallelization strategies~\cite{shoeybi2019megatron,rasley2020deepspeed,bian2021colossal,narayanan2021efficient}. Up-sampling is often employed when pre-training data is partly limited, such as data from a high-quality domain like Wikipedia or text in a rare language for training multilingual LLMs~\cite{lin2021few,orlanski2023measuring}. \citet{hernandez2022scaling} study up-sampling of data subsets and find that repeating only 0.1\% of training data 100 times significantly degrades performance. In contrast, our work focuses on repeating the entire pre-training corpus for multiple epochs rather than up-sampling parts of it.

\noindent
\textbf{Alternative data strategies} Large pre-training datasets are commonly filtered to remove undesired samples or reduce noise~\cite{sorscher2022beyond}. Perplexity-based filtering, whereby a trained model is used to filter out samples with high perplexity, has been found beneficial to reduce noise in web-crawled datasets~\cite{wenzek2019ccnet}. Mixing of data is employed for the pre-training data of multilingual LLMs, where text data from different languages is combined~\cite{conneau2019unsupervised,xue2020mt5,soltan2022alexatm,muennighoff2022crosslingual}. However, both for code and natural language models, mixing different (programming) languages has been reported to under-perform monolingual models~\cite{nijkamp2022codegen,virtanen2019multilingual}. Some work has investigated mixing code and natural language data for prediction tasks, such as summarizing code snippets~\cite{iyer2016summarizing} or predicting function names~\cite{allamanis2015suggesting}. Several pre-training datasets for LLMs include low amounts of code data~\cite{gao2020pile,rae2021scaling,scao2022bloom}. However, these past works generally do not provide any ablation on the drawbacks of including code or the benefits for natural language task performance. We perform a detailed benchmarking of mixing Python and natural language in LLM pre-training at 10 different mixing rates.

\section{Conclusion}

This work studies data-constrained scaling, focusing on the optimal use of computational resources when unique data is limited. We propose an extension to the Chinchilla scaling laws that takes into account the decay in value of repeated data, and we fit this function using a large set of controlled experiments. We find that despite recommendations of earlier work, training large language models for multiple epochs by repeating data is beneficial and that scaling laws continue to hold in the multi-epoch regime, albeit with diminishing returns. We also consider complementary approaches to continue scaling models, and find that code gives the ability to scale an additional 2$\times$. We believe that our findings will enable further scaling of language models to unlock new capabilities with current data. However, our work also indicates that there are limits on the scaling horizon. In addition to collecting additional data, researchers should explore using current data in a more effective manner. 

\begin{ack}

This work was co-funded by the European Union under grant agreement No 101070350. The authors wish to acknowledge CSC – IT Center for Science, Finland, for generous computational resources on the LUMI supercomputer.\footnote{\url{https://www.lumi-supercomputer.eu/}} We are thankful for the immense support from teams at LUMI and AMD, especially Samuel Antao. Hugging Face provided storage and additional compute instances. This work was supported by a Simons Investigator Fellowship, NSF grant DMS-2134157, DARPA grant W911NF2010021, and DOE grant DE-SC0022199. We are grateful to Harm de Vries, Woojeong Kim, Mengzhou Xia and the EleutherAI community for exceptional feedback. We thank Loubna Ben Allal for help with the Python data and Big Code members for insightful discussions on scaling laws. We thank Thomas Wang, Helen Ngo and TurkuNLP members for support on early experiments. We thank Mengning Wu for pointing out a bug in our $R^2$ calculation.

\end{ack}

\bibliography{custom}

\begin{thebibliography}{135}
\expandafter\ifx\csname natexlab\endcsname\relax\def\natexlab#1{#1}\fi

\bibitem[{Aghajanyan et~al.(2023)Aghajanyan, Yu, Conneau, Hsu, Hambardzumyan,
  Zhang, Roller, Goyal, Levy, and Zettlemoyer}]{aghajanyan2023scaling}
Armen Aghajanyan, Lili Yu, Alexis Conneau, Wei-Ning Hsu, Karen Hambardzumyan,
  Susan Zhang, Stephen Roller, Naman Goyal, Omer Levy, and Luke Zettlemoyer.
  2023.
\newblock Scaling Laws for Generative Mixed-Modal Language Models.
\newblock \emph{arXiv preprint arXiv:2301.03728}.

\bibitem[{Alabdulmohsin et~al.(2022)Alabdulmohsin, Neyshabur, and
  Zhai}]{alabdulmohsin2022revisiting}
Ibrahim~M Alabdulmohsin, Behnam Neyshabur, and Xiaohua Zhai. 2022.
\newblock Revisiting neural scaling laws in language and vision.
\newblock \emph{Advances in Neural Information Processing Systems},
  35:22300--22312.

\bibitem[{Allal et~al.(2023)Allal, Li, Kocetkov, Mou, Akiki, Ferrandis,
  Muennighoff, Mishra, Gu, Dey et~al.}]{allal2023santacoder}
Loubna~Ben Allal, Raymond Li, Denis Kocetkov, Chenghao Mou, Christopher Akiki,
  Carlos~Munoz Ferrandis, Niklas Muennighoff, Mayank Mishra, Alex Gu, Manan
  Dey, et~al. 2023.
\newblock SantaCoder: don't reach for the stars!
\newblock \emph{arXiv preprint arXiv:2301.03988}.

\bibitem[{Allamanis et~al.(2015)Allamanis, Barr, Bird, and
  Sutton}]{allamanis2015suggesting}
Miltiadis Allamanis, Earl~T Barr, Christian Bird, and Charles Sutton. 2015.
\newblock Suggesting accurate method and class names.
\newblock In \emph{Proceedings of the 2015 10th joint meeting on foundations of
  software engineering}, pages 38--49.

\bibitem[{Bach et~al.(2022)Bach, Sanh, Yong, Webson, Raffel, Nayak, Sharma,
  Kim, Bari, Fevry, Alyafeai, Dey, Santilli, Sun, Ben-David, Xu, Chhablani,
  Wang, Fries, Al-shaibani, Sharma, Thakker, Almubarak, Tang, Tang, Jiang, and
  Rush}]{promptsource}
Stephen~H. Bach, Victor Sanh, Zheng-Xin Yong, Albert Webson, Colin Raffel,
  Nihal~V. Nayak, Abheesht Sharma, Taewoon Kim, M~Saiful Bari, Thibault Fevry,
  Zaid Alyafeai, Manan Dey, Andrea Santilli, Zhiqing Sun, Srulik Ben-David,
  Canwen Xu, Gunjan Chhablani, Han Wang, Jason~Alan Fries, Maged~S.
  Al-shaibani, Shanya Sharma, Urmish Thakker, Khalid Almubarak, Xiangru Tang,
  Xiangru Tang, Mike Tian-Jian Jiang, and Alexander~M. Rush. 2022.
\newblock \href {http://arxiv.org/abs/2202.01279} {PromptSource: An Integrated
  Development Environment and Repository for Natural Language Prompts}.

\bibitem[{Bahri et~al.(2021)Bahri, Dyer, Kaplan, Lee, and
  Sharma}]{bahri2021explaining}
Yasaman Bahri, Ethan Dyer, Jared Kaplan, Jaehoon Lee, and Utkarsh Sharma. 2021.
\newblock Explaining neural scaling laws.
\newblock \emph{arXiv preprint arXiv:2102.06701}.

\bibitem[{Bansal et~al.(2022)Bansal, Ghorbani, Garg, Zhang, Cherry, Neyshabur,
  and Firat}]{bansal2022data}
Yamini Bansal, Behrooz Ghorbani, Ankush Garg, Biao Zhang, Colin Cherry, Behnam
  Neyshabur, and Orhan Firat. 2022.
\newblock Data scaling laws in NMT: The effect of noise and architecture.
\newblock In \emph{International Conference on Machine Learning}, pages
  1466--1482. PMLR.

\bibitem[{Bender et~al.(2021)Bender, Gebru, McMillan-Major, and
  Shmitchell}]{bender2021dangers}
Emily~M. Bender, Timnit Gebru, Angelina McMillan-Major, and Shmargaret
  Shmitchell. 2021.
\newblock On the Dangers of Stochastic Parrots: Can Language Models Be Too Big?
\newblock In \emph{Proceedings of the 2021 ACM Conference on Fairness,
  Accountability, and Transparency}, pages 610--623.

\bibitem[{Bian et~al.(2021)Bian, Liu, Wang, Huang, Li, Wang, Cui, and
  You}]{bian2021colossal}
Zhengda Bian, Hongxin Liu, Boxiang Wang, Haichen Huang, Yongbin Li, Chuanrui
  Wang, Fan Cui, and Yang You. 2021.
\newblock Colossal-AI: A unified deep learning system for large-scale parallel
  training.
\newblock \emph{arXiv preprint arXiv:2110.14883}.

\bibitem[{Biderman et~al.(2023{\natexlab{a}})Biderman, Prashanth, Sutawika,
  Schoelkopf, Anthony, Purohit, and Raf}]{biderman2023emergent}
Stella Biderman, USVSN~Sai Prashanth, Lintang Sutawika, Hailey Schoelkopf,
  Quentin Anthony, Shivanshu Purohit, and Edward Raf. 2023{\natexlab{a}}.
\newblock Emergent and Predictable Memorization in Large Language Models.
\newblock \emph{arXiv preprint arXiv:2304.11158}.

\bibitem[{Biderman et~al.(2023{\natexlab{b}})Biderman, Schoelkopf, Anthony,
  Bradley, O'Brien, Hallahan, Khan, Purohit, Prashanth, Raff
  et~al.}]{biderman2023pythia}
Stella Biderman, Hailey Schoelkopf, Quentin Anthony, Herbie Bradley, Kyle
  O'Brien, Eric Hallahan, Mohammad~Aflah Khan, Shivanshu Purohit, USVSN~Sai
  Prashanth, Edward Raff, et~al. 2023{\natexlab{b}}.
\newblock Pythia: A suite for analyzing large language models across training
  and scaling.
\newblock \emph{arXiv preprint arXiv:2304.01373}.

\bibitem[{Bisk et~al.(2020)Bisk, Zellers, Bras, Gao, and Choi}]{Bisk2020}
Yonatan Bisk, Rowan Zellers, Ronan~Le Bras, Jianfeng Gao, and Yejin Choi. 2020.
\newblock PIQA: Reasoning about Physical Commonsense in Natural Language.
\newblock In \emph{Thirty-Fourth AAAI Conference on Artificial Intelligence}.

\bibitem[{Black et~al.(2022)Black, Biderman, Hallahan, Anthony, Gao, Golding,
  He, Leahy, McDonell, Phang et~al.}]{black2022gpt}
Sid Black, Stella Biderman, Eric Hallahan, Quentin Anthony, Leo Gao, Laurence
  Golding, Horace He, Connor Leahy, Kyle McDonell, Jason Phang, et~al. 2022.
\newblock GPT-NeoX-20B: An Open-Source Autoregressive Language Model.
\newblock \emph{arXiv preprint arXiv:2204.06745}.

\bibitem[{Black et~al.(2021)Black, Gao, Wang, Leahy, and
  Biderman}]{black2021gpt}
Sid Black, Leo Gao, Phil Wang, Connor Leahy, and Stella Biderman. 2021.
\newblock GPT-Neo: Large scale autoregressive language modeling with
  mesh-tensorflow.
\newblock \emph{If you use this software, please cite it using these metadata},
  58.

\bibitem[{Brown et~al.(2020)Brown, Mann, Ryder, Subbiah, Kaplan, Dhariwal,
  Neelakantan, Shyam, Sastry, Askell et~al.}]{brown2020language}
Tom Brown, Benjamin Mann, Nick Ryder, Melanie Subbiah, Jared~D Kaplan, Prafulla
  Dhariwal, Arvind Neelakantan, Pranav Shyam, Girish Sastry, Amanda Askell,
  et~al. 2020.
\newblock Language models are few-shot learners.
\newblock \emph{Advances in neural information processing systems},
  33:1877--1901.

\bibitem[{Carlini et~al.(2022)Carlini, Ippolito, Jagielski, Lee, Tramer, and
  Zhang}]{carlini2022quantifying}
Nicholas Carlini, Daphne Ippolito, Matthew Jagielski, Katherine Lee, Florian
  Tramer, and Chiyuan Zhang. 2022.
\newblock Quantifying memorization across neural language models.
\newblock \emph{arXiv preprint arXiv:2202.07646}.

\bibitem[{Castro~Ferreira et~al.(2020)Castro~Ferreira, Gardent, Ilinykh,
  van~der Lee, Mille, Moussallem, and
  Shimorina}]{castro-ferreira20:bilin-bi-direc-webnl-shared}
Thiago Castro~Ferreira, Claire Gardent, Nikolai Ilinykh, Chris van~der Lee,
  Simon Mille, Diego Moussallem, and Anastasia Shimorina. 2020.
\newblock The 2020 Bilingual, Bi-Directional WebNLG+ Shared Task Overview and
  Evaluation Results (WebNLG+ 2020).
\newblock In \emph{Proceedings of the 3rd WebNLG Workshop on Natural Language
  Generation from the Semantic Web (WebNLG+ 2020)}, pages 55--76, Dublin,
  Ireland (Virtual). Association for Computational Linguistics.

\bibitem[{Chen et~al.(2020)Chen, Radford, Child, Wu, Jun, Luan, and
  Sutskever}]{chen2020generative}
Mark Chen, Alec Radford, Rewon Child, Jeffrey Wu, Heewoo Jun, David Luan, and
  Ilya Sutskever. 2020.
\newblock Generative pretraining from pixels.
\newblock In \emph{International conference on machine learning}, pages
  1691--1703. PMLR.

\bibitem[{Chowdhery et~al.(2022)Chowdhery, Narang, Devlin, Bosma, Mishra,
  Roberts, Barham, Chung, Sutton, Gehrmann et~al.}]{chowdhery2022palm}
Aakanksha Chowdhery, Sharan Narang, Jacob Devlin, Maarten Bosma, Gaurav Mishra,
  Adam Roberts, Paul Barham, Hyung~Won Chung, Charles Sutton, Sebastian
  Gehrmann, et~al. 2022.
\newblock Palm: Scaling language modeling with pathways.
\newblock \emph{arXiv preprint arXiv:2204.02311}.

\bibitem[{Chung et~al.(2022)Chung, Hou, Longpre, Zoph, Tay, Fedus, Li, Wang,
  Dehghani, Brahma, Webson, Gu, Dai, Suzgun, Chen, Chowdhery, Narang, Mishra,
  Yu, Zhao, Huang, Dai, Yu, Petrov, Chi, Dean, Devlin, Roberts, Zhou, Le, and
  Wei}]{chung2022scaling}
Hyung~Won Chung, Le~Hou, Shayne Longpre, Barret Zoph, Yi~Tay, William Fedus,
  Eric Li, Xuezhi Wang, Mostafa Dehghani, Siddhartha Brahma, Albert Webson,
  Shixiang~Shane Gu, Zhuyun Dai, Mirac Suzgun, Xinyun Chen, Aakanksha
  Chowdhery, Sharan Narang, Gaurav Mishra, Adams Yu, Vincent Zhao, Yanping
  Huang, Andrew Dai, Hongkun Yu, Slav Petrov, Ed~H. Chi, Jeff Dean, Jacob
  Devlin, Adam Roberts, Denny Zhou, Quoc~V. Le, and Jason Wei. 2022.
\newblock \href {https://arxiv.org/abs/2210.11416} {Scaling
  Instruction-Finetuned Language Models}.
\newblock \emph{arXiv preprint arXiv:2210.11416}.

\bibitem[{Clark et~al.(2019)Clark, Lee, Chang, Kwiatkowski, Collins, and
  Toutanova}]{clark2019boolq}
Christopher Clark, Kenton Lee, Ming-Wei Chang, Tom Kwiatkowski, Michael
  Collins, and Kristina Toutanova. 2019.
\newblock BoolQ: Exploring the Surprising Difficulty of Natural Yes/No
  Questions.
\newblock In \emph{NAACL}.

\bibitem[{Clark et~al.(2018)Clark, Cowhey, Etzioni, Khot, Sabharwal, Schoenick,
  and Tafjord}]{allenai:arc}
Peter Clark, Isaac Cowhey, Oren Etzioni, Tushar Khot, Ashish Sabharwal, Carissa
  Schoenick, and Oyvind Tafjord. 2018.
\newblock Think you have Solved Question Answering? Try ARC, the AI2 Reasoning
  Challenge.
\newblock \emph{arXiv:1803.05457v1}.

\bibitem[{Conneau et~al.(2019)Conneau, Khandelwal, Goyal, Chaudhary, Wenzek,
  Guzm{\'a}n, Grave, Ott, Zettlemoyer, and Stoyanov}]{conneau2019unsupervised}
Alexis Conneau, Kartikay Khandelwal, Naman Goyal, Vishrav Chaudhary, Guillaume
  Wenzek, Francisco Guzm{\'a}n, Edouard Grave, Myle Ott, Luke Zettlemoyer, and
  Veselin Stoyanov. 2019.
\newblock Unsupervised Cross-lingual Representation Learning at Scale.
\newblock \emph{arXiv preprint arXiv:1911.02116}.

\bibitem[{Dagan et~al.(2006)Dagan, Glickman, and Magnini}]{dagan2006pascal}
Ido Dagan, Oren Glickman, and Bernardo Magnini. 2006.
\newblock The pascal recognising textual entailment challenge.
\newblock In \emph{Machine Learning Challenges. Evaluating Predictive
  Uncertainty, Visual Object Classification, and Recognising Tectual
  Entailment: First PASCAL Machine Learning Challenges Workshop, MLCW 2005,
  Southampton, UK, April 11-13, 2005, Revised Selected Papers}, pages 177--190.
  Springer.

\bibitem[{De~Marneffe et~al.(2019)De~Marneffe, Simons, and
  Tonhauser}]{de2019commitmentbank}
Marie-Catherine De~Marneffe, Mandy Simons, and Judith Tonhauser. 2019.
\newblock The commitmentbank: Investigating projection in naturally occurring
  discourse.
\newblock In \emph{proceedings of Sinn und Bedeutung}, volume~23, pages
  107--124.

\bibitem[{Dehghani et~al.(2023)Dehghani, Djolonga, Mustafa, Padlewski, Heek,
  Gilmer, Steiner, Caron, Geirhos, Alabdulmohsin et~al.}]{dehghani2023scaling}
Mostafa Dehghani, Josip Djolonga, Basil Mustafa, Piotr Padlewski, Jonathan
  Heek, Justin Gilmer, Andreas Steiner, Mathilde Caron, Robert Geirhos, Ibrahim
  Alabdulmohsin, et~al. 2023.
\newblock Scaling vision transformers to 22 billion parameters.
\newblock \emph{arXiv preprint arXiv:2302.05442}.

\bibitem[{Dehghani et~al.(2018)Dehghani, Gouws, Vinyals, Uszkoreit, and
  Kaiser}]{dehghani2018universal}
Mostafa Dehghani, Stephan Gouws, Oriol Vinyals, Jakob Uszkoreit, and {\L}ukasz
  Kaiser. 2018.
\newblock Universal transformers.
\newblock \emph{arXiv preprint arXiv:1807.03819}.

\bibitem[{Du et~al.(2022)Du, Huang, Dai, Tong, Lepikhin, Xu, Krikun, Zhou, Yu,
  Firat et~al.}]{du2022glam}
Nan Du, Yanping Huang, Andrew~M Dai, Simon Tong, Dmitry Lepikhin, Yuanzhong Xu,
  Maxim Krikun, Yanqi Zhou, Adams~Wei Yu, Orhan Firat, et~al. 2022.
\newblock Glam: Efficient scaling of language models with mixture-of-experts.
\newblock In \emph{International Conference on Machine Learning}, pages
  5547--5569. PMLR.

\bibitem[{Du{\v{s}}ek et~al.(2020)Du{\v{s}}ek, Novikova, and
  Rieser}]{dusek.etal2020:csl}
Ond\v{r}ej Du{\v{s}}ek, Jekaterina Novikova, and Verena Rieser. 2020.
\newblock \href {https://doi.org/10.1016/j.csl.2019.06.009} {Evaluating the
  {{State}}-of-the-{{Art}} of {{End}}-to-{{End Natural Language Generation}}:
  {{The E2E NLG Challenge}}}.
\newblock \emph{Computer Speech \& Language}, 59:123--156.

\bibitem[{Fedus et~al.(2021)Fedus, Zoph, and Shazeer}]{fedus2021switch}
William Fedus, Barret Zoph, and Noam Shazeer. 2021.
\newblock Switch transformers: Scaling to trillion parameter models with simple
  and efficient sparsity.
\newblock \emph{J. Mach. Learn. Res}, 23:1--40.

\bibitem[{Gao et~al.(2020)Gao, Biderman, Black, Golding, Hoppe, Foster, Phang,
  He, Thite, Nabeshima, Presser, and Leahy}]{gao2020pile}
Leo Gao, Stella Biderman, Sid Black, Laurence Golding, Travis Hoppe, Charles
  Foster, Jason Phang, Horace He, Anish Thite, Noa Nabeshima, Shawn Presser,
  and Connor Leahy. 2020.
\newblock The {P}ile: An 800{GB} Dataset of Diverse Text for Language Modeling.
\newblock \emph{arXiv preprint arXiv:2101.00027}.

\bibitem[{Gao et~al.(2021)Gao, Tow, Biderman, Black, DiPofi, Foster, Golding,
  Hsu, McDonell, Muennighoff, Phang, Reynolds, Tang, Thite, Wang, Wang, and
  Zou}]{eval-harness}
Leo Gao, Jonathan Tow, Stella Biderman, Sid Black, Anthony DiPofi, Charles
  Foster, Laurence Golding, Jeffrey Hsu, Kyle McDonell, Niklas Muennighoff,
  Jason Phang, Laria Reynolds, Eric Tang, Anish Thite, Ben Wang, Kevin Wang,
  and Andy Zou. 2021.
\newblock \href {https://doi.org/10.5281/zenodo.5371628} {A framework for
  few-shot language model evaluation}.

\bibitem[{Gehman et~al.(2020)Gehman, Gururangan, Sap, Choi, and
  Smith}]{gehman2020realtoxicityprompts}
Samuel Gehman, Suchin Gururangan, Maarten Sap, Yejin Choi, and Noah~A Smith.
  2020.
\newblock Realtoxicityprompts: Evaluating neural toxic degeneration in language
  models.
\newblock \emph{arXiv preprint arXiv:2009.11462}.

\bibitem[{Gehrmann et~al.(2021)Gehrmann, Adewumi, Aggarwal, Ammanamanchi,
  Anuoluwapo, Bosselut, Chandu, Clinciu, Das, Dhole et~al.}]{gehrmann2021gem}
Sebastian Gehrmann, Tosin Adewumi, Karmanya Aggarwal, Pawan~Sasanka
  Ammanamanchi, Aremu Anuoluwapo, Antoine Bosselut, Khyathi~Raghavi Chandu,
  Miruna Clinciu, Dipanjan Das, Kaustubh~D Dhole, et~al. 2021.
\newblock The gem benchmark: Natural language generation, its evaluation and
  metrics.
\newblock \emph{arXiv preprint arXiv:2102.01672}.

\bibitem[{Ghorbani et~al.(2021)Ghorbani, Firat, Freitag, Bapna, Krikun, Garcia,
  Chelba, and Cherry}]{ghorbani2021scaling}
Behrooz Ghorbani, Orhan Firat, Markus Freitag, Ankur Bapna, Maxim Krikun,
  Xavier Garcia, Ciprian Chelba, and Colin Cherry. 2021.
\newblock Scaling laws for neural machine translation.
\newblock \emph{arXiv preprint arXiv:2109.07740}.

\bibitem[{Gupta et~al.(2023)Gupta, Sawant, Mishra, Nakamura, Mitra, Mashetty,
  and Baral}]{gupta2023instruction}
Himanshu Gupta, Saurabh~Arjun Sawant, Swaroop Mishra, Mutsumi Nakamura, Arindam
  Mitra, Santosh Mashetty, and Chitta Baral. 2023.
\newblock Instruction Tuned Models are Quick Learners.
\newblock \emph{arXiv preprint arXiv:2306.05539}.

\bibitem[{Heafield(2011)}]{heafield-2011-kenlm}
Kenneth Heafield. 2011.
\newblock \href {https://aclanthology.org/W11-2123} {{K}en{LM}: Faster and
  Smaller Language Model Queries}.
\newblock In \emph{Proceedings of the Sixth Workshop on Statistical Machine
  Translation}, pages 187--197, Edinburgh, Scotland. Association for
  Computational Linguistics.

\bibitem[{Henderson et~al.(2022)Henderson, Krass, Zheng, Guha, Manning,
  Jurafsky, and Ho}]{henderson2022pile}
Peter Henderson, Mark Krass, Lucia Zheng, Neel Guha, Christopher~D Manning, Dan
  Jurafsky, and Daniel Ho. 2022.
\newblock Pile of law: Learning responsible data filtering from the law and a
  256gb open-source legal dataset.
\newblock \emph{Advances in Neural Information Processing Systems},
  35:29217--29234.

\bibitem[{Henighan et~al.(2020)Henighan, Kaplan, Katz, Chen, Hesse, Jackson,
  Jun, Brown, Dhariwal, Gray et~al.}]{henighan2020scaling}
Tom Henighan, Jared Kaplan, Mor Katz, Mark Chen, Christopher Hesse, Jacob
  Jackson, Heewoo Jun, Tom~B Brown, Prafulla Dhariwal, Scott Gray, et~al. 2020.
\newblock Scaling laws for autoregressive generative modeling.
\newblock \emph{arXiv preprint arXiv:2010.14701}.

\bibitem[{Hernandez et~al.(2022)Hernandez, Brown, Conerly, DasSarma, Drain,
  El-Showk, Elhage, Hatfield-Dodds, Henighan, Hume
  et~al.}]{hernandez2022scaling}
Danny Hernandez, Tom Brown, Tom Conerly, Nova DasSarma, Dawn Drain, Sheer
  El-Showk, Nelson Elhage, Zac Hatfield-Dodds, Tom Henighan, Tristan Hume,
  et~al. 2022.
\newblock Scaling Laws and Interpretability of Learning from Repeated Data.
\newblock \emph{arXiv preprint arXiv:2205.10487}.

\bibitem[{Hernandez et~al.(2021)Hernandez, Kaplan, Henighan, and
  McCandlish}]{hernandez2021scaling}
Danny Hernandez, Jared Kaplan, Tom Henighan, and Sam McCandlish. 2021.
\newblock Scaling laws for transfer.
\newblock \emph{arXiv preprint arXiv:2102.01293}.

\bibitem[{Hoffmann et~al.(2022)Hoffmann, Borgeaud, Mensch, Buchatskaya, Cai,
  Rutherford, Casas, Hendricks, Welbl, Clark et~al.}]{hoffmann2022training}
Jordan Hoffmann, Sebastian Borgeaud, Arthur Mensch, Elena Buchatskaya, Trevor
  Cai, Eliza Rutherford, Diego de~Las Casas, Lisa~Anne Hendricks, Johannes
  Welbl, Aidan Clark, et~al. 2022.
\newblock Training Compute-Optimal Large Language Models.
\newblock \emph{arXiv preprint arXiv:2203.15556}.

\bibitem[{Huang et~al.(2018)Huang, Vaswani, Uszkoreit, Shazeer, Simon,
  Hawthorne, Dai, Hoffman, Dinculescu, and Eck}]{huang2018music}
Cheng-Zhi~Anna Huang, Ashish Vaswani, Jakob Uszkoreit, Noam Shazeer, Ian Simon,
  Curtis Hawthorne, Andrew~M Dai, Matthew~D Hoffman, Monica Dinculescu, and
  Douglas Eck. 2018.
\newblock Music transformer.
\newblock \emph{arXiv preprint arXiv:1809.04281}.

\bibitem[{Iyer et~al.(2016)Iyer, Konstas, Cheung, and
  Zettlemoyer}]{iyer2016summarizing}
Srinivasan Iyer, Ioannis Konstas, Alvin Cheung, and Luke Zettlemoyer. 2016.
\newblock Summarizing source code using a neural attention model.
\newblock In \emph{Proceedings of the 54th Annual Meeting of the Association
  for Computational Linguistics (Volume 1: Long Papers)}, pages 2073--2083.

\bibitem[{Kandpal et~al.(2022)Kandpal, Wallace, and
  Raffel}]{kandpal2022deduplicating}
Nikhil Kandpal, Eric Wallace, and Colin Raffel. 2022.
\newblock \href {http://arxiv.org/abs/2202.06539} {Deduplicating Training Data
  Mitigates Privacy Risks in Language Models}.

\bibitem[{Kaplan et~al.(2020)Kaplan, McCandlish, Henighan, Brown, Chess, Child,
  Gray, Radford, Wu, and Amodei}]{kaplan2020scaling}
Jared Kaplan, Sam McCandlish, Tom Henighan, Tom~B Brown, Benjamin Chess, Rewon
  Child, Scott Gray, Alec Radford, Jeffrey Wu, and Dario Amodei. 2020.
\newblock Scaling laws for neural language models.
\newblock \emph{arXiv preprint arXiv:2001.08361}.

\bibitem[{Khrushchev et~al.(2022)Khrushchev, Vasilev, Petrov, and
  Zinov}]{Khrushchev_YaLM_100B_2022}
Mikhail Khrushchev, Ruslan Vasilev, Alexey Petrov, and Nikolay Zinov. 2022.
\newblock \href {https://github.com/yandex/YaLM-100B} {{YaLM 100B}}.

\bibitem[{Kingma and Ba(2014)}]{kingma2014adam}
Diederik~P Kingma and Jimmy Ba. 2014.
\newblock Adam: A method for stochastic optimization.
\newblock \emph{arXiv preprint arXiv:1412.6980}.

\bibitem[{Kocetkov et~al.(2022)Kocetkov, Li, Allal, Li, Mou, Ferrandis,
  Jernite, Mitchell, Hughes, Wolf et~al.}]{kocetkov2022stack}
Denis Kocetkov, Raymond Li, Loubna~Ben Allal, Jia Li, Chenghao Mou,
  Carlos~Mu{\~n}oz Ferrandis, Yacine Jernite, Margaret Mitchell, Sean Hughes,
  Thomas Wolf, et~al. 2022.
\newblock The Stack: 3 TB of permissively licensed source code.
\newblock \emph{arXiv preprint arXiv:2211.15533}.

\bibitem[{Kojima et~al.(2022)Kojima, Gu, Reid, Matsuo, and
  Iwasawa}]{kojima2022large}
Takeshi Kojima, Shixiang~Shane Gu, Machel Reid, Yutaka Matsuo, and Yusuke
  Iwasawa. 2022.
\newblock Large language models are zero-shot reasoners.
\newblock \emph{arXiv preprint arXiv:2205.11916}.

\bibitem[{Komatsuzaki(2019)}]{komatsuzaki2019one}
Aran Komatsuzaki. 2019.
\newblock One epoch is all you need.
\newblock \emph{arXiv preprint arXiv:1906.06669}.

\bibitem[{Kudo and Richardson(2018)}]{kudo-richardson-2018-sentencepiece}
Taku Kudo and John Richardson. 2018.
\newblock \href {https://doi.org/10.18653/v1/D18-2012} {{S}entence{P}iece: A
  simple and language independent subword tokenizer and detokenizer for Neural
  Text Processing}.
\newblock In \emph{Proceedings of the 2018 Conference on Empirical Methods in
  Natural Language Processing: System Demonstrations}, pages 66--71, Brussels,
  Belgium. Association for Computational Linguistics.

\bibitem[{Ladhak et~al.(2020)Ladhak, Durmus, Cardie, and
  McKeown}]{ladhak2020wikilingua}
Faisal Ladhak, Esin Durmus, Claire Cardie, and Kathleen McKeown. 2020.
\newblock WikiLingua: A new benchmark dataset for cross-lingual abstractive
  summarization.
\newblock \emph{arXiv preprint arXiv:2010.03093}.

\bibitem[{Lauren{\c{c}}on et~al.(2022)Lauren{\c{c}}on, Saulnier, Wang, Akiki,
  del Moral, Le~Scao, Von~Werra, Mou, Ponferrada, Nguyen
  et~al.}]{laurencconbigscience}
Hugo Lauren{\c{c}}on, Lucile Saulnier, Thomas Wang, Christopher Akiki,
  Albert~Villanova del Moral, Teven Le~Scao, Leandro Von~Werra, Chenghao Mou,
  Eduardo~Gonz{\'a}lez Ponferrada, Huu Nguyen, et~al. 2022.
\newblock The BigScience ROOTS Corpus: A 1.6 TB Composite Multilingual Dataset.
\newblock In \emph{Thirty-sixth Conference on Neural Information Processing
  Systems Datasets and Benchmarks Track}.

\bibitem[{Lee et~al.(2021)Lee, Ippolito, Nystrom, Zhang, Eck, Callison-Burch,
  and Carlini}]{deduplicatinglee2021}
Katherine Lee, Daphne Ippolito, Andrew Nystrom, Chiyuan Zhang, Douglas Eck,
  Chris Callison-Burch, and Nicholas Carlini. 2021.
\newblock Deduplicating Training Data Makes Language Models Better.
\newblock \emph{arXiv preprint arXiv:2107.06499}.

\bibitem[{Li et~al.(2023)Li, Allal, Zi, Muennighoff, Kocetkov, Mou, Marone,
  Akiki, Li, Chim, Liu, Zheltonozhskii, Zhuo, Wang, Dehaene, Davaadorj,
  Lamy-Poirier, Monteiro, Shliazhko, Gontier, Meade, Zebaze, Yee, Umapathi,
  Zhu, Lipkin, Oblokulov, Wang, Murthy, Stillerman, Patel, Abulkhanov, Zocca,
  Dey, Zhang, Fahmy, Bhattacharyya, Yu, Singh, Luccioni, Villegas, Kunakov,
  Zhdanov, Romero, Lee, Timor, Ding, Schlesinger, Schoelkopf, Ebert, Dao,
  Mishra, Gu, Robinson, Anderson, Dolan-Gavitt, Contractor, Reddy, Fried,
  Bahdanau, Jernite, Ferrandis, Hughes, Wolf, Guha, von Werra, and
  de~Vries}]{li2023starcoder}
Raymond Li, Loubna~Ben Allal, Yangtian Zi, Niklas Muennighoff, Denis Kocetkov,
  Chenghao Mou, Marc Marone, Christopher Akiki, Jia Li, Jenny Chim, Qian Liu,
  Evgenii Zheltonozhskii, Terry~Yue Zhuo, Thomas Wang, Olivier Dehaene, Mishig
  Davaadorj, Joel Lamy-Poirier, João Monteiro, Oleh Shliazhko, Nicolas
  Gontier, Nicholas Meade, Armel Zebaze, Ming-Ho Yee, Logesh~Kumar Umapathi,
  Jian Zhu, Benjamin Lipkin, Muhtasham Oblokulov, Zhiruo Wang, Rudra Murthy,
  Jason Stillerman, Siva~Sankalp Patel, Dmitry Abulkhanov, Marco Zocca, Manan
  Dey, Zhihan Zhang, Nour Fahmy, Urvashi Bhattacharyya, Wenhao Yu, Swayam
  Singh, Sasha Luccioni, Paulo Villegas, Maxim Kunakov, Fedor Zhdanov, Manuel
  Romero, Tony Lee, Nadav Timor, Jennifer Ding, Claire Schlesinger, Hailey
  Schoelkopf, Jan Ebert, Tri Dao, Mayank Mishra, Alex Gu, Jennifer Robinson,
  Carolyn~Jane Anderson, Brendan Dolan-Gavitt, Danish Contractor, Siva Reddy,
  Daniel Fried, Dzmitry Bahdanau, Yacine Jernite, Carlos~Muñoz Ferrandis, Sean
  Hughes, Thomas Wolf, Arjun Guha, Leandro von Werra, and Harm de~Vries. 2023.
\newblock \href {http://arxiv.org/abs/2305.06161} {StarCoder: may the source be
  with you!}
\newblock \emph{arXiv preprint arXiv:2305.06161}.

\bibitem[{Liang et~al.(2022)Liang, Bommasani, Lee, Tsipras, Soylu, Yasunaga,
  Zhang, Narayanan, Wu, Kumar et~al.}]{liang2022holistic}
Percy Liang, Rishi Bommasani, Tony Lee, Dimitris Tsipras, Dilara Soylu,
  Michihiro Yasunaga, Yian Zhang, Deepak Narayanan, Yuhuai Wu, Ananya Kumar,
  et~al. 2022.
\newblock Holistic evaluation of language models.
\newblock \emph{arXiv preprint arXiv:2211.09110}.

\bibitem[{Lieber et~al.(2021)Lieber, Sharir, Lenz, and
  Shoham}]{lieber2021jurassic}
Opher Lieber, Or~Sharir, Barak Lenz, and Yoav Shoham. 2021.
\newblock Jurassic-1: Technical details and evaluation.
\newblock \emph{White Paper. AI21 Labs}, 1.

\bibitem[{Lin(2004)}]{lin2004rouge}
Chin-Yew Lin. 2004.
\newblock Rouge: A package for automatic evaluation of summaries.
\newblock In \emph{Text summarization branches out}, pages 74--81.

\bibitem[{Lin et~al.(2021)Lin, Mihaylov, Artetxe, Wang, Chen, Simig, Ott,
  Goyal, Bhosale, Du et~al.}]{lin2021few}
Xi~Victoria Lin, Todor Mihaylov, Mikel Artetxe, Tianlu Wang, Shuohui Chen,
  Daniel Simig, Myle Ott, Naman Goyal, Shruti Bhosale, Jingfei Du, et~al. 2021.
\newblock Few-shot learning with multilingual language models.
\newblock \emph{arXiv preprint arXiv:2112.10668}.

\bibitem[{Lin et~al.(2019)Lin, Chen, Lee, Li, Zhang, Xia, Rijhwani, He, Zhang,
  Ma et~al.}]{lin2019choosing}
Yu-Hsiang Lin, Chian-Yu Chen, Jean Lee, Zirui Li, Yuyan Zhang, Mengzhou Xia,
  Shruti Rijhwani, Junxian He, Zhisong Zhang, Xuezhe Ma, et~al. 2019.
\newblock Choosing transfer languages for cross-lingual learning.
\newblock \emph{arXiv preprint arXiv:1905.12688}.

\bibitem[{Longpre et~al.(2023{\natexlab{a}})Longpre, Hou, Vu, Webson, Chung,
  Tay, Zhou, Le, Zoph, Wei et~al.}]{longpre2023flan}
Shayne Longpre, Le~Hou, Tu~Vu, Albert Webson, Hyung~Won Chung, Yi~Tay, Denny
  Zhou, Quoc~V Le, Barret Zoph, Jason Wei, et~al. 2023{\natexlab{a}}.
\newblock The Flan Collection: Designing Data and Methods for Effective
  Instruction Tuning.
\newblock \emph{arXiv preprint arXiv:2301.13688}.

\bibitem[{Longpre et~al.(2023{\natexlab{b}})Longpre, Mahari, Chen, Obeng-Marnu,
  Sileo, Brannon, Muennighoff, Khazam, Kabbara, Perisetla, Wu, Shippole,
  Bollacker, Wu, Villa, Pentland, Roy, and Hooker}]{longpre2023data}
Shayne Longpre, Robert Mahari, Anthony Chen, Naana Obeng-Marnu, Damien Sileo,
  William Brannon, Niklas Muennighoff, Nathan Khazam, Jad Kabbara, Kartik
  Perisetla, Xinyi Wu, Enrico Shippole, Kurt Bollacker, Tongshuang Wu, Luis
  Villa, Sandy Pentland, Deb Roy, and Sara Hooker. 2023{\natexlab{b}}.
\newblock \href {http://arxiv.org/abs/2310.16787} {The Data Provenance
  Initiative: A Large Scale Audit of Dataset Licensing \& Attribution in AI}.

\bibitem[{Longpre et~al.(2023{\natexlab{c}})Longpre, Yauney, Reif, Lee,
  Roberts, Zoph, Zhou, Wei, Robinson, Mimno, and
  Ippolito}]{longpre2023pretrainers}
Shayne Longpre, Gregory Yauney, Emily Reif, Katherine Lee, Adam Roberts, Barret
  Zoph, Denny Zhou, Jason Wei, Kevin Robinson, David Mimno, and Daphne
  Ippolito. 2023{\natexlab{c}}.
\newblock \href {http://arxiv.org/abs/2305.13169} {A Pretrainer's Guide to
  Training Data: Measuring the Effects of Data Age, Domain Coverage, Quality,
  \& Toxicity}.

\bibitem[{Luukkonen et~al.(2023)Luukkonen, Komulainen, Luoma, Eskelinen,
  Kanerva, Kupari, Ginter, Laippala, Muennighoff, Piktus, Wang, Tazi, Scao,
  Wolf, Suominen, Sairanen, Merioksa, Heinonen, Vahtola, Antao, and
  Pyysalo}]{luukkonen2023fingpt}
Risto Luukkonen, Ville Komulainen, Jouni Luoma, Anni Eskelinen, Jenna Kanerva,
  Hanna-Mari Kupari, Filip Ginter, Veronika Laippala, Niklas Muennighoff,
  Aleksandra Piktus, Thomas Wang, Nouamane Tazi, Teven~Le Scao, Thomas Wolf,
  Osma Suominen, Samuli Sairanen, Mikko Merioksa, Jyrki Heinonen, Aija Vahtola,
  Samuel Antao, and Sampo Pyysalo. 2023.
\newblock Fin{GPT}: Large Generative Models for a Small Language.
\newblock In \emph{The 2023 Conference on Empirical Methods in Natural Language
  Processing}.

\bibitem[{Madani et~al.(2020)Madani, McCann, Naik, Keskar, Anand, Eguchi,
  Huang, and Socher}]{madani2020progen}
Ali Madani, Bryan McCann, Nikhil Naik, Nitish~Shirish Keskar, Namrata Anand,
  Raphael~R Eguchi, Po-Ssu Huang, and Richard Socher. 2020.
\newblock Progen: Language modeling for protein generation.
\newblock \emph{arXiv preprint arXiv:2004.03497}.

\bibitem[{McCann et~al.(2018)McCann, Keskar, Xiong, and
  Socher}]{DBLP:journals/corr/abs-1806-08730}
Bryan McCann, Nitish~Shirish Keskar, Caiming Xiong, and Richard Socher. 2018.
\newblock \href {http://arxiv.org/abs/1806.08730} {The Natural Language
  Decathlon: Multitask Learning as Question Answering}.
\newblock \emph{CoRR}, abs/1806.08730.

\bibitem[{Min et~al.(2021)Min, Lewis, Zettlemoyer, and
  Hajishirzi}]{min2021metaicl}
Sewon Min, Mike Lewis, Luke Zettlemoyer, and Hannaneh Hajishirzi. 2021.
\newblock Metaicl: Learning to learn in context.
\newblock \emph{arXiv preprint arXiv:2110.15943}.

\bibitem[{Mostafazadeh et~al.(2017)Mostafazadeh, Roth, Louis, Chambers, and
  Allen}]{mostafazadeh2017lsdsem}
Nasrin Mostafazadeh, Michael Roth, Annie Louis, Nathanael Chambers, and James
  Allen. 2017.
\newblock Lsdsem 2017 shared task: The story cloze test.
\newblock In \emph{Proceedings of the 2nd Workshop on Linking Models of
  Lexical, Sentential and Discourse-level Semantics}, pages 46--51.

\bibitem[{Muennighoff(2020)}]{muennighoff2020vilio}
Niklas Muennighoff. 2020.
\newblock Vilio: State-of-the-art visio-linguistic models applied to hateful
  memes.
\newblock \emph{arXiv preprint arXiv:2012.07788}.

\bibitem[{Muennighoff(2022)}]{muennighoff2022sgpt}
Niklas Muennighoff. 2022.
\newblock SGPT: GPT Sentence Embeddings for Semantic Search.
\newblock \emph{arXiv preprint arXiv:2202.08904}.

\bibitem[{Muennighoff et~al.(2023)Muennighoff, Liu, Zebaze, Zheng, Hui, Zhuo,
  Singh, Tang, von Werra, and Longpre}]{muennighoff2023octopack}
Niklas Muennighoff, Qian Liu, Armel Zebaze, Qinkai Zheng, Binyuan Hui,
  Terry~Yue Zhuo, Swayam Singh, Xiangru Tang, Leandro von Werra, and Shayne
  Longpre. 2023.
\newblock OctoPack: Instruction Tuning Code Large Language Models.
\newblock \emph{arXiv preprint arXiv:2308.07124}.

\bibitem[{Muennighoff et~al.(2022{\natexlab{a}})Muennighoff, Tazi, Magne, and
  Reimers}]{muennighoff2022mteb}
Niklas Muennighoff, Nouamane Tazi, Lo{\"\i}c Magne, and Nils Reimers.
  2022{\natexlab{a}}.
\newblock \href {https://doi.org/10.48550/ARXIV.2210.07316} {MTEB: Massive Text
  Embedding Benchmark}.
\newblock \emph{arXiv preprint arXiv:2210.07316}.

\bibitem[{Muennighoff et~al.(2022{\natexlab{b}})Muennighoff, Wang, Sutawika,
  Roberts, Biderman, Scao, Bari, Shen, Yong, Schoelkopf
  et~al.}]{muennighoff2022crosslingual}
Niklas Muennighoff, Thomas Wang, Lintang Sutawika, Adam Roberts, Stella
  Biderman, Teven~Le Scao, M~Saiful Bari, Sheng Shen, Zheng-Xin Yong, Hailey
  Schoelkopf, et~al. 2022{\natexlab{b}}.
\newblock Crosslingual generalization through multitask finetuning.
\newblock \emph{arXiv preprint arXiv:2211.01786}.

\bibitem[{Nakkiran et~al.(2021{\natexlab{a}})Nakkiran, Kaplun, Bansal, Yang,
  Barak, and Sutskever}]{nakkiran2021deep}
Preetum Nakkiran, Gal Kaplun, Yamini Bansal, Tristan Yang, Boaz Barak, and Ilya
  Sutskever. 2021{\natexlab{a}}.
\newblock Deep double descent: Where bigger models and more data hurt.
\newblock \emph{Journal of Statistical Mechanics: Theory and Experiment},
  2021(12):124003.

\bibitem[{Nakkiran et~al.(2021{\natexlab{b}})Nakkiran, Neyshabur, and
  Sedghi}]{NakkiranNS21}
Preetum Nakkiran, Behnam Neyshabur, and Hanie Sedghi. 2021{\natexlab{b}}.
\newblock \href {https://openreview.net/forum?id=guetrIHLFGI} {The Deep
  Bootstrap Framework: Good Online Learners are Good Offline Generalizers}.
\newblock In \emph{9th International Conference on Learning Representations,
  {ICLR} 2021, Virtual Event, Austria, May 3-7, 2021}. OpenReview.net.

\bibitem[{Narayan et~al.(2018)Narayan, Cohen, and Lapata}]{xsum-emnlp}
Shashi Narayan, Shay~B. Cohen, and Mirella Lapata. 2018.
\newblock Don't Give Me the Details, Just the Summary! {T}opic-Aware
  Convolutional Neural Networks for Extreme Summarization.
\newblock In \emph{Proceedings of the 2018 Conference on Empirical Methods in
  Natural Language Processing}, Brussels, Belgium.

\bibitem[{Narayanan et~al.(2021)Narayanan, Shoeybi, Casper, LeGresley, Patwary,
  Korthikanti, Vainbrand, Kashinkunti, Bernauer, Catanzaro
  et~al.}]{narayanan2021efficient}
Deepak Narayanan, Mohammad Shoeybi, Jared Casper, Patrick LeGresley, Mostofa
  Patwary, Vijay Korthikanti, Dmitri Vainbrand, Prethvi Kashinkunti, Julie
  Bernauer, Bryan Catanzaro, et~al. 2021.
\newblock Efficient large-scale language model training on gpu clusters using
  megatron-lm.
\newblock In \emph{Proceedings of the International Conference for High
  Performance Computing, Networking, Storage and Analysis}, pages 1--15.

\bibitem[{Nie et~al.(2020)Nie, Williams, Dinan, Bansal, Weston, and
  Kiela}]{nie2019adversarial}
Yixin Nie, Adina Williams, Emily Dinan, Mohit Bansal, Jason Weston, and Douwe
  Kiela. 2020.
\newblock Adversarial NLI: A New Benchmark for Natural Language Understanding.
\newblock In \emph{Proceedings of the 58th Annual Meeting of the Association
  for Computational Linguistics}. Association for Computational Linguistics.

\bibitem[{Nijkamp et~al.(2022)Nijkamp, Pang, Hayashi, Tu, Wang, Zhou, Savarese,
  and Xiong}]{nijkamp2022codegen}
Erik Nijkamp, Bo~Pang, Hiroaki Hayashi, Lifu Tu, Huan Wang, Yingbo Zhou, Silvio
  Savarese, and Caiming Xiong. 2022.
\newblock Codegen: An open large language model for code with multi-turn
  program synthesis.
\newblock \emph{arXiv preprint arXiv:2203.13474}.

\bibitem[{nostalgebraist(2022)}]{nostalgebraist}
nostalgebraist. 2022.
\newblock chinchilla’s wild implications.
\newblock \emph{lesswrong}.

\bibitem[{Orlanski et~al.(2023)Orlanski, Xiao, Garcia, Hui, Howland, Malmaud,
  Austin, Singh, and Catasta}]{orlanski2023measuring}
Gabriel Orlanski, Kefan Xiao, Xavier Garcia, Jeffrey Hui, Joshua Howland,
  Jonathan Malmaud, Jacob Austin, Rishah Singh, and Michele Catasta. 2023.
\newblock Measuring The Impact Of Programming Language Distribution.
\newblock \emph{arXiv preprint arXiv:2302.01973}.

\bibitem[{Ortiz~Su{'a}rez et~al.(2020)Ortiz~Su{'a}rez, Romary, and
  Sagot}]{ortiz-suarez-etal-2020-monolingual}
Pedro~Javier Ortiz~Su{'a}rez, Laurent Romary, and Benoit Sagot. 2020.
\newblock \href {https://www.aclweb.org/anthology/2020.acl-main.156} {A
  Monolingual Approach to Contextualized Word Embeddings for Mid-Resource
  Languages}.
\newblock In \emph{Proceedings of the 58th Annual Meeting of the Association
  for Computational Linguistics}, pages 1703--1714, Online. Association for
  Computational Linguistics.

\bibitem[{{Ortiz Su{'a}rez} et~al.(2019){Ortiz Su{'a}rez}, Sagot, and
  Romary}]{OrtizSuarezSagotRomary2019}
Pedro~Javier {Ortiz Su{'a}rez}, Benoit Sagot, and Laurent Romary. 2019.
\newblock \href {https://doi.org/10.14618/ids-pub-9021} {Asynchronous pipelines
  for processing huge corpora on medium to low resource infrastructures}.
\newblock Proceedings of the Workshop on Challenges in the Management of Large
  Corpora (CMLC-7) 2019. Cardiff, 22nd July 2019, pages 9 -- 16, Mannheim.
  Leibniz-Institut f{"u}r Deutsche Sprache.

\bibitem[{Ouyang et~al.(2022)Ouyang, Wu, Jiang, Almeida, Wainwright, Mishkin,
  Zhang, Agarwal, Slama, Ray et~al.}]{ouyang2022training}
Long Ouyang, Jeff Wu, Xu~Jiang, Diogo Almeida, Carroll~L Wainwright, Pamela
  Mishkin, Chong Zhang, Sandhini Agarwal, Katarina Slama, Alex Ray, et~al.
  2022.
\newblock Training language models to follow instructions with human feedback.
\newblock \emph{arXiv preprint arXiv:2203.02155}.

\bibitem[{Penedo et~al.(2023)Penedo, Malartic, Hesslow, Cojocaru, Cappelli,
  Alobeidli, Pannier, Almazrouei, and Launay}]{penedo2023refinedweb}
Guilherme Penedo, Quentin Malartic, Daniel Hesslow, Ruxandra Cojocaru,
  Alessandro Cappelli, Hamza Alobeidli, Baptiste Pannier, Ebtesam Almazrouei,
  and Julien Launay. 2023.
\newblock The RefinedWeb Dataset for Falcon LLM: Outperforming Curated Corpora
  with Web Data, and Web Data Only.
\newblock \emph{arXiv preprint arXiv:2306.01116}.

\bibitem[{Prabhumoye et~al.(2023)Prabhumoye, Patwary, Shoeybi, and
  Catanzaro}]{prabhumoye2023adding}
Shrimai Prabhumoye, Mostofa Patwary, Mohammad Shoeybi, and Bryan Catanzaro.
  2023.
\newblock Adding Instructions during Pretraining: Effective Way of Controlling
  Toxicity in Language Models.
\newblock \emph{arXiv preprint arXiv:2302.07388}.

\bibitem[{Radford et~al.(2019)Radford, Wu, Child, Luan, Amodei, Sutskever
  et~al.}]{radford2019language}
Alec Radford, Jeffrey Wu, Rewon Child, David Luan, Dario Amodei, Ilya
  Sutskever, et~al. 2019.
\newblock Language models are unsupervised multitask learners.
\newblock \emph{OpenAI blog}, 1(8):9.

\bibitem[{Rae et~al.(2021)Rae, Borgeaud, Cai, Millican, Hoffmann, Song,
  Aslanides, Henderson, Ring, Young et~al.}]{rae2021scaling}
Jack~W Rae, Sebastian Borgeaud, Trevor Cai, Katie Millican, Jordan Hoffmann,
  Francis Song, John Aslanides, Sarah Henderson, Roman Ring, Susannah Young,
  et~al. 2021.
\newblock Scaling language models: Methods, analysis \& insights from training
  gopher.
\newblock \emph{arXiv preprint arXiv:2112.11446}.

\bibitem[{Raffel et~al.(2020)Raffel, Shazeer, Roberts, Lee, Narang, Matena,
  Zhou, Li, Liu et~al.}]{raffel2020exploring}
Colin Raffel, Noam Shazeer, Adam Roberts, Katherine Lee, Sharan Narang, Michael
  Matena, Yanqi Zhou, Wei Li, Peter~J Liu, et~al. 2020.
\newblock Exploring the limits of transfer learning with a unified text-to-text
  transformer.
\newblock \emph{J. Mach. Learn. Res.}, 21(140):1--67.

\bibitem[{Rasley et~al.(2020)Rasley, Rajbhandari, Ruwase, and
  He}]{rasley2020deepspeed}
Jeff Rasley, Samyam Rajbhandari, Olatunji Ruwase, and Yuxiong He. 2020.
\newblock Deepspeed: System optimizations enable training deep learning models
  with over 100 billion parameters.
\newblock In \emph{Proceedings of the 26th ACM SIGKDD International Conference
  on Knowledge Discovery \& Data Mining}, pages 3505--3506.

\bibitem[{Roemmele et~al.(2011)Roemmele, Bejan, and
  Gordon}]{roemmele2011choice}
Melissa Roemmele, Cosmin~Adrian Bejan, and Andrew~S Gordon. 2011.
\newblock Choice of Plausible Alternatives: An Evaluation of Commonsense Causal
  Reasoning.
\newblock In \emph{AAAI spring symposium: logical formalizations of commonsense
  reasoning}, pages 90--95.

\bibitem[{Sakaguchi et~al.(2021)Sakaguchi, Bras, Bhagavatula, and
  Choi}]{sakaguchi2021winogrande}
Keisuke Sakaguchi, Ronan~Le Bras, Chandra Bhagavatula, and Yejin Choi. 2021.
\newblock Winogrande: An adversarial winograd schema challenge at scale.
\newblock \emph{Communications of the ACM}, 64(9):99--106.

\bibitem[{Sanh et~al.(2022)Sanh, Webson, Raffel, Bach, Sutawika, Alyafeai,
  Chaffin, Stiegler, Le~Scao, Raja et~al.}]{sanh2022multitask}
Victor Sanh, Albert Webson, Colin Raffel, Stephen Bach, Lintang Sutawika, Zaid
  Alyafeai, Antoine Chaffin, Arnaud Stiegler, Teven Le~Scao, Arun Raja, et~al.
  2022.
\newblock Multitask Prompted Training Enables Zero-Shot Task Generalization.
\newblock In \emph{The Tenth International Conference on Learning
  Representations}.

\bibitem[{Scao et~al.(2022{\natexlab{a}})Scao, Fan, Akiki, Pavlick, Ili{\'c},
  Hesslow, Castagn{\'e}, Luccioni, Yvon, Gall{\'e} et~al.}]{scao2022bloom}
Teven~Le Scao, Angela Fan, Christopher Akiki, Ellie Pavlick, Suzana Ili{\'c},
  Daniel Hesslow, Roman Castagn{\'e}, Alexandra~Sasha Luccioni, Fran{\c{c}}ois
  Yvon, Matthias Gall{\'e}, et~al. 2022{\natexlab{a}}.
\newblock Bloom: A 176b-parameter open-access multilingual language model.
\newblock \emph{arXiv preprint arXiv:2211.05100}.

\bibitem[{Scao et~al.(2022{\natexlab{b}})Scao, Wang, Hesslow, Saulnier, Bekman,
  Bari, Bideman, Elsahar, Muennighoff, Phang et~al.}]{scao2022language}
Teven~Le Scao, Thomas Wang, Daniel Hesslow, Lucile Saulnier, Stas Bekman,
  M~Saiful Bari, Stella Bideman, Hady Elsahar, Niklas Muennighoff, Jason Phang,
  et~al. 2022{\natexlab{b}}.
\newblock What Language Model to Train if You Have One Million GPU Hours?
\newblock \emph{arXiv preprint arXiv:2210.15424}.

\bibitem[{Shin et~al.(2022)Shin, Lee, Ahn, Kim, Kim, Kim, Cho, Lee, Park, Ha
  et~al.}]{shin2022effect}
Seongjin Shin, Sang-Woo Lee, Hwijeen Ahn, Sungdong Kim, HyoungSeok Kim, Boseop
  Kim, Kyunghyun Cho, Gichang Lee, Woomyoung Park, Jung-Woo Ha, et~al. 2022.
\newblock On the effect of pretraining corpora on in-context learning by a
  large-scale language model.
\newblock \emph{arXiv preprint arXiv:2204.13509}.

\bibitem[{Shoeybi et~al.(2019)Shoeybi, Patwary, Puri, LeGresley, Casper, and
  Catanzaro}]{shoeybi2019megatron}
Mohammad Shoeybi, Mostofa Patwary, Raul Puri, Patrick LeGresley, Jared Casper,
  and Bryan Catanzaro. 2019.
\newblock Megatron-lm: Training multi-billion parameter language models using
  model parallelism.
\newblock \emph{arXiv preprint arXiv:1909.08053}.

\bibitem[{Smith et~al.(2022)Smith, Patwary, Norick, LeGresley, Rajbhandari,
  Casper, Liu, Prabhumoye, Zerveas, Korthikanti et~al.}]{smith2022using}
Shaden Smith, Mostofa Patwary, Brandon Norick, Patrick LeGresley, Samyam
  Rajbhandari, Jared Casper, Zhun Liu, Shrimai Prabhumoye, George Zerveas,
  Vijay Korthikanti, et~al. 2022.
\newblock Using deepspeed and megatron to train megatron-turing nlg 530b, a
  large-scale generative language model.
\newblock \emph{arXiv preprint arXiv:2201.11990}.

\bibitem[{Soltan et~al.(2022)Soltan, Ananthakrishnan, FitzGerald, Gupta, Hamza,
  Khan, Peris, Rawls, Rosenbaum, Rumshisky et~al.}]{soltan2022alexatm}
Saleh Soltan, Shankar Ananthakrishnan, Jack FitzGerald, Rahul Gupta, Wael
  Hamza, Haidar Khan, Charith Peris, Stephen Rawls, Andy Rosenbaum, Anna
  Rumshisky, et~al. 2022.
\newblock Alexatm 20b: Few-shot learning using a large-scale multilingual
  seq2seq model.
\newblock \emph{arXiv preprint arXiv:2208.01448}.

\bibitem[{Sorscher et~al.(2022)Sorscher, Geirhos, Shekhar, Ganguli, and
  Morcos}]{sorscher2022beyond}
Ben Sorscher, Robert Geirhos, Shashank Shekhar, Surya Ganguli, and Ari~S
  Morcos. 2022.
\newblock Beyond neural scaling laws: beating power law scaling via data
  pruning.
\newblock \emph{arXiv preprint arXiv:2206.14486}.

\bibitem[{Srivastava et~al.(2022)Srivastava, Rastogi, Rao, Shoeb, Abid, Fisch,
  Brown, Santoro, Gupta, Garriga-Alonso et~al.}]{srivastava2022beyond}
Aarohi Srivastava, Abhinav Rastogi, Abhishek Rao, Abu Awal~Md Shoeb, Abubakar
  Abid, Adam Fisch, Adam~R Brown, Adam Santoro, Aditya Gupta, Adri{\`a}
  Garriga-Alonso, et~al. 2022.
\newblock Beyond the imitation game: Quantifying and extrapolating the
  capabilities of language models.
\newblock \emph{arXiv preprint arXiv:2206.04615}.

\bibitem[{Su et~al.(2022)Su, Zhou, Yu, Chen, Zhu, Yu, and Zhou}]{su2022welm}
Hui Su, Xiao Zhou, Houjing Yu, Yuwen Chen, Zilin Zhu, Yang Yu, and Jie Zhou.
  2022.
\newblock WeLM: A Well-Read Pre-trained Language Model for Chinese.
\newblock \emph{arXiv preprint arXiv:2209.10372}.

\bibitem[{Tay et~al.(2022{\natexlab{a}})Tay, Dehghani, Abnar, Chung, Fedus,
  Rao, Narang, Tran, Yogatama, and Metzler}]{tay2022scaling}
Yi~Tay, Mostafa Dehghani, Samira Abnar, Hyung~Won Chung, William Fedus, Jinfeng
  Rao, Sharan Narang, Vinh~Q Tran, Dani Yogatama, and Donald Metzler.
  2022{\natexlab{a}}.
\newblock Scaling Laws vs Model Architectures: How does Inductive Bias
  Influence Scaling?
\newblock \emph{arXiv preprint arXiv:2207.10551}.

\bibitem[{Tay et~al.(2021)Tay, Dehghani, Rao, Fedus, Abnar, Chung, Narang,
  Yogatama, Vaswani, and Metzler}]{tay2021scale}
Yi~Tay, Mostafa Dehghani, Jinfeng Rao, William Fedus, Samira Abnar, Hyung~Won
  Chung, Sharan Narang, Dani Yogatama, Ashish Vaswani, and Donald Metzler.
  2021.
\newblock Scale efficiently: Insights from pre-training and fine-tuning
  transformers.
\newblock \emph{arXiv preprint arXiv:2109.10686}.

\bibitem[{Tay et~al.(2022{\natexlab{b}})Tay, Dehghani, Tran, Garcia, Bahri,
  Schuster, Zheng, Houlsby, and Metzler}]{tay2022unifying}
Yi~Tay, Mostafa Dehghani, Vinh~Q Tran, Xavier Garcia, Dara Bahri, Tal Schuster,
  Huaixiu~Steven Zheng, Neil Houlsby, and Donald Metzler. 2022{\natexlab{b}}.
\newblock Unifying Language Learning Paradigms.
\newblock \emph{arXiv preprint arXiv:2205.05131}.

\bibitem[{Tay et~al.(2022{\natexlab{c}})Tay, Wei, Chung, Tran, So, Shakeri,
  Garcia, Zheng, Rao, Chowdhery et~al.}]{tay2022transcending}
Yi~Tay, Jason Wei, Hyung~Won Chung, Vinh~Q Tran, David~R So, Siamak Shakeri,
  Xavier Garcia, Huaixiu~Steven Zheng, Jinfeng Rao, Aakanksha Chowdhery, et~al.
  2022{\natexlab{c}}.
\newblock Transcending scaling laws with 0.1\% extra compute.
\newblock \emph{arXiv preprint arXiv:2210.11399}.

\bibitem[{Taylor et~al.(2022)Taylor, Kardas, Cucurull, Scialom, Hartshorn,
  Saravia, Poulton, Kerkez, and Stojnic}]{taylor2022galactica}
Ross Taylor, Marcin Kardas, Guillem Cucurull, Thomas Scialom, Anthony
  Hartshorn, Elvis Saravia, Andrew Poulton, Viktor Kerkez, and Robert Stojnic.
  2022.
\newblock Galactica: A large language model for science.
\newblock \emph{arXiv preprint arXiv:2211.09085}.

\bibitem[{Thoppilan et~al.(2022)Thoppilan, De~Freitas, Hall, Shazeer,
  Kulshreshtha, Cheng, Jin, Bos, Baker, Du et~al.}]{thoppilan2022lamda}
Romal Thoppilan, Daniel De~Freitas, Jamie Hall, Noam Shazeer, Apoorv
  Kulshreshtha, Heng-Tze Cheng, Alicia Jin, Taylor Bos, Leslie Baker, Yu~Du,
  et~al. 2022.
\newblock Lamda: Language models for dialog applications.
\newblock \emph{arXiv preprint arXiv:2201.08239}.

\bibitem[{Touvron et~al.(2023)Touvron, Lavril, Izacard, Martinet, Lachaux,
  Lacroix, Rozi{\`e}re, Goyal, Hambro, Azhar et~al.}]{touvron2023llama}
Hugo Touvron, Thibaut Lavril, Gautier Izacard, Xavier Martinet, Marie-Anne
  Lachaux, Timoth{\'e}e Lacroix, Baptiste Rozi{\`e}re, Naman Goyal, Eric
  Hambro, Faisal Azhar, et~al. 2023.
\newblock Llama: Open and efficient foundation language models.
\newblock \emph{arXiv preprint arXiv:2302.13971}.

\bibitem[{Vaswani et~al.(2017)Vaswani, Shazeer, Parmar, Uszkoreit, Jones,
  Gomez, Kaiser, and Polosukhin}]{vaswani2017attention}
Ashish Vaswani, Noam Shazeer, Niki Parmar, Jakob Uszkoreit, Llion Jones,
  Aidan~N Gomez, {\L}ukasz Kaiser, and Illia Polosukhin. 2017.
\newblock Attention is all you need.
\newblock \emph{Advances in neural information processing systems}, 30.

\bibitem[{Villalobos et~al.(2022)Villalobos, Sevilla, Heim, Besiroglu,
  Hobbhahn, and Ho}]{villalobos2022will}
Pablo Villalobos, Jaime Sevilla, Lennart Heim, Tamay Besiroglu, Marius
  Hobbhahn, and Anson Ho. 2022.
\newblock Will we run out of data? An analysis of the limits of scaling
  datasets in Machine Learning.
\newblock \emph{arXiv preprint arXiv:2211.04325}.

\bibitem[{Virtanen et~al.(2019)Virtanen, Kanerva, Ilo, Luoma, Luotolahti,
  Salakoski, Ginter, and Pyysalo}]{virtanen2019multilingual}
Antti Virtanen, Jenna Kanerva, Rami Ilo, Jouni Luoma, Juhani Luotolahti, Tapio
  Salakoski, Filip Ginter, and Sampo Pyysalo. 2019.
\newblock Multilingual is not enough: BERT for Finnish.
\newblock \emph{arXiv preprint arXiv:1912.07076}.

\bibitem[{Wang et~al.(2019)Wang, Pruksachatkun, Nangia, Singh, Michael, Hill,
  Levy, and Bowman}]{wang2019superglue}
Alex Wang, Yada Pruksachatkun, Nikita Nangia, Amanpreet Singh, Julian Michael,
  Felix Hill, Omer Levy, and Samuel~R. Bowman. 2019.
\newblock \href {http://arxiv.org/abs/1905.00537} {SuperGLUE: {A} Stickier
  Benchmark for General-Purpose Language Understanding Systems}.
\newblock \emph{CoRR}, abs/1905.00537.

\bibitem[{Wang et~al.(2023)Wang, Ivison, Dasigi, Hessel, Khot, Chandu, Wadden,
  MacMillan, Smith, Beltagy et~al.}]{wang2023far}
Yizhong Wang, Hamish Ivison, Pradeep Dasigi, Jack Hessel, Tushar Khot,
  Khyathi~Raghavi Chandu, David Wadden, Kelsey MacMillan, Noah~A Smith,
  Iz~Beltagy, et~al. 2023.
\newblock How Far Can Camels Go? Exploring the State of Instruction Tuning on
  Open Resources.
\newblock \emph{arXiv preprint arXiv:2306.04751}.

\bibitem[{Wang et~al.(2022)Wang, Kordi, Mishra, Liu, Smith, Khashabi, and
  Hajishirzi}]{wang2022self}
Yizhong Wang, Yeganeh Kordi, Swaroop Mishra, Alisa Liu, Noah~A Smith, Daniel
  Khashabi, and Hannaneh Hajishirzi. 2022.
\newblock Self-Instruct: Aligning Language Model with Self Generated
  Instructions.
\newblock \emph{arXiv preprint arXiv:2212.10560}.

\bibitem[{Wei et~al.(2021)Wei, Bosma, Zhao, Guu, Yu, Lester, Du, Dai, and
  Le}]{wei2021finetuned}
Jason Wei, Maarten Bosma, Vincent~Y Zhao, Kelvin Guu, Adams~Wei Yu, Brian
  Lester, Nan Du, Andrew~M Dai, and Quoc~V Le. 2021.
\newblock Finetuned language models are zero-shot learners.
\newblock \emph{arXiv preprint arXiv:2109.01652}.

\bibitem[{Wei et~al.(2022)Wei, Wang, Schuurmans, Bosma, Chi, Le, and
  Zhou}]{wei2022chain}
Jason Wei, Xuezhi Wang, Dale Schuurmans, Maarten Bosma, Ed~Chi, Quoc Le, and
  Denny Zhou. 2022.
\newblock Chain of thought prompting elicits reasoning in large language
  models.
\newblock \emph{arXiv preprint arXiv:2201.11903}.

\bibitem[{Weidinger et~al.(2021)Weidinger, Mellor, Rauh, Griffin, Uesato,
  Huang, Cheng, Glaese, Balle, Kasirzadeh et~al.}]{weidinger2021ethical}
Laura Weidinger, John Mellor, Maribeth Rauh, Conor Griffin, Jonathan Uesato,
  Po-Sen Huang, Myra Cheng, Mia Glaese, Borja Balle, Atoosa Kasirzadeh, et~al.
  2021.
\newblock Ethical and social risks of harm from language models.
\newblock \emph{arXiv preprint arXiv:2112.04359}.

\bibitem[{Welbl et~al.(2017)Welbl, Liu, and Gardner}]{welbl2017crowdsourcing}
Johannes Welbl, Nelson~F Liu, and Matt Gardner. 2017.
\newblock Crowdsourcing multiple choice science questions.
\newblock \emph{arXiv preprint arXiv:1707.06209}.

\bibitem[{Wenzek et~al.(2019)Wenzek, Lachaux, Conneau, Chaudhary, Guzm{\'a}n,
  Joulin, and Grave}]{wenzek2019ccnet}
Guillaume Wenzek, Marie-Anne Lachaux, Alexis Conneau, Vishrav Chaudhary,
  Francisco Guzm{\'a}n, Armand Joulin, and Edouard Grave. 2019.
\newblock CCNet: Extracting high quality monolingual datasets from web crawl
  data.
\newblock \emph{arXiv preprint arXiv:1911.00359}.

\bibitem[{Weston et~al.(2015)Weston, Bordes, Chopra, Rush, Van~Merri{\"e}nboer,
  Joulin, and Mikolov}]{weston2015towards}
Jason Weston, Antoine Bordes, Sumit Chopra, Alexander~M Rush, Bart
  Van~Merri{\"e}nboer, Armand Joulin, and Tomas Mikolov. 2015.
\newblock Towards ai-complete question answering: A set of prerequisite toy
  tasks.
\newblock \emph{arXiv preprint arXiv:1502.05698}.

\bibitem[{Xia et~al.(2022)Xia, Artetxe, Zhou, Lin, Pasunuru, Chen, Zettlemoyer,
  and Stoyanov}]{xia2022training}
Mengzhou Xia, Mikel Artetxe, Chunting Zhou, Xi~Victoria Lin, Ramakanth
  Pasunuru, Danqi Chen, Luke Zettlemoyer, and Ves Stoyanov. 2022.
\newblock Training Trajectories of Language Models Across Scales.
\newblock \emph{arXiv preprint arXiv:2212.09803}.

\bibitem[{Xia et~al.(2021)Xia, Zheng, Mukherjee, Shokouhi, Neubig, and
  Awadallah}]{xia2021metaxl}
Mengzhou Xia, Guoqing Zheng, Subhabrata Mukherjee, Milad Shokouhi, Graham
  Neubig, and Ahmed~Hassan Awadallah. 2021.
\newblock MetaXL: Meta representation transformation for low-resource
  cross-lingual learning.
\newblock \emph{arXiv preprint arXiv:2104.07908}.

\bibitem[{Xu et~al.(2023)Xu, Sun, Zheng, Geng, Zhao, Feng, Tao, and
  Jiang}]{xu2023wizardlm}
Can Xu, Qingfeng Sun, Kai Zheng, Xiubo Geng, Pu~Zhao, Jiazhan Feng, Chongyang
  Tao, and Daxin Jiang. 2023.
\newblock Wizardlm: Empowering large language models to follow complex
  instructions.
\newblock \emph{arXiv preprint arXiv:2304.12244}.

\bibitem[{Xue et~al.(2020)Xue, Constant, Roberts, Kale, Al-Rfou, Siddhant,
  Barua, and Raffel}]{xue2020mt5}
Linting Xue, Noah Constant, Adam Roberts, Mihir Kale, Rami Al-Rfou, Aditya
  Siddhant, Aditya Barua, and Colin Raffel. 2020.
\newblock mT5: A massively multilingual pre-trained text-to-text transformer.
\newblock \emph{arXiv preprint arXiv:2010.11934}.

\bibitem[{Yang et~al.(2021)Yang, Hu, Babuschkin, Sidor, Liu, Farhi, Ryder,
  Pachocki, Chen, and Gao}]{yang2021tuning}
Ge~Yang, Edward Hu, Igor Babuschkin, Szymon Sidor, Xiaodong Liu, David Farhi,
  Nick Ryder, Jakub Pachocki, Weizhu Chen, and Jianfeng Gao. 2021.
\newblock Tuning large neural networks via zero-shot hyperparameter transfer.
\newblock \emph{Advances in Neural Information Processing Systems},
  34:17084--17097.

\bibitem[{Yong et~al.(2022)Yong, Schoelkopf, Muennighoff, Aji, Adelani,
  Almubarak, Bari, Sutawika, Kasai, Baruwa et~al.}]{yong2022bloom+}
Zheng-Xin Yong, Hailey Schoelkopf, Niklas Muennighoff, Alham~Fikri Aji,
  David~Ifeoluwa Adelani, Khalid Almubarak, M~Saiful Bari, Lintang Sutawika,
  Jungo Kasai, Ahmed Baruwa, et~al. 2022.
\newblock BLOOM+ 1: Adding Language Support to BLOOM for Zero-Shot Prompting.
\newblock \emph{arXiv preprint arXiv:2212.09535}.

\bibitem[{Zellers et~al.(2019)Zellers, Holtzman, Bisk, Farhadi, and
  Choi}]{zellers2019hellaswag}
Rowan Zellers, Ari Holtzman, Yonatan Bisk, Ali Farhadi, and Yejin Choi. 2019.
\newblock HellaSwag: Can a Machine Really Finish Your Sentence?
\newblock In \emph{Proceedings of the 57th Annual Meeting of the Association
  for Computational Linguistics}.

\bibitem[{Zeng et~al.(2022)Zeng, Liu, Du, Wang, Lai, Ding, Yang, Xu, Zheng, Xia
  et~al.}]{zeng2022glm}
Aohan Zeng, Xiao Liu, Zhengxiao Du, Zihan Wang, Hanyu Lai, Ming Ding, Zhuoyi
  Yang, Yifan Xu, Wendi Zheng, Xiao Xia, et~al. 2022.
\newblock GLM-130B: An Open Bilingual Pre-trained Model.
\newblock \emph{arXiv preprint arXiv:2210.02414}.

\bibitem[{Zeng et~al.(2021)Zeng, Ren, Su, Wang, Liao, Wang, Jiang, Yang, Wang,
  Zhang et~al.}]{zeng2021pangu}
Wei Zeng, Xiaozhe Ren, Teng Su, Hui Wang, Yi~Liao, Zhiwei Wang, Xin Jiang,
  ZhenZhang Yang, Kaisheng Wang, Xiaoda Zhang, et~al. 2021.
\newblock PanGu-alpha: Large-scale Autoregressive Pretrained Chinese Language
  Models with Auto-parallel Computation.
\newblock \emph{arXiv preprint arXiv:2104.12369}.

\bibitem[{Zhang et~al.(2022)Zhang, Roller, Goyal, Artetxe, Chen, Chen, Dewan,
  Diab, Li, Lin et~al.}]{zhang2022opt}
Susan Zhang, Stephen Roller, Naman Goyal, Mikel Artetxe, Moya Chen, Shuohui
  Chen, Christopher Dewan, Mona Diab, Xian Li, Xi~Victoria Lin, et~al. 2022.
\newblock Opt: Open pre-trained transformer language models.
\newblock \emph{arXiv preprint arXiv:2205.01068}.

\bibitem[{Zhao et~al.(2023)Zhao, Zhou, Li, Tang, Wang, Hou, Min, Zhang, Zhang,
  Dong et~al.}]{zhao2023survey}
Wayne~Xin Zhao, Kun Zhou, Junyi Li, Tianyi Tang, Xiaolei Wang, Yupeng Hou,
  Yingqian Min, Beichen Zhang, Junjie Zhang, Zican Dong, et~al. 2023.
\newblock A survey of large language models.
\newblock \emph{arXiv preprint arXiv:2303.18223}.

\bibitem[{Zhou et~al.(2023)Zhou, Liu, Xu, Iyer, Sun, Mao, Ma, Efrat, Yu, Yu
  et~al.}]{zhou2023lima}
Chunting Zhou, Pengfei Liu, Puxin Xu, Srini Iyer, Jiao Sun, Yuning Mao, Xuezhe
  Ma, Avia Efrat, Ping Yu, Lili Yu, et~al. 2023.
\newblock Lima: Less is more for alignment.
\newblock \emph{arXiv preprint arXiv:2305.11206}.

\bibitem[{Zoph et~al.(2022)Zoph, Bello, Kumar, Du, Huang, Dean, Shazeer, and
  Fedus}]{zoph2022designing}
Barret Zoph, Irwan Bello, Sameer Kumar, Nan Du, Yanping Huang, Jeff Dean, Noam
  Shazeer, and William Fedus. 2022.
\newblock Designing effective sparse expert models.
\newblock \emph{arXiv preprint arXiv:2202.08906}.

\end{thebibliography}
\bibliographystyle{acl_natbib}

\newpage
\appendix


\part{}
\section*{\centering \LARGE{Appendix}}
\mtcsettitle{parttoc}{Contents}
\parttoc

\clearpage

\section{Derivation of Data-Constrained Scaling Laws}
\label{sec:scalinglaws}

Let $N$ be the number of model parameters, $D$ be the training tokens and $U$ be the "unique" training tokens i.e. the size of the dataset that is to be trained on for one or more epochs. Chinchilla~\cite{hoffmann2022training} only deals with non-repeated tokens, thus $D=U$ and we can write their formula  (``Approach 3'') as:

\begin{equation}\label{eq:cc}
L(N,U) = \tfrac{A}{N^\alpha} + \tfrac{B}{U^\beta} + E
\end{equation}

where $E$ represents the irreducible loss. $A$, $B$, $\alpha$ and $\beta$ are learned parameters.


We now want to generalize this expression to multiple epochs where tokens are repeated. We repeat the data $R_D$ times, where $R_D=0$ corresponds to the base case of a single epoch. We let $D'$ be the ``effective data size'': the number of unique data needed to get the same value as repeating $U$ unique tokens for $R_D$ repeats. Hence, if $R_D=0$, the effective data is the same as the total data processed. Intuitively, each time a sample is repeated, it is worth less as the model has already learned some of its information. Assume that each time a model trains on a token, it learns a $1-\delta$ fraction of the information in it for some constant $0 \leq \delta \leq 1$. (Thus,
if $\delta=0$ repeated tokens are as good as new ones, and if $\delta=1$, repeated tokens are worth nothing.) In other words, we expect the decrease in value of each repetition to be proportional to the value of the prior repetition, which is equivalent to exponential decay. As we would like to sum up the value of all repetitions, we temporarily assume an integral number of repeats and express it as a geometric series:

\begin{equation}\label{eq:ccu}
D' = U + (1-\delta)U + (1-\delta)^{2}U + \cdots + (1-\delta)^{R_D}U
\end{equation}

We know that the sum $S$ of a geometric series with a common ratio $r$ is:

\begin{equation}\label{eq:geos}
S = \frac{a(1-r^n)}{1-r}
\end{equation}

where $a$ is the first term and $n$ the number of terms in the series. As $r=(1-\delta)$ and $a=(1-\delta)U$:


\begin{equation}\label{eq:ar}
D' = U + U\sum_{k=1}^{R_D} (1-\delta)^k = U + (1-\delta)U\tfrac{(1-(1-\delta)^{R_D})}{\delta}
\end{equation}

Note that \autoref{eq:ar} can also be used with a non-integer number of repetitions. We can directly use \autoref{eq:ar} as our effective data and learn $\delta$ but for convenience and interpretability, we redefine it in terms of the number of epochs beyond which repeating does not help. Note that as more data is repeated, the right-hand side tends to $\tfrac{(1-\delta)U}{\delta}$, as $\lim_{R_D\to\infty}(1 - (1-\delta)^{R_D}) = 1$. Let $R^*_D = \tfrac{1-\delta}{\delta}$, hence $D'$ ``plateaus'' at $U + R^*_D U$ as $R_D$ goes to infinity.


If we assume $\delta$ to be small, ${1-\delta}$ tends to one and we can approximate $1/R^*_D = \tfrac{\delta}{1-\delta} \approx \delta$.


Next, define $e^x$ in terms of its Taylor series expansion:

\begin{equation}\label{eq:approx2}
e^x = 1 + x + \frac{x^2}{2!} + \frac{x^3}{3!} + \dots \approx 1 + x
\end{equation}

If $x$ is small later terms become increasingly small, thus $e^x \approx 1 + x$. As we have assumed $\delta$ to be small, let $x = -\delta$, which yields 

\begin{equation}\label{eq:approx3}
(1 + x) = (1 - \delta) \approx e^{-\delta} \approx e^{-1/R^*_D}
\end{equation}

Now inserting $(1-\delta)/\delta=R_D^*$ and $(1-\delta)^{R_D}= e^{(-1/R^*_D)^{R_D}}$ into \autoref{eq:ar} we get our final equation representing the \emph{effective data}:

\begin{equation}\label{eq:dterm}
D' = U + U\cdot R_D^* \cdot(1- e^{-R_D/R_D^*})
\end{equation}

where $U$ and $R_D$ are given while $R_D^*$ is a learned constant. If no repeats are done, the second part of the sum is zero and the term simplifies to the single-epoch scaling laws from \autoref{eq:cc}. While $R_D \ll R_D^*$, the second term is approximated as $U\cdot R_D$ and for $R_D \gg R_D^*$, it plateaus at $U\cdot R_D^*$. Hence $R^*_D$ corresponds to the number of times we can repeat tokens before seeing sharply diminishing returns.


Let us consider a concrete example to show that \autoref{eq:dterm} is a very good approximation of \autoref{eq:ar} and make the equations more intuitive. Suppose repeated data retains 75\% of its value ($\delta=0.25$) and we train on a single token or data unit ($U=1$) for five epochs, i.e. we repeat it four times ($R_D=4$). In that case \autoref{eq:ar} yields $D'=U + (1-\delta)U\tfrac{(1-(1-\delta)^{R_D})}{\delta}=1+(0.75)*4*(1-0.75^4)=3.05$. Thus despite training for 5 total units (4 of which are repetitions), we only get the value equivalent to $3.05$ units. As we have defined $R_D^*=(1 - \delta)/\delta$, the corresponding $R_D^*$ value is $3$. Setting $R_D^*=3$ in \autoref{eq:dterm} yields $D'=U + U\cdot R_D^* \cdot(1- e^{-R_D/R_D^*})=1 + 3 * (1 - e^{-4/3})=3.21$. Due to our approximations, the results are not the same, i.e. $3.21$ is slightly higher than $3.05$. However, note that the data term is additionally raised to a power of $\beta=0.353$ (see \autoref{eq:cc}; \autoref{sec:c4scaling}), thus the actual difference calculated as $((3.21^{0.353}) / (3.05^{0.353})) - 1$ is a mere 1.8\% despite this relatively large $\delta$ of $0.25$. \autoref{eq:dterm} has the benefit that we can interpret $R_D^*$ as the number of repetitions beyond which repeating yields sharply diminishing returns and flattens out soon after. Consider $R_D=100$ then $D'=1 + 3 * (1 - e^{-100/3})=3.99$. No matter how many repeats are done the effective data will never exceed $4$ i.e. it plateaus at $U + R_D^*U$ as $R_D$ tends to infinity.

Similarly, we consider repeating parameters. Symmetric to seeing the same data, excess parameters learn the same features and do not add any value in the extreme. For the Chinchilla equation (\autoref{eq:cc}) increasing parameters from 1 billion to 10 billion yields the same absolute decrease in loss regardless of whether the dataset is a single token or 1 billion tokens. However, intuition and our data (\autoref{sec:toomany}) suggest that in the first case, adding parameters should not decrease loss at all, as the additional 9 billion parameters cannot possibly learn anything from the single token that the first 1 billion parameters have not already learned. Thus, to allow excess parameters to decay to adding nothing, we also replace $N$ with a symmetric version of \autoref{eq:dterm} yielding our final equation:

\begin{equation}\label{eq:fin}
\begin{aligned}
L(U_N,U_D,R_N,R_D)=\frac{A}{(U_N + U_N R_N^* (1 - e^{\frac{-R_N}{R_N^*}}))^\alpha} + \frac{B}{(U_D + U_D R_D^* (1 - e^{\frac{-R_D}{R_D^*}}))^\beta} + E
\end{aligned}
\end{equation}

We define $U_N$, as the number of "unique" parameters that provide an optimal fit for $U_D$. Additional parameters decay with a symmetric version of the expression for repeated data. $R_N$ is the number that the "unique" parameters are repeated i.e. $R_N = \max\{(N / U_N) - 1, 0\}$. If $R_N^*=\infty$, additional parameters do not decay at all and $(U_N + U_N R_N^* (1 - e^{\frac{-R_N}{R_N^*}}))$ reduces to $N$. We compute $U_N$ from $U_D$ by setting $D_{opt}=U_D$ and rearranging \autoref{eq:ccopt} to map from $D_{opt}$ to $N_{opt}$. $U_N$ is then $\min\{N_{opt}, N\}$. This is equivalent to the following:

\begin{equation}\label{eq:optunud}
\begin{aligned}
U_N = \min\{ ((U_D \cdot G)^{\beta/\alpha}) \cdot G, N \}
\quad \text{ where } \quad G = {\left(\frac{\alpha A}{\beta B} \right)}^{\frac{1}{\alpha + \beta}}
\end{aligned}
\end{equation}


\autoref{eq:fin} is a generalization of \autoref{eq:cc}: It provides the same estimates for optimal model and data size in the single epoch case, but allows for decay in the value of parameters and tokens, thus generalizing to training for multiple epochs and with excess parameters. It can thus be used as a direct replacement of \autoref{eq:cc}. If $R_N^*$ and $R_D^*$ are unknown, one can simply set them to infinity by default, which will make \autoref{eq:fin} completely equivalent to \autoref{eq:cc}.

To learn the parameters $R_N^*$ and $R_D^*$, we largely follow the approach from \cite{hoffmann2022training}. We fix $a$, $b$, $e$, $\alpha$, $\beta$ to the values learned on C4 in \autoref{sec:c4scaling} and minimize:

\begin{equation}\label{eq:lsedata}
\begin{aligned}
 \min_{R_N^*,R_D^*}  \sum_{\text{Run }i} \text{Huber}_\delta \Big( & \text{LSE}\big(a - \alpha \log (U_N^i + U_N^i R_N^* (1 - e^{\frac{-R_N^i}{R_N^*}})), \\
 & b- \beta \log (U_D^i + U_D^i R_D^* (1 - e^{\frac{-R_D^i}{R_D^*}})), e \big) - \log L^i
 \Big)
\end{aligned}
\end{equation}

We use the LBFGS algorithm to find local minima of the objective above, started on a grid of initialization given by: $R_N^* \in \{0., 4.,\dots, 20. \}$ and $R_D^* \in \{0., 4.,\dots, 20. \}$. We fit on 182 samples with parameters varying from 7 million up to 9 billion and epochs ranging from 1 to 500. We removed outliers referenced in \autoref{sec:toomany} from our fitting, as our formulas do not allow for excess parameters or excess epochs to negatively impact performance. We assume excess parameters or epochs only cause performance to plateau but never to worsen. However, it is difficult to identify all samples where excess parameters or epochs hurt, as for some data budgets we only train a single model, thus we do not know if the loss of that model is already in the range where it starts to increase again. Further, there are samples where loss initially increases and then decreases as a function of epochs (double descent, see~\autoref{sec:dd}), which further contributes to noise in the fitting. Nevertheless, we are able to get a fairly stable fit resulting in $R_N^*=5.309743$ and $R_D^*=15.387756$. Since $R_D^* > R_N^*$, excess parameters decay faster. Hence, the data-constrained efficient frontiers in Figures~\ref{fig:returnalloc},\ref{fig:100misoloss} suggest scaling compute allocated to epochs faster than to parameters. This value of $R_D^*$ yields $\delta \approx 6 * 10^{-2}$ ($0.19$ for $R_N^*$), which respects the assumption that $\delta$ is small. Inserting these learned parameters and the parameters from \autoref{sec:c4scaling}, and simplifying \autoref{eq:optunud} yields the precise formulation we use to predict loss ($L$) given unique tokens ($U_N$), parameter repetitions ($R_N$) and data repetitions ($R_D$): 

\begin{equation}\label{eq:finvals}
\begin{aligned}
L(U_D,R_N,R_D)=&\frac{521}{(U_N + 5.3 \cdot U_N (1 - e^{\frac{-R_N}{5.3}}))^{0.35}} + \frac{1488}{(U_D + 15.4 \cdot U_D (1 - e^{\frac{-R_D}{15.4}}))^{0.35}} + 1.87\\ \\
& \text{ where } U_N = U_D \cdot 0.051
\end{aligned}
\end{equation}


\begin{table}[htbp]
\centering
\caption{\textbf{Comparison of different versions of our parametric fit.} All versions are fitted on the same 182 samples. We report the fitting loss and the $R^2$ (coefficient of determination) of the predicted loss compared to the actual loss. No decay corresponds to assuming Chinchilla holds for repeated data without modification necessary. For \autoref{eq:ar}, we use the same equation for $D$ and $N$ renaming the $\delta$ to $R_D^*$ and $R_N^*$.}
\begin{tabular}{c | c c | c c}
\toprule
Parametric Fit & $R_D^*$ & $R_N^*$ & Loss ($\downarrow$) & $R^2$ ($\uparrow$) \\
\midrule
No decay & - & - & - & 0.4452 \\
\autoref{eq:fin} but only decay $N$ & - & 713.0015 & 0.0241 & 0.4488 \\
\autoref{eq:fin} but only decay $D$ & 2.9157 & - & 0.0169 & 0.7354 \\
\autoref{eq:fin} & 15.3878 & 5.3097 & 0.0158 & 0.7722 \\
\autoref{eq:ar} for both $N$ and $D$ & 0.0104 & 0.3676 & 0.0155 & 0.7988 \\
\autoref{eq:expdecay} for both $N$ and $D$ & 0.0105 & 0.3676 &  0.0155 & 0.7987 \\
\autoref{eq:alphabeta} & 26530.611 & 2040.8163 & 0.0596 & 0.5110 \\
\bottomrule
\end{tabular}
\label{tab:fits}
\end{table}

We experiment with different versions of our formula and display the learned values in \autoref{tab:fits}. No decay or decaying only $D$ or $N$ of \autoref{eq:fin} leads to worse loss and $R^2$ than \autoref{eq:fin}. Thus, it is important to decay both the value of excess parameters and data repetitions. We also consider an explicit exponential where $D'=\sum_{k=0}^{R_D} U*e^{-R_D^*k}$, hence from \autoref{eq:geos} it follows:

\begin{equation}\label{eq:expdecay}
\begin{aligned}
D'=U\tfrac{1-(e^{-R_D^*})^{R_D + 1}}{1-e^{-R_D^*}}
\end{aligned}
\end{equation}

This explicit decay, \autoref{eq:ar}, and \autoref{eq:fin} all yield similar results with $R^{2}$ around 80. \autoref{eq:fin} fits the data slightly worse than \autoref{eq:ar}, likely due to our approximations. Nevertheless, we use \autoref{eq:fin} throughout as it has fewer terms, and we find it easier to interpret.

\FloatBarrier

\subsection{Analytical properties of compute-optimal point}

\begin{figure}[t]
    \centering
    \includegraphics[width=4in]{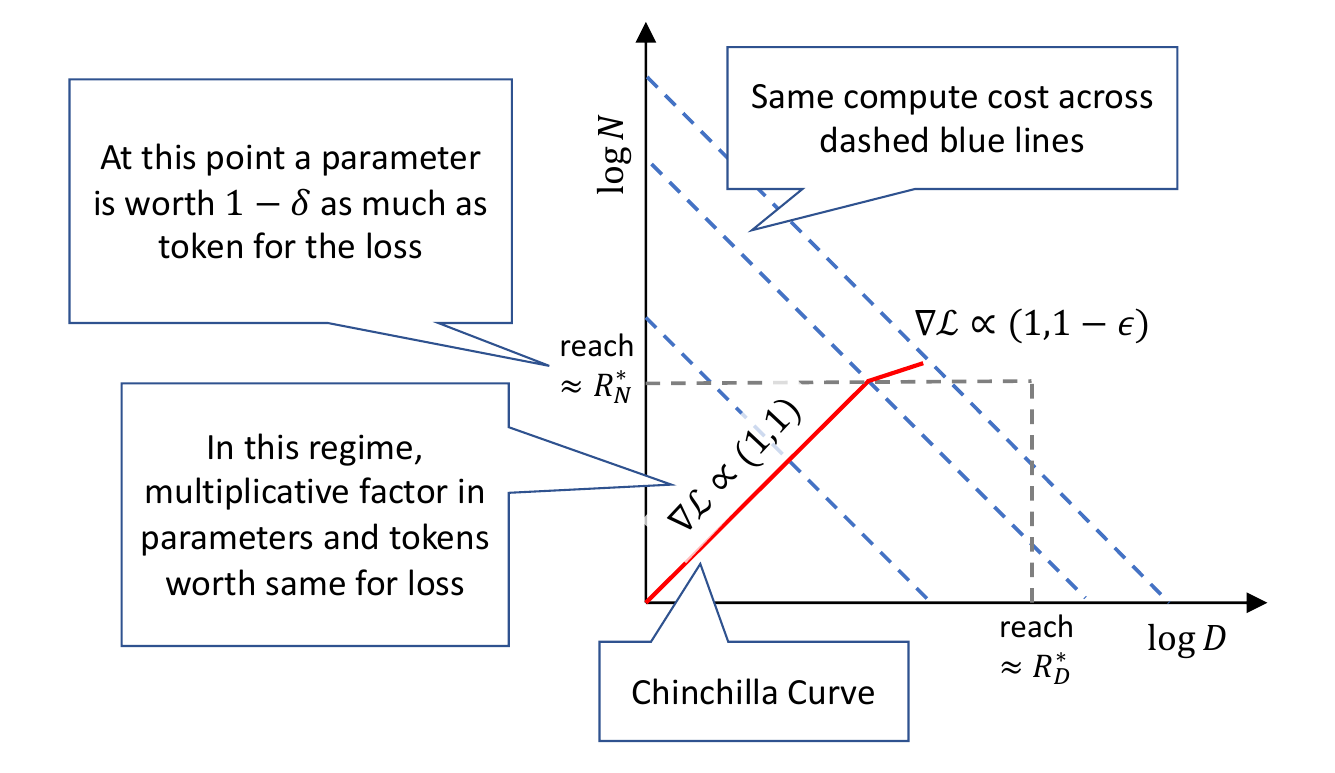}
    \caption{A cartoon of how the compute-optimal tradeoff deviates from Chinchilla as we increase the number of epochs. Initially the model size and tokens processed grow proportionally ($R_N=R_D$) but since $R^*_N < R^*_D$, at some point adding parameters offers worse returns compared to increasing the number of tokens processed, and hence we deviate from the Chinchilla curve.}
    \label{fig:cartoonmultiepoch}
\end{figure}

In our case, consider the setting of a fixed compute budget $C$ and a fixed budget of unique tokens $U_D$ implying a set of unique parameters $U_N$. Let $R_D$  denote the number of times we repeat data (we assume that we are in the multi-epoch regime and hence $R_D>0$).

Write $U_D = cU_N$ (for Chinchilla $c\approx 20$).
When $R_D \ll R^*_D$ and $R_N \ll R^*_N$, our scaling agrees with Chinchilla, and so the point $(U_N,U_D)$, corresponding to $R_D=R_N=0$ is on the optimal compute curve.
Increasing $R_D$ by $\epsilon$ corresponds to increasing the number of tokens by $\epsilon U_D = \epsilon c U_N$, while increasing $R_N$ by $\epsilon$ corresponds to increasing the number of parameters by $\epsilon U_N$.
For small positive $R_D,R_N$, our curve agrees with Chinchilla and so we need to increase $R_N,R_D$ by the same amount to maintain the proportionality. Hence up to some value $r>0$, the optimal compute curve corresponds to $R_N=R_D=r$.
Our curve differs from Chinchilla when $r$ gets closer to either $R^*_N$ or $R^*_D$. At this point, we start to see sharply diminishing returns.

In our setting, $R^*_D > R^*_N$ which means that we reach the point $r \approx R^*_N$ first. 
At this point, each added parameter is worth less (specifically worth $e^{-r/R^*_N}$), than an added data point, despite them having equal computational cost.
Hence processing more tokens will be more effective than increasing the number of parameters, and we expect the optimal compute curve to break away from proportionality.
This is indeed what we see.

\FloatBarrier

\section{C4 Scaling Coefficients}
\label{sec:c4scaling}

While \citet{hoffmann2022training} have shown that the equal scaling of model parameters and training tokens holds across different training datasets, the precise ratios vary considerably across datasets and approaches. For example given the Gopher~\cite{rae2021scaling} compute budget of $5.76 \times 10^{23}$ FLOPs, their parametric loss function fitted on MassiveWeb predicts an optimal allocation of 40 billion parameters. Meanwhile, if the training dataset is C4~\cite{raffel2020exploring} their IsoFLOP approach predicts 73 billion parameters to be optimal, almost twice as much. However, for C4, which is our training dataset, they do not provide the coefficients necessary to compute loss with their parametric loss function. Based on their IsoFLOP training runs on C4, they only provide the information that for C4, compute ($C$) allocated to data ($D$) and parameters ($N$) should be scaled \emph{exactly} equally for optimality, i.e. $a=b=0.5$ in the relationship $N_{opt} \propto C^a$ and $D_{opt} \propto C^b$. This corresponds to $\alpha=\beta$ in the parametric loss function (\autoref{eq:ccbase}). Thus, we use this information together with the methodology and C4 data points from \cite{hoffmann2022training} to fit the parametric loss function. We tie the parameters $\alpha$ and $\beta$ to be equal and optimize

\begin{equation}\label{eq:lse}
     \min_{a, b, e, \alpha, \beta}  \sum_{\text{Run }i} \text{Huber}_\delta \Big(\text{LSE}\big(a - \alpha \log N_i, b- \beta \log D_i, e \big) - \log L_i\Big)
\end{equation}

where $\text{LSE}$ is the log-sum-exp operator and $N_i$, $D_i$ and $L_i$ the model size, dataset size and loss of the $i$th run, and $\delta = 10^{-3}$. We fit on 54 samples on a grid of initialization given by: $\alpha \in \{0., 0.5,\dots, 2. \}$, $\beta \in \{ 0., 0.5,\dots, 2.\}$, $e \in \{-1., -.5, \dots, 1. \}$, $a \in \{0, 5, \dots, 25 \}$, and $b \in \{0, 5, \dots, 25 \}$.
Our fit results in $a=6.255414$, $b=7.3049974$, $e=0.6254804$, $\alpha=\beta=0.3526596$. Exponentiating $a$, $b$ and $e$ to get $A$, $B$ and $E$ and inserting all learned coefficients into \autoref{eq:ccbase} then allows us to compute loss ($L$) as a function of parameters and data: 

\begin{equation}
    L(N, D) = 1.87 + \frac{521}{N^{0.353}} + \frac{1488}{D^{0.353}}
\end{equation}

To verify the accuracy of our fit, we benchmark the predictions with those of the IsoFLOP C4 curves in \cite{hoffmann2022training}. Following \cite{hoffmann2022training}, we can compute the optimal number of parameters $N_{opt}$ and tokens $D_{opt}$ for our fit using:

\begin{equation}\label{eq:ccoptapp}
\begin{aligned}
N_{opt}(C) = G {\left(\frac{C}{6}\right)}^{a}, \quad
D_{opt}(C) = G^{-1} {\left(\frac{C}{6}\right)}^{b}\\
\quad \text{ where } \quad G = {\left(\frac{\alpha A}{\beta B} \right)}^{\frac{1}{\alpha + \beta}},\quad
a = \frac{\beta}{\alpha+\beta}, \text{ and } b = \frac{\alpha}{\alpha + \beta}
\end{aligned}
\end{equation}

Given the Gopher compute budget of $C=5.76 \times 10^{23}$ our fitted parameters predict an optimal allocation of $N_{opt}=70.0$ billion parameters and $D_{opt}=1.37$ trillion tokens. This is very close to the 73 billion parameters and 1.3 trillion tokens predicted by the IsoFLOP curves on C4 from \cite{hoffmann2022training} and thus we consider it a good fit. We use these fitted parameters rather than the MassiveWeb parameters for all computations involving Chinchilla scaling laws.

\FloatBarrier

\section{Additional Contour Plots}
\label{sec:addcon}

\begin{figure*}[htbp]
    \centering
    \begin{center}
        \includegraphics[width=\textwidth]{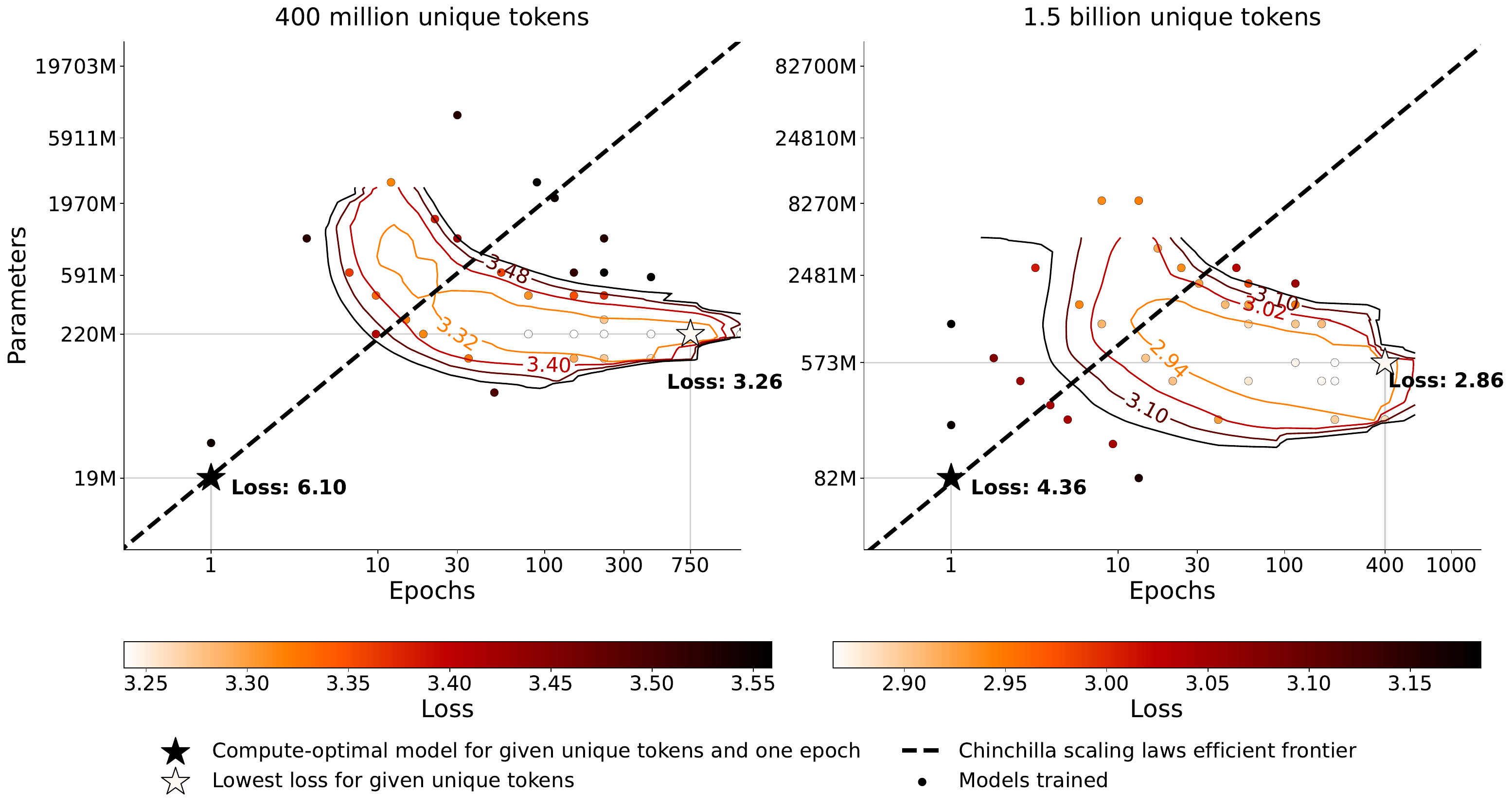}
        \caption{\textbf{Empirical isoLoss curves for 400 million and 1.5 billion unique tokens.} 34 models trained on 400 million unique tokens and 37 models trained on 1.5 billion unique tokens with varying parameters and epochs.}
        \label{fig:400m1b5isoloss}
    \end{center}
\end{figure*}

\autoref{fig:400m1b5isoloss} contains additional empirical isoLoss contours for 400 million and 1.5 billion unique tokens. Results show that like in \autoref{fig:100misoloss} significantly lower loss can be achieved by increasing parameters and epochs beyond what is compute-optimal at a single epoch. The lowest loss is also achieved by allocating more extra compute to repeating data rather than to adding parameters.

\FloatBarrier

\section{Double Descent}
\label{sec:dd}

\begin{figure*}[htbp]
    \centering
    \begin{center}
        \includegraphics[width=\textwidth]{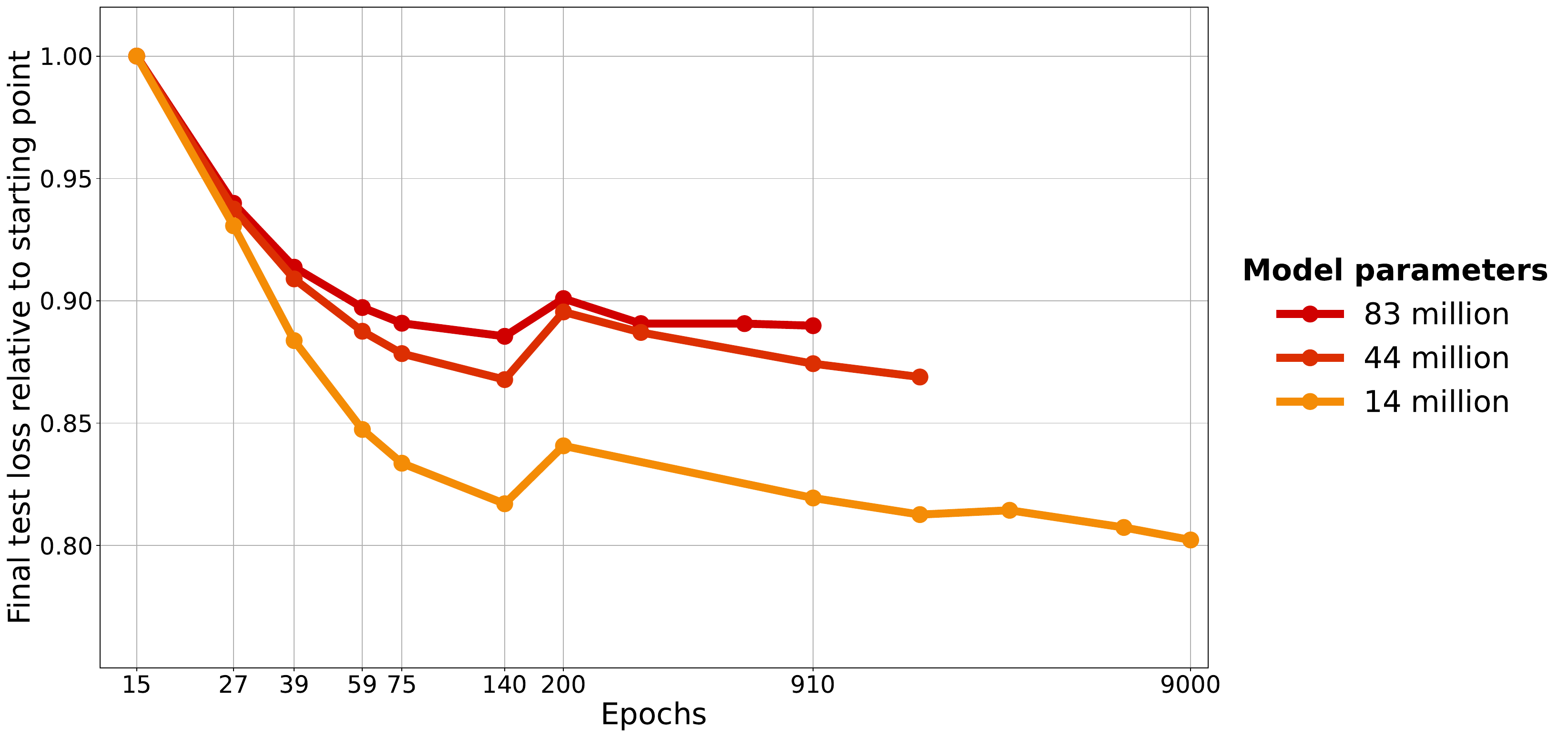}
        \caption{\textbf{Double descent.} Each dot is a model trained on 100 million unique tokens. Loss initially increases at 200 epochs and then decreases again; this is known as epoch-wise double descent~\cite{nakkiran2021deep}.}
        \label{fig:dd}
    \end{center}
\end{figure*}

Prior work has reported double descent phenomena when repeating data, where the loss initially increases and then decreases again as the model is trained for more epochs~\cite{nakkiran2021deep,hernandez2022scaling}. In \autoref{fig:dd}, we plot the loss curves of several models trained for varying epochs on 100 million tokens. We find double descent phenomena with the loss of all models increasing at 200 epochs before decreasing again. This contributes to additional noise in the fitting of our functions in \autoref{sec:scalinglaws}, as our functional form assumes loss to be monotonically decreasing as epochs increase. Thus, we remove most such examples from the fitting.

\FloatBarrier

\section{Repeating on Heavily Deduplicated Data}

\begin{figure*}[htbp]
    \centering
    \begin{center}
        \includegraphics[width=\textwidth]{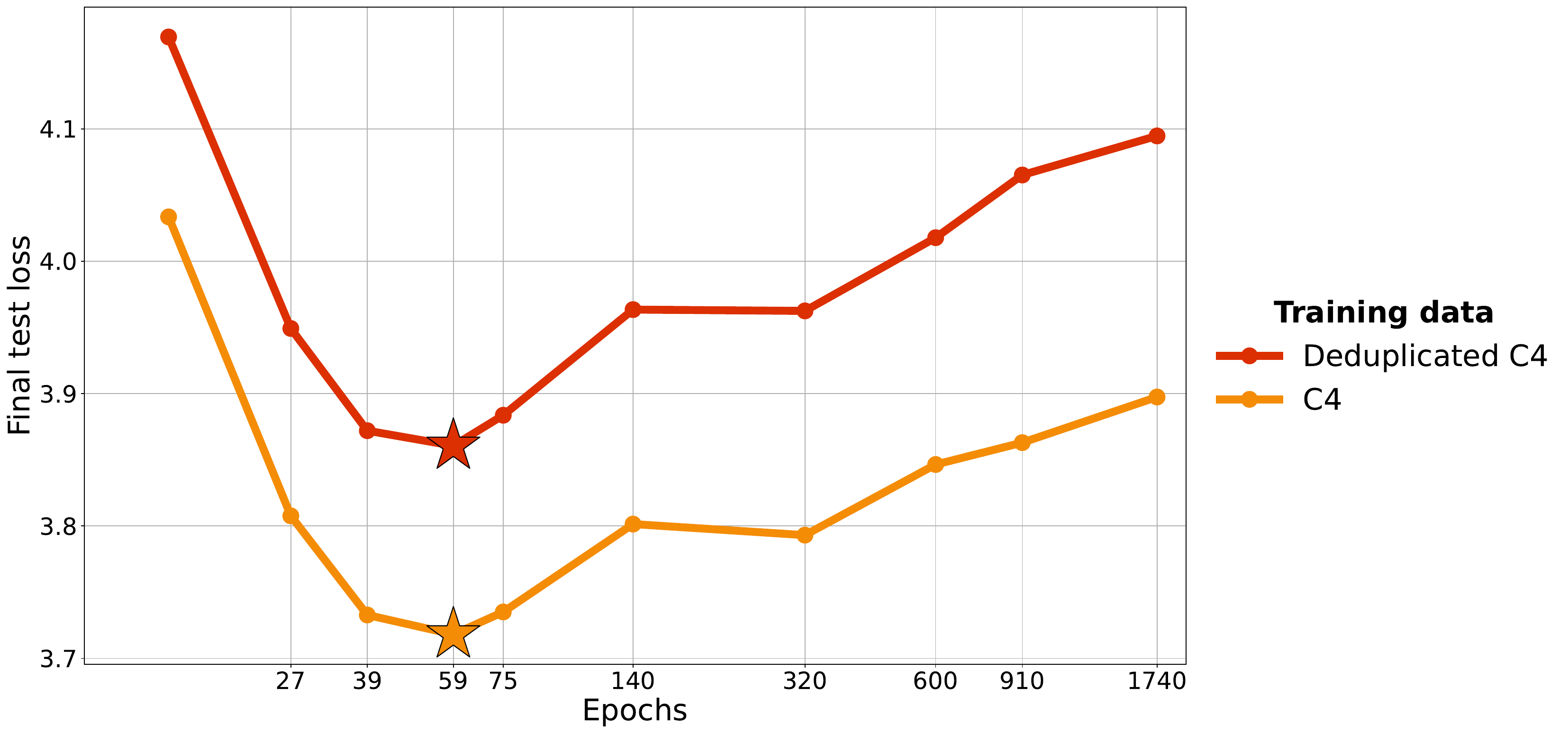}
        \caption{\textbf{Optimal loss on deduplicated data.} 146 million parameter models trained on 100 million unique tokens that are either directly from C4 or undergo additional deduplication. Each dot is a single model. While deduplication results in a higher test loss, the optimal number of epochs remains the same whether or not deduplication is performed (see also \autoref{fig:100misoloss}).}
        \label{fig:dedup}
    \end{center}
\end{figure*}

To investigate whether \autoref{fig:100misoloss} is dependent on the inherent amount of duplicates in the selected 100 million tokens, we train several models on a deduplicated version of C4 (see \autoref{sec:filtering}). We plot the performance of the models trained on the deduplicated C4 versus the regular C4 in \autoref{fig:dedup}. All models are evaluated on the same validation dataset from the regular C4. Regardless of deduplication we find 59 epochs to be optimal and the overall trend to be very similar. Together with our results on OSCAR~(\autoref{sec:fixoscar}), this suggests that our work generalizes to different datasets with different inherent amounts of duplicates.

\FloatBarrier

\section{Do Excess Parameters Hurt, Plateau or Help?}
\label{sec:toomany}

\begin{figure*}[htbp]
    \centering
    \begin{center}
        \includegraphics[width=\textwidth]{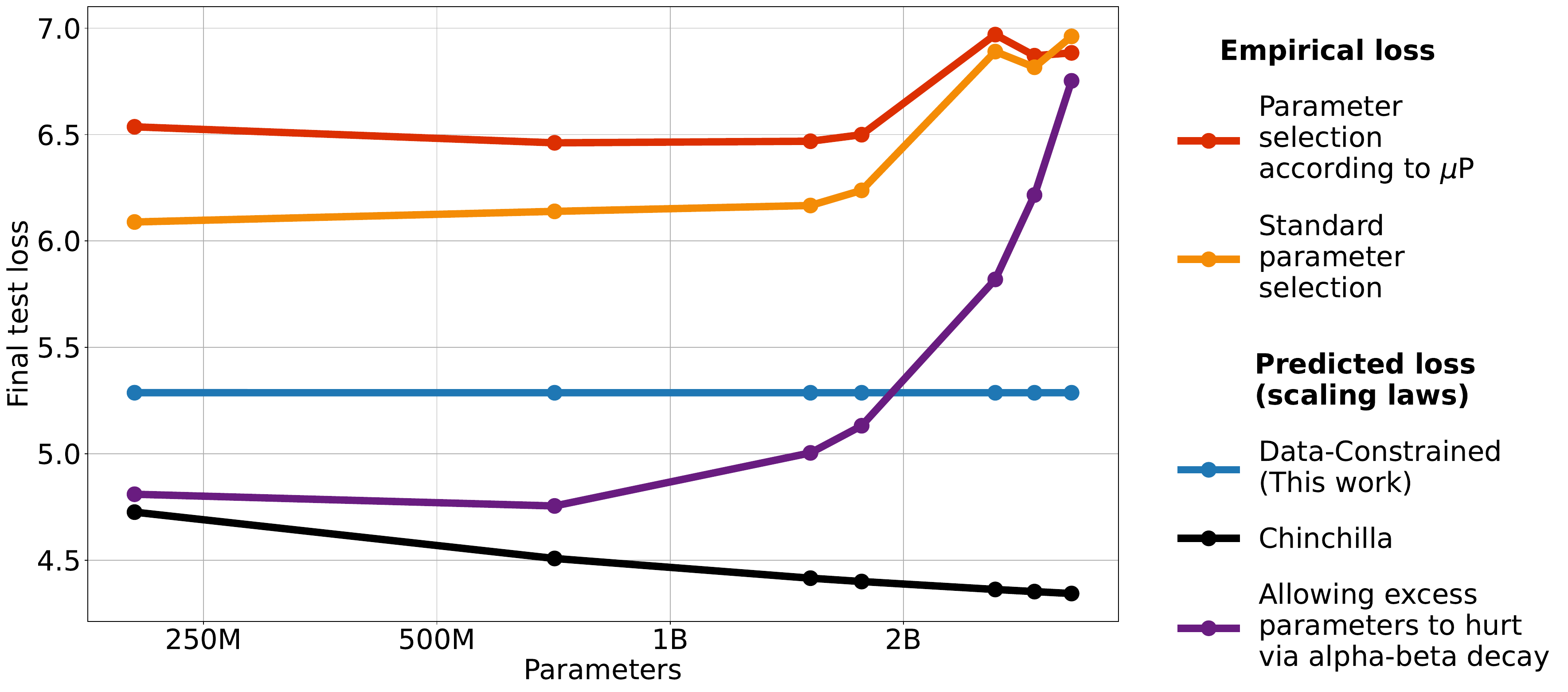}
        \caption{\textbf{Empirical and predicted losses of LLMs trained on 100 million tokens for a single epoch.} Excess parameters empirically hurt performance, but this may be due to a lack of regularization. Thus, our scaling formula predicts loss to plateau, while Chinchilla predicts loss to improve. By decaying the exponent $\alpha$ (and $\beta$) instead, one can allow excess parameters to hurt.}
        \label{fig:mup}
    \end{center}
\end{figure*}

Figures~\ref{fig:100misoloss},~\ref{fig:400m1b5isoloss} suggest that excess parameters (or epochs) can harm performance. We hypothesize that this is due to suboptimal hyperparameters and could be prevented with better regularization. Thus, we expect with optimal regularization hyperparameters excess parameters would never hurt, but performance would merely plateau, as in extreme cases regularization could just take the form of removing the excess parameters. One approach to selecting optimal hyperparameters is $\mu$P~\cite{yang2021tuning}. We compare excessively large models trained with a data constraint of $D_C=100$ million tokens in \autoref{fig:mup} across $\mu$P, our default hyperparameters (\autoref{sec:arch}) and scaling law predictions. Surprisingly, $\mu$P leads to even higher test loss than our default hyperparameters. Nevertheless, we find that also with $\mu$P excessive parameters hurt: The models with more than 2 billion parameters have significantly higher validation loss after training than the models with 200 million to 1 billion parameters when trained on only 100 million tokens. However, $\mu$P only covers hyperparameters such as the learning rate, but not explicit regularization hyperparameters like dropout rates, which we hypothesize would prevent this behavior. Thus, our proposed scaling equations predict loss to plateau, as seen in the straight line. As the compute-optimal parameter count for 100 million tokens is around 7 million, all depicted models have a significant amount of excess parameters and data-constrained scaling laws predict their losses to be all the same ($R_N^* \ll R_N$). Meanwhile, the default Chinchilla scaling law~\cite{hoffmann2022training} predicts loss to continue decreasing as parameters are added, which is in stark contrast to the empirical data.

If one wants to incorporate excess parameters hurting performance into the scaling law equations, one could consider \textbf{(a)} Modifying the exponential decay formulation introduced in \autoref{sec:scalinglaws} such that instead of the value of repeated data decaying to $0$ it decays to a large negative value \textbf{(b)} decaying the exponents $\alpha$ and $\beta$ in \autoref{eq:cc} instead of $D$ and $N$. Decaying the exponents to $0$ has the effect of more repetitions eventually hurting performance as $\lim_{\alpha\to0}D^{\alpha} = 1$ and the same for $\beta$. Thus, initially as $D$ and $N$ increase loss decreases, but ultimately the decay of $\alpha$ and $\beta$ pushes $D$ and $N$ back to $1$ resulting in loss to increase. Specifically, approach \textbf{(b)} could take the form of:

\begin{equation}
\label{eq:alphabeta}
L(N,D,R_N,R_D)=E + \frac{A}{N^{\alpha * max(0, 1 - (R_N/R_N^*))}} + \frac{B}{D^{\beta * max(0, 1 - (R_D/R_D^*))}}
\end{equation}

\begin{figure*}[htbp]
    \centering
    \begin{center}
        \includegraphics[width=\textwidth]{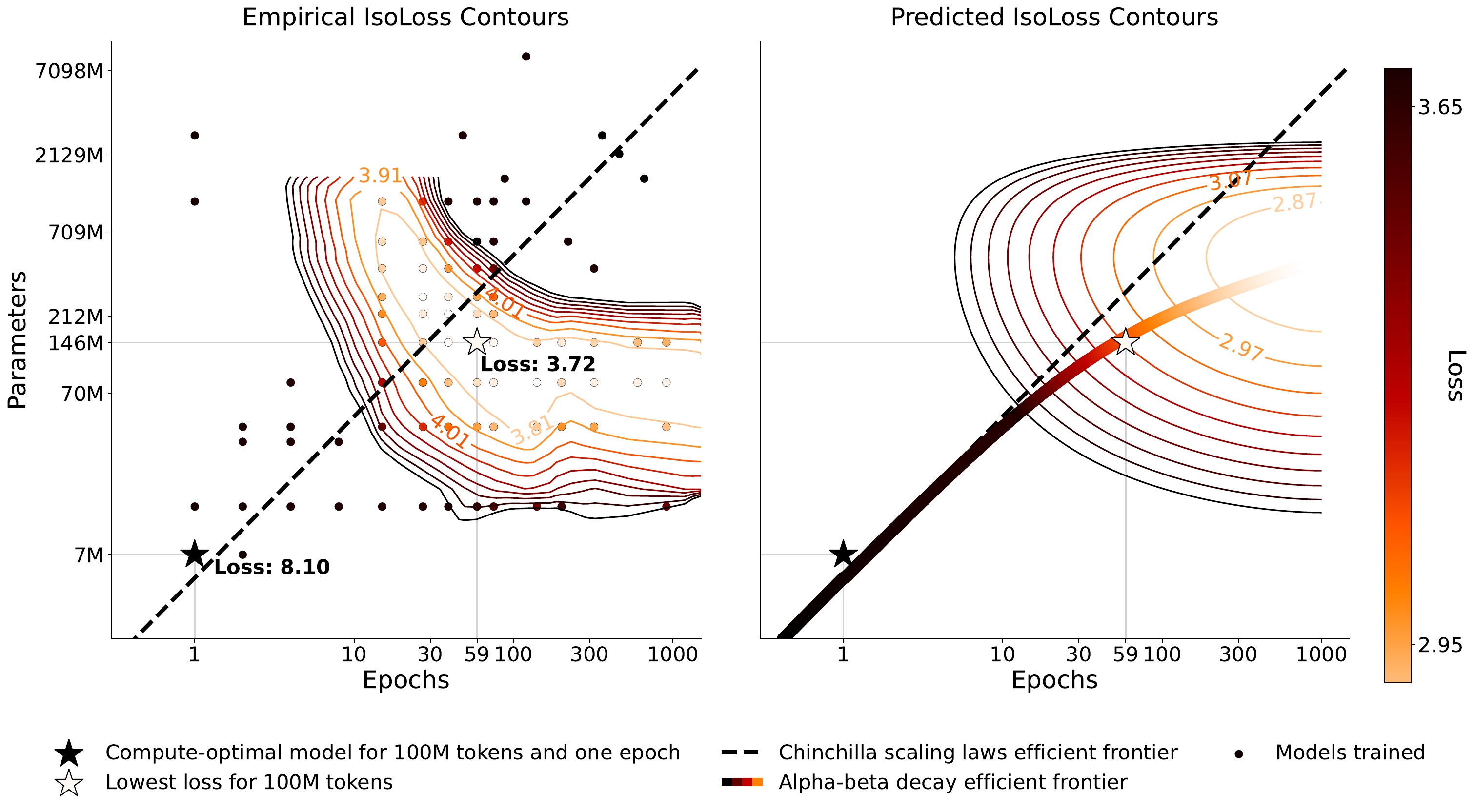}
        \caption{\textbf{IsoLoss contours for 100 million unique tokens with contours predicted by parametric decay of alpha and beta.} The same models from \autoref{fig:100misoloss} with the contour predictions being done by the alpha-beta decay formulation introduced in \autoref{sec:toomany}.}
        \label{fig:alphabeta}
    \end{center}
\end{figure*}

Like the equations in \autoref{sec:scalinglaws} this formulation also reduces to the Chinchilla scaling laws in the base case of $R_D=0$ or $R_N=0$. As the exponents decrease with more repetitions adding parameters or epochs becomes less beneficial. Eventually, the decay in $\alpha$ or $\beta$ causes loss to increase again as it pushes $N$ or $D$ back down to $1$. We fit this formula using the same approach outlined in \autoref{sec:scalinglaws} but including samples where excess parameters or epochs hurt (296 total samples). We use a grid of initialization given by: $R_N^* \in \{0., 2000.,\dots, 100000. \}$ and $R_D^* \in \{0., 2000.,\dots, 100000. \}$. This results in $R_D^*=26530.611$ and $R_N^*=2040.8163$. $R_N^*$ is significantly lower resulting in excess parameters hurting faster than excess epochs, which is in line with empirical data from~\autoref{fig:100misoloss}. We visualize \autoref{fig:100misoloss} with the predictions from this alpha-beta decay formulation in \autoref{fig:alphabeta}. Expected parameters eventually hurt resulting in circle-shaped contours. Due to the very high $R_D^*$ the area where epochs start to hurt is outside of the boundaries of~\autoref{fig:alphabeta}. While the predicted optimal allocation (efficient frontier) is similar to~\autoref{fig:100misoloss}, the predicted return from repeated data differs significantly. The alpha-beta decay formulation incorrectly predicts returns to diminish significantly slower as seen by the longer efficient frontier and the smaller distance in contours early on as compared to~\autoref{fig:100misoloss}. Beyond its potentially useful properties, we do not have a rigorous mathematical justification for this alpha-beta decay formulation which could be the cause of the incorrect return predictions.

Ultimately, we settle on our exponential decay formulation from~\autoref{sec:scalinglaws} that does not allow excess parameters or epochs to hurt, as preventing such behavior is trivial by stopping training (in the case of epochs hurting) or removing excess parameters (in the case of model parameters hurting). Further, accurately predicting how much loss increases in the limit is not very useful, as in practice one would want to stop training when it's expected to plateau anyways.

\FloatBarrier

\section{Case Study: Galactica}
\label{sec:galactica}

\begin{figure*}[htbp]
    \centering
    \begin{center}
        \includegraphics[width=\textwidth]{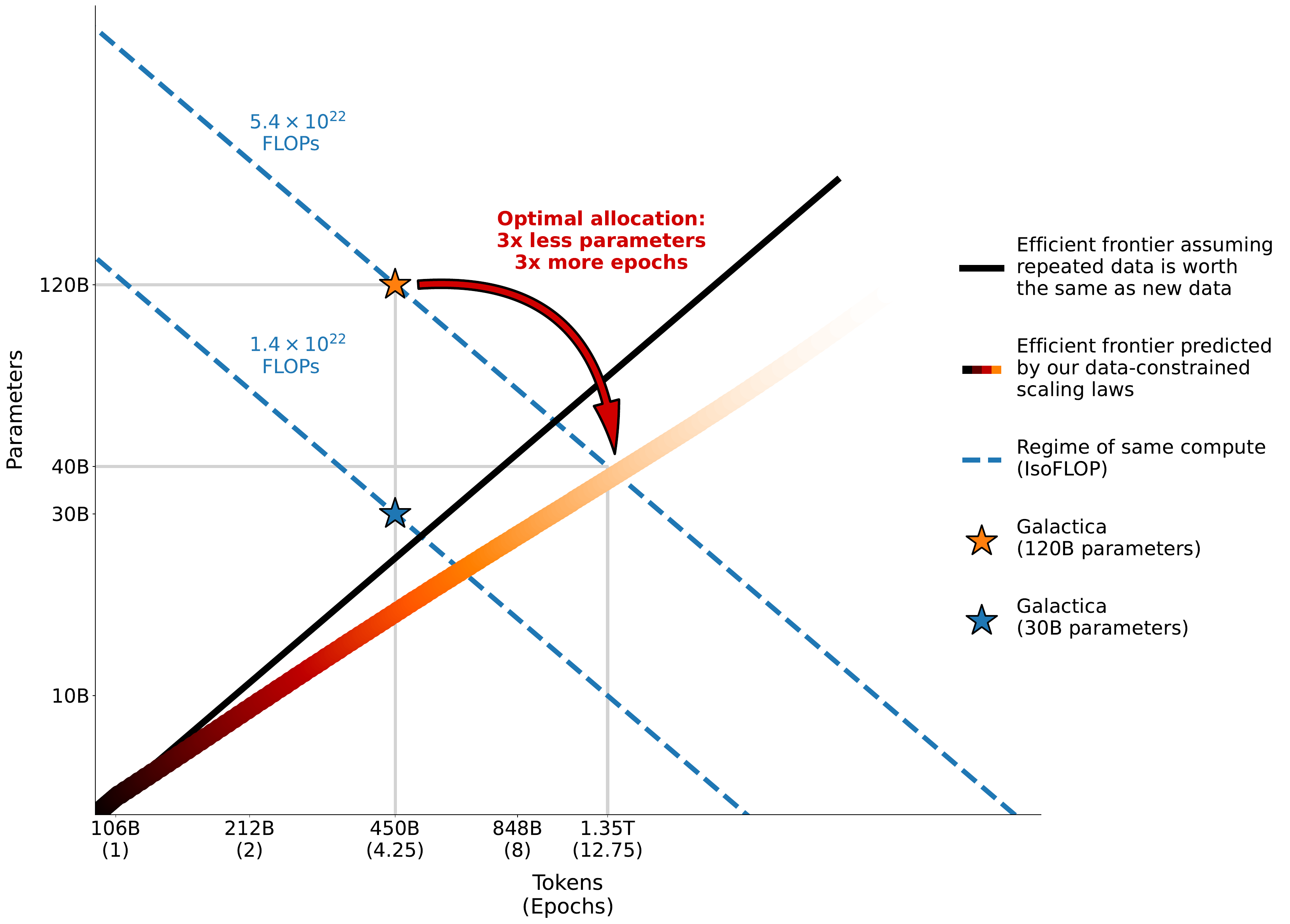}
        \caption{\textbf{Optimal compute allocation for Galactica.} Efficient frontier assuming repeated data is worth the same as new data (Chinchilla scaling laws) and data-constrained efficient frontier assuming a unique token budget of 106 billion tokens like for the Galactica models~\cite{taylor2022galactica}. For optimal compute allocation according to our proposed data-constrained scaling laws, the 120 billion Galactica model should have been significantly smaller and trained for more epochs.}
        \label{fig:galactica}
    \end{center}
\end{figure*}

The Galactica models~\cite{taylor2022galactica} are the only publicly known LLMs that explicitly trained for a significant number of epochs prior to this work. They trained their models on 106 billion unique tokens for 4.25 epochs. Our findings on~\emph{Return} from repeated data agree with their conclusion that multiple epochs are beneficial, however, we find that even more epochs can be beneficial and a small spike in validation loss does not justify stopping training (\autoref{sec:addvalloss}). Meanwhile, our findings on \emph{Allocation} significantly deviate from Galactica. \autoref{fig:galactica} visualizes the Galactica models with our predicted efficient frontier in the same style as \autoref{fig:returnalloc}. The creators of Galactica decided to train a 120 billion parameter model on 450 billion tokens, a significant overallocation to parameters even in Chinchilla terms (black efficient frontier). This decision was likely driven by the intuition that repeated data is worth less, thus one should spend more compute on parameters. However, our empirical data contradicts this. Parameters learning from repeated data are worth \emph{even less} than repeated data, thus one should overallocate to epochs, not parameters. Our data-constrained scaling laws thus predict that a better model could have been trained by allocating significantly more FLOPs to epochs rather than parameters for the largest Galactica model with 120 billion parameters. Specifically, 40 billion parameters trained for 1.35 trillion tokens (12.75 epochs) would have been optimal according to data-constrained scaling laws. Note that these scaling laws have been fitted on C4, which is not the dataset used to pre-train Galactica. The Galactica models are pre-trained on a predominantly scientific dataset, which includes code data among other data sources. Results from~\cite{hoffmann2022training} show that there are differences in the scaling coefficients when training on C4 as compared to GitHub code, however, the overall allocation trend is the same. Thus, while we expect a smaller model trained for more epochs to be better than the 120 billion parameter model, the optimal allocation is unlikely to be exactly 40 billion parameters and 1.35 trillion tokens.

\FloatBarrier

\section{Training Loss}
\label{sec:trainlossc4}

\begin{figure*}[htbp]
    \centering
    \begin{center}
        \includegraphics[width=\textwidth]{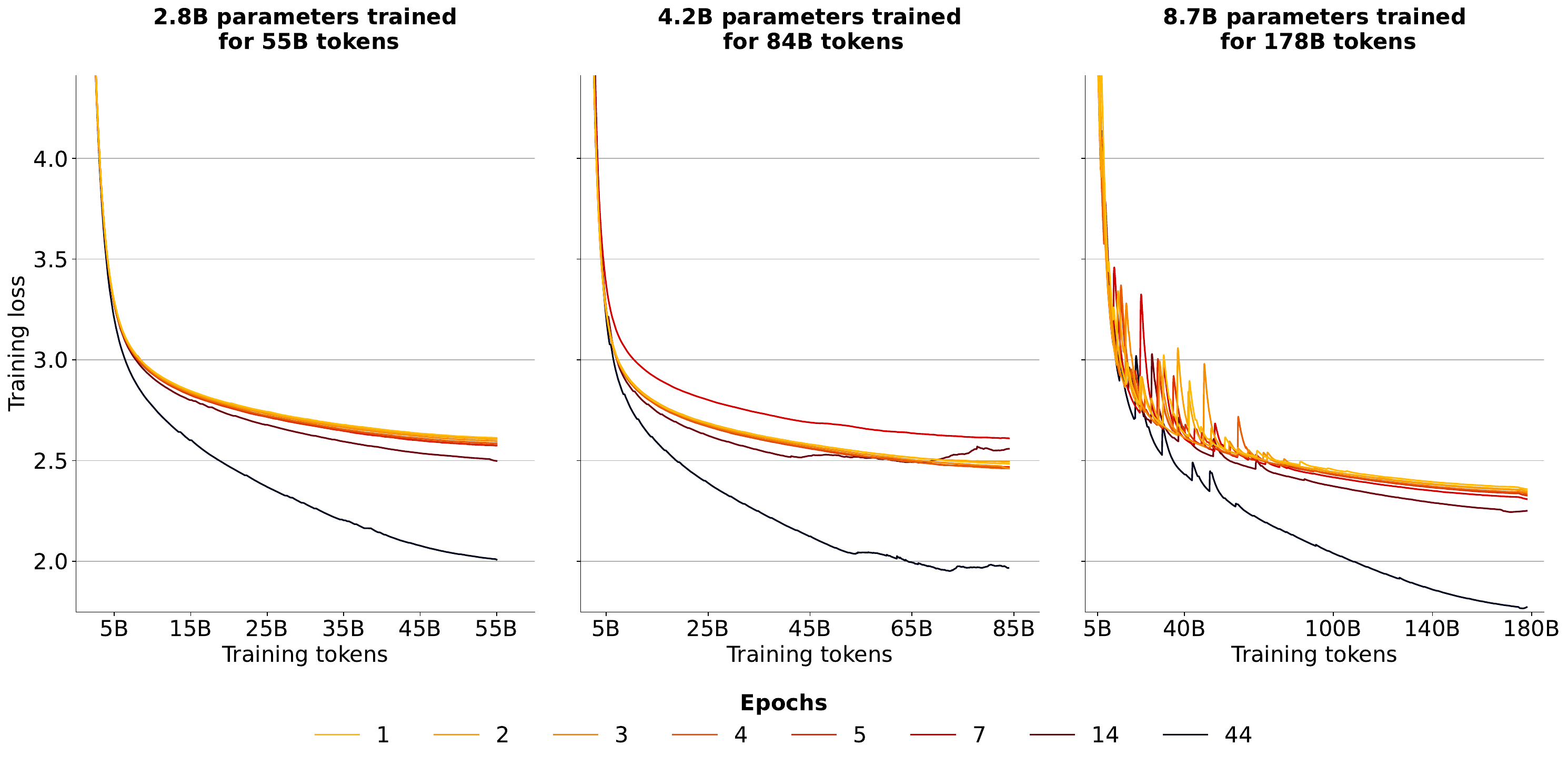}
        \caption{\textbf{Training loss smoothed with exponential moving average smoothing and a weight of 0.999.} Models trained on fewer unique tokens (more epochs) have better training loss as they overfit.}
        \label{fig:trainingc4}
    \end{center}
\end{figure*}

\citet{hoffmann2022training} use training loss as their core metric. However, when repeating data for multiple epochs, training loss is a bad metric as models will overfit to the limited data available as shown in \autoref{fig:trainingc4}. Thus, we use loss on a held-out test set as our key performance metric.

\FloatBarrier

\section{Scaling Curves on the OSCAR Corpus}
\label{sec:fixoscar}

\begin{figure*}[htbp]
    \centering
    \begin{center}
        \includegraphics[width=\textwidth]{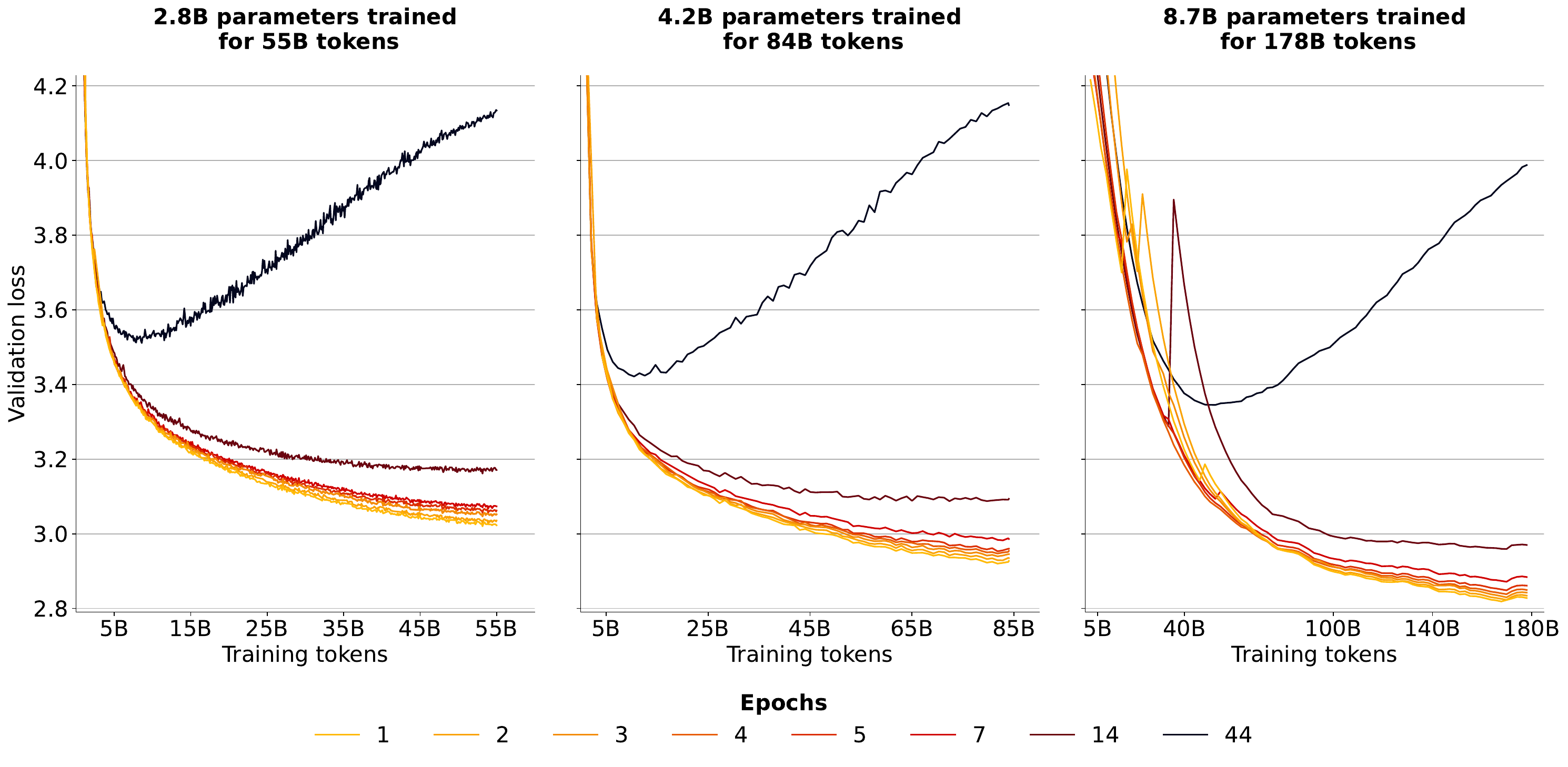}
        \caption{\textbf{Validation loss during training for models trained on OSCAR.} Models trained on tokens that are repeated for multiple epochs have consistently worse loss.}
        \label{fig:validationoscar}
    \end{center}
\end{figure*}

\begin{figure*}[htbp]
    \centering
    \begin{center}
        \includegraphics[width=\textwidth]{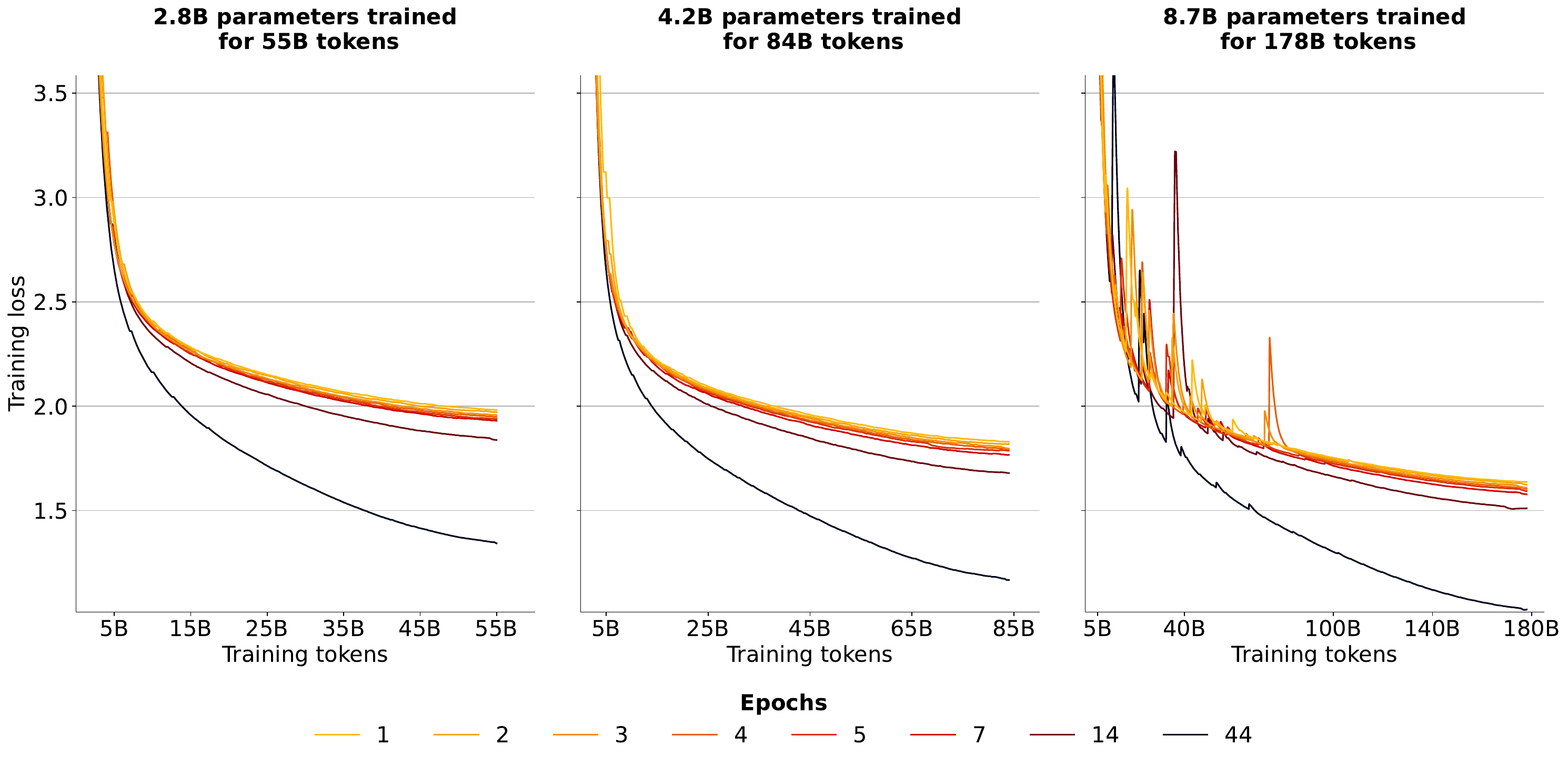}
        \caption{\textbf{Training loss for models trained on OSCAR smoothed with exponential moving average smoothing and a weight of 0.999.} Models trained on fewer unique tokens (more epochs) have better training loss as they overfit.}
        \label{fig:trainingoscar}
    \end{center}
\end{figure*}

To ensure our findings are not dataset-dependent, we train models with the same configurations from \autoref{fig:validation} on the OSCAR corpus~\cite{ortiz-suarez-etal-2020-monolingual}. OSCAR is considered noisier than C4~\cite{raffel2020exploring} due to its less stringent duplication. Figures~\ref{fig:validationoscar},\ref{fig:trainingoscar} depict the validation and training loss of these models. We find the trend to be the same as for models trained on C4: While models with fewer repeats have better loss, differences for a few repeats are insignificant.

\FloatBarrier

\section{Validation Loss by Epoch}
\label{sec:addvalloss}

\begin{figure*}[htbp]
    \centering
    \begin{center}
        \includegraphics[width=\textwidth]{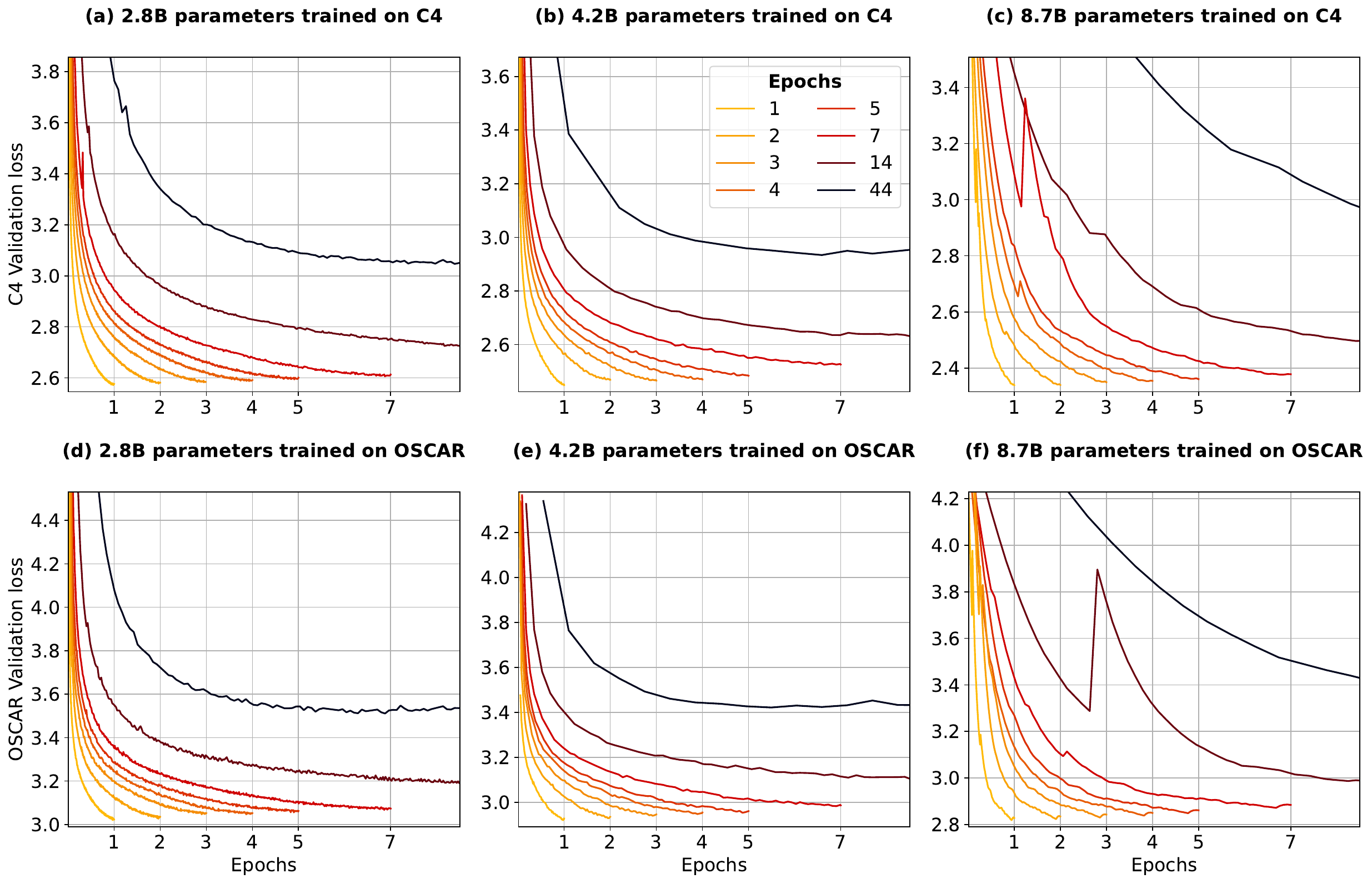}
        \caption{\textbf{Validation loss during training visualized by epochs.} Loss progresses smoothly throughout training. There are temporary spikes for 8.7 billion parameter models, commonly at the start of a new epoch.}
        \label{fig:vallossepochs}
    \end{center}
\end{figure*}

\citet{taylor2022galactica} decided to early-stop pre-training of the Galactica models due to a small increase in validation loss at the start of the fifth epoch. In \autoref{fig:vallossepochs} we plot the validation loss curves of our isoFLOP models as a function of epochs. We do find small increases in validation loss when models enter a new epoch. For example, upon entering the third and fourth epoch, the 7-epoch 8.7 billion parameter OSCAR model shows loss spikes. However, these are temporary and loss continues to go down smoothly thereafter. Thus, we hypothesize that the Galactica models could have attained better performance by continuing pre-training beyond the loss spike experienced at the beginning of the fifth epoch.

\FloatBarrier

\section{Evaluation Details}
\label{sec:eval}

\begin{table}[htbp]
\centering
\caption{\textbf{Setup for computing validation loss during training.} At every \emph{Evaluation Interval}, loss is computed on \emph{Evaluation Tokens} many tokens from the validation set. The evaluation tokens vary with the interval, i.e. the evaluation tokens at 100 steps are not the same as at 200 steps. However, the tokens do not vary across data budgets for the same FLOP budget (\autoref{fig:validation}). For example, $N=2.8$ billion parameter models with $D_C=55$ billion tokens are evaluated on the same data as models with $D_C=28$ billion tokens at each evaluation interval.}
\begin{tabular}{c c c c}
\toprule
FLOP budget & Parameters & Evaluation Interval & Evaluation Tokens \\
\midrule
$9.3 \times 10^{20}$ & $2.8$B & 100 & 105 million\\
$2.1 \times 10^{21}$ & $4.2$B & 1000 & 105 million\\
$9.3 \times 10^{21}$ & $8.7$B & 1000 & 2.1 million\\
\bottomrule
\end{tabular}
\label{tab:setupval}
\end{table}

\paragraph{Loss evaluation} For all models trained on C4, the final test loss is computed on the same 210 million tokens from the C4 validation set after training. For held-out evaluation during training, such as in \autoref{fig:validation}, the configurations are displayed in \autoref{tab:setupval}. The small number of evaluation tokens for the 8.7 billion parameter models likely contributes to the loss spikes for 8.7 billion parameter models seen in \autoref{fig:validation}. Thus, we smooth the validation loss curves of 8.7 billion parameter models with exponential moving average smoothing and a weight of 0.85. For training on OSCAR, configurations are the same, however, the validation split used is a held-out part from the OSCAR training split, as there is no official validation split for OSCAR. All training loss curves for C4 and OSCAR models are smoothed with exponential moving average smoothing and a weight of 0.999.

\paragraph{Downstream evaluation} We provide statistics of all downstream evaluation datasets in \autoref{tab:eval}. We use the evaluation-harness frameworks from BigScience and EleutherAI~\cite{eval-harness} to evaluate models on 19 evaluation datasets. For each dataset, a maximum of 3000 samples are evaluated with 0,1,2,3,4 and 5 few-shots~\cite{brown2020language} to produce six scores which are then averaged.  We normalize scores to range from the random baseline of each task to 1 and report them as percentages. For example, if random guessing produces 50\% accuracy and the maximum accuracy possible is 100\%, then a raw accuracy
of 55\% would be normalized to 10\%, and a raw accuracy of 45\% would be normalized to -10\% since it is worse than random. This is done to give all tasks the same weight. Otherwise average performance would heavily depend on generative tasks, where the random baselines are 0. Prompts are sourced from GPT-3~\cite{brown2020language} and PromptSource~\cite{promptsource} and detailed in \autoref{sec:samples}. We note that our evaluation is in no means comprehensive and a larger benchmarking would be helpful~\cite{srivastava2022beyond,muennighoff2022mteb}. However, by training five seeds for most models benchmarked, always averaging 0-5 fewshots, and ensuring maximum data overlap for repeated data (\autoref{sec:exp}) we significantly reduce uncertainty.

\begin{table}[htbp]
  \centering
  \caption{\textbf{Downstream evaluation datasets.} We evaluate on 19 datasets: The first 14 are evaluated using accuracy (ANLI counted as three), the next 4 using ROUGE-2 f-measure~\cite{lin2004rouge} and bAbI using exact match.}
  \resizebox{\textwidth}{!}{\begin{tabular}{lllll}
    \toprule
    Dataset & Split(s) & Samples & Baseline & URL \\
    \midrule
    ANLI~\cite{nie2019adversarial} & dev\_r{1,2,3}  & 3000 & 33.3 & \rurl{hf.co/datasets/anli} \\
    ARC-Easy~\cite{allenai:arc} & test & 1172 & 25.0 & \rurl{hf.co/datasets/ai2_arc} \\
    ARC-Challenge~\cite{allenai:arc} & test & 2376 & 25.0 & \rurl{hf.co/datasets/ai2_arc} \\
    BoolQ~\cite{clark2019boolq} & validation & 3270 & 50.0 & \rurl{hf.co/datasets/boolq} \\
    CB~\cite{de2019commitmentbank} & validation & 56 & 33.3 & \rurl{hf.co/datasets/super_glue} \\
    Copa~\cite{roemmele2011choice} & validation & 100 & 50.0 & \rurl{hf.co/datasets/super_glue} \\
    HellaSwag~\cite{zellers2019hellaswag} & test & 10003 & 25.0 & \rurl{hf.co/datasets/hellaswag} \\
    PiQA~\cite{Bisk2020} & validation & 1838 & 50.0 & \rurl{hf.co/datasets/piqa} \\
    RTE~\cite{dagan2006pascal,wang2019superglue} & validation & 277 & 50.0 & \rurl{hf.co/datasets/super_glue} \\
    SciQ~\cite{welbl2017crowdsourcing} & test & 1000 & 25.0 & \rurl{hf.co/datasets/sciq} \\
    StoryCloze 2016~\cite{mostafazadeh2017lsdsem} & test & 1871 & 25.0 & \rurl{hf.co/datasets/story_cloze} \\
    WinoGrande XL~\cite{sakaguchi2021winogrande} & test & 1267 & 50.0 & \rurl{hf.co/datasets/winogrande} \\
    \midrule
    E2E NLG~\cite{dusek.etal2020:csl} & test & 4693 & 0.0 & \rurl{hf.co/datasets/e2e_nlg_cleaned} \\
    XSUM~\cite{xsum-emnlp,gehrmann2021gem} & test & 11334 & 0.0 & \rurl{hf.co/datasets/GEM/xsum} \\
    WebNLG EN~\cite{castro-ferreira20:bilin-bi-direc-webnl-shared,gehrmann2021gem} & test & 5150 & 0.0 & \rurl{hf.co/datasets/GEM/web_nlg} \\
    WikiLingua EN~\cite{ladhak2020wikilingua,gehrmann2021gem} & sampled\_test & 3000 & 0.0 & \rurl{hf.co/datasets/GEM/wiki_lingua} \\
    \midrule
    bAbI ~\cite{weston2015towards} & test & 19000 & 0.0 & \rurl{hf.co/datasets/Muennighoff/babi} \\
    \bottomrule
  \end{tabular}}
  \label{tab:eval}
\end{table}

\FloatBarrier

\section{Downstream Repetition Results}
\label{sec:downrep}

In Tables~\ref{tab:2b8rep}-\ref{tab:8b7oscarrep} we report downstream results of all models trained on C4~\cite{raffel2020exploring} and OSCAR~\cite{ortiz-suarez-etal-2020-monolingual} according to the configurations in \autoref{fig:validation}. All scores are from the final checkpoints at the end of training. OSCAR is a noisier dataset than C4 due to less filtering, thus models trained on C4 generally perform better. Notably, models trained on C4 completely fail on bAbI~\cite{weston2015towards}, while OSCAR models are able to perform better than random. This is likely due to code data being present in OSCAR, which enables state-tracking capabilities like for code augmented models in \autoref{sec:beyond}. For C4 the creators strictly removed all data that resembles code \cite{raffel2020exploring}. There are no significant differences between models trained for a single epoch and models trained for up to 4 epochs. Even models trained for more epochs (and thus on less unique data) have similar performance.

\begin{table}[htbp]
  \centering
  \caption{\textbf{Results for 2.8B parameter models trained on repeated data on C4 for 55B total tokens.} Scores are normalized averages of 0-5 few-shots and reported as percentages. We report mean/std. err. across five different models, each trained with a different random seed.}
  \resizebox{\textwidth}{!}{
    \begin{tabular}{l|llllllll}
\toprule
Data Budget & 55B & 28B & 18B & 14B & 11B & 9B & 4B & 1.25B \\
\midrule
Epochs & 1 & 2 & 3 & 4 & 5 & 7 & 14 & 44 \\
\midrule
ANLI R1 & 0.4 ± 1.6 & 0.7 ± 0.8 & 0.3 ± 0.5 & -0.3 ± 1.8 & 0.4 ± 1.8 & 0.4 ± 0.7 & 0.0 ± 0.9 & -0.6 ± 0.6 \\
ANLI R2 & 0.9 ± 0.4 & 1.4 ± 0.8 & 0.8 ± 0.8 & 1.1 ± 0.7 & 0.5 ± 0.7 & 0.6 ± 1.0 & 1.1 ± 1.1 & 2.7 ± 1.6 \\
ANLI R3 & 1.7 ± 0.5 & 1.2 ± 0.4 & 0.4 ± 0.5 & 1.9 ± 0.7 & 0.6 ± 1.0 & 0.8 ± 0.8 & 1.7 ± 0.7 & 0.7 ± 1.7 \\
ARC-Challenge & 1.6 ± 1.0 & 0.9 ± 0.5 & 1.2 ± 0.6 & 1.1 ± 0.6 & 1.1 ± 1.2 & 1.3 ± 0.5 & 0.3 ± 0.6 & -2.9 ± 1.0 \\
ARC-Easy & 44.5 ± 0.5 & 44.9 ± 0.4 & 44.7 ± 0.7 & 44.3 ± 0.4 & 44.0 ± 0.5 & 44.2 ± 0.9 & 41.4 ± 0.2 & 28.9 ± 0.7 \\
BoolQ & 18.8 ± 3.4 & 16.2 ± 5.2 & 16.1 ± 2.7 & 19.7 ± 1.8 & 15.0 ± 3.8 & 16.9 ± 3.2 & 13.1 ± 4.9 & -2.1 ± 4.7 \\
CB & 20.0 ± 4.7 & 17.4 ± 6.4 & 14.6 ± 5.1 & 17.5 ± 4.2 & 12.3 ± 12.2 & 14.4 ± 7.5 & 21.6 ± 8.4 & 21.3 ± 5.6 \\
COPA & 49.7 ± 3.5 & 50.3 ± 3.4 & 49.9 ± 2.3 & 50.1 ± 2.5 & 50.9 ± 1.2 & 48.1 ± 2.4 & 43.5 ± 3.1 & 33.3 ± 1.9 \\
HellaSwag & 24.7 ± 0.3 & 24.6 ± 0.2 & 24.3 ± 0.1 & 24.3 ± 0.0 & 24.3 ± 0.3 & 24.1 ± 0.1 & 22.8 ± 0.2 & 16.7 ± 0.4 \\
PiQA & 47.9 ± 0.6 & 47.6 ± 0.8 & 47.3 ± 0.3 & 47.6 ± 0.6 & 47.6 ± 0.7 & 47.0 ± 0.2 & 45.6 ± 0.5 & 37.0 ± 0.4 \\
RTE & 5.1 ± 4.0 & 2.5 ± 4.5 & 8.4 ± 2.6 & 6.0 ± 2.5 & 5.1 ± 1.6 & 2.3 ± 3.9 & 7.8 ± 2.5 & 2.6 ± 4.3 \\
SciQ & 83.2 ± 0.6 & 82.5 ± 0.6 & 82.7 ± 1.1 & 81.9 ± 0.6 & 81.9 ± 0.8 & 81.6 ± 0.9 & 78.5 ± 1.1 & 59.3 ± 1.6 \\
StoryCloze 2016 & 58.7 ± 0.2 & 58.7 ± 0.5 & 58.5 ± 0.3 & 58.3 ± 0.3 & 58.5 ± 0.6 & 58.4 ± 0.3 & 56.7 ± 0.5 & 52.0 ± 0.6 \\
WinoGrande XL & 11.6 ± 0.8 & 10.8 ± 1.1 & 10.9 ± 1.3 & 10.6 ± 0.5 & 11.1 ± 0.9 & 10.6 ± 0.9 & 6.4 ± 1.3 & 2.9 ± 1.3 \\
\midrule
E2E NLG & 17.0 ± 1.4 & 17.7 ± 0.5 & 17.0 ± 1.2 & 16.9 ± 1.1 & 15.1 ± 2.3 & 13.3 ± 2.2 & 14.9 ± 0.9 & 9.8 ± 0.9 \\
XSUM & 2.4 ± 0.1 & 2.4 ± 0.1 & 2.5 ± 0.1 & 2.3 ± 0.2 & 2.4 ± 0.1 & 2.4 ± 0.1 & 2.1 ± 0.1 & 1.6 ± 0.1 \\
WebNLG EN & 5.3 ± 0.1 & 5.5 ± 0.2 & 5.4 ± 0.1 & 5.4 ± 0.1 & 5.1 ± 0.1 & 5.4 ± 0.2 & 5.1 ± 0.3 & 2.9 ± 0.2 \\
WikiLingua EN & 3.0 ± 0.1 & 3.1 ± 0.1 & 2.9 ± 0.1 & 2.9 ± 0.3 & 2.9 ± 0.2 & 2.9 ± 0.1 & 2.6 ± 0.1 & 2.0 ± 0.2 \\
\midrule
bAbI & 0.0 ± 0.0 & 0.0 ± 0.0 & 0.0 ± 0.0 & 0.0 ± 0.0 & 0.0 ± 0.0 & 0.0 ± 0.0 & 0.0 ± 0.0 & 0.0 ± 0.0 \\
\midrule
Average & 20.9 ± 0.4 & 20.4 ± 0.3 & 20.4 ± 0.2 & 20.6 ± 0.2 & 19.9 ± 0.9 & 19.7 ± 0.2 & 19.2 ± 0.5 & 14.1 ± 0.4 \\
\bottomrule
  \end{tabular}
  }
  \label{tab:2b8rep}
\end{table}

\begin{table}[htbp]
  \centering
  \caption{\textbf{Results for 2.8B parameter models trained on repeated data on OSCAR for 55B total tokens.} Scores are normalized averages of 0-5 few-shots and reported as percentages. We report mean/std. err. across five different models, each trained with a different random seed.}
  \resizebox{\textwidth}{!}{
    \begin{tabular}{l|llllllll}
\toprule    
Data Budget & 55B & 28B & 18B & 14B & 11B & 9B & 4B & 1.25B \\
\midrule
Epochs & 1 & 2 & 3 & 4 & 5 & 7 & 14 & 44 \\
\midrule
ANLI R1 & -0.3 ± 0.5 & -0.6 ± 1.3 & 0.2 ± 0.6 & 0.3 ± 1.1 & 0.2 ± 1.2 & -0.1 ± 1.1 & -0.1 ± 0.5 & -0.7 ± 1.3 \\
ANLI R2 & 1.0 ± 1.0 & 1.1 ± 0.3 & 1.7 ± 0.8 & 2.3 ± 0.8 & 1.4 ± 1.0 & 0.8 ± 0.7 & 1.0 ± 0.7 & 2.3 ± 0.7 \\
ANLI R3 & 0.4 ± 0.8 & 0.5 ± 0.5 & -0.2 ± 0.8 & -0.1 ± 1.0 & 1.1 ± 0.6 & 0.7 ± 0.4 & -0.2 ± 0.9 & 0.5 ± 1.2 \\
ARC-Challenge & -1.4 ± 0.8 & -0.6 ± 0.8 & -1.7 ± 0.1 & -1.6 ± 0.7 & -1.6 ± 0.6 & -1.4 ± 0.5 & -1.9 ± 0.8 & -5.0 ± 1.1 \\
ARC-Easy & 39.7 ± 0.3 & 39.6 ± 0.8 & 39.5 ± 0.6 & 39.3 ± 0.5 & 38.7 ± 0.6 & 38.7 ± 0.4 & 36.9 ± 0.4 & 25.4 ± 0.7 \\
BoolQ & 12.8 ± 4.4 & 7.8 ± 3.8 & 7.9 ± 3.8 & 3.3 ± 5.4 & 0.2 ± 3.0 & 2.3 ± 6.1 & -2.1 ± 2.4 & 7.4 ± 6.1 \\
CB & 19.7 ± 5.1 & 15.4 ± 7.3 & 13.2 ± 5.1 & 12.6 ± 2.6 & 21.7 ± 3.6 & 15.4 ± 3.7 & 16.2 ± 5.2 & 9.7 ± 5.7 \\
COPA & 42.7 ± 2.2 & 39.5 ± 2.2 & 40.9 ± 2.0 & 41.5 ± 2.1 & 38.5 ± 2.4 & 40.4 ± 2.4 & 38.6 ± 2.6 & 28.5 ± 3.1 \\
HellaSwag & 16.3 ± 0.1 & 16.3 ± 0.2 & 16.3 ± 0.2 & 16.1 ± 0.2 & 16.0 ± 0.1 & 15.9 ± 0.2 & 15.0 ± 0.2 & 11.7 ± 0.1 \\
PiQA & 41.2 ± 0.7 & 41.4 ± 0.5 & 40.3 ± 0.4 & 40.6 ± 0.5 & 40.3 ± 0.9 & 39.8 ± 0.6 & 38.8 ± 1.1 & 31.0 ± 0.4 \\
RTE & 3.9 ± 1.1 & 2.1 ± 1.6 & 2.3 ± 3.3 & 1.6 ± 3.0 & 0.5 ± 2.1 & 2.9 ± 2.5 & 0.9 ± 3.4 & -3.2 ± 2.7 \\
SciQ & 83.2 ± 0.6 & 82.4 ± 0.6 & 82.1 ± 0.9 & 82.6 ± 0.7 & 81.5 ± 0.9 & 80.5 ± 0.6 & 76.5 ± 1.3 & 57.7 ± 1.8 \\
StoryCloze 2016 & 52.8 ± 0.3 & 52.9 ± 0.4 & 52.6 ± 0.3 & 53.0 ± 0.4 & 52.3 ± 0.4 & 52.4 ± 0.4 & 51.8 ± 0.7 & 47.9 ± 0.5 \\
WinoGrande XL & 5.8 ± 0.9 & 4.4 ± 1.4 & 4.5 ± 0.3 & 4.2 ± 1.3 & 4.5 ± 0.6 & 4.1 ± 0.7 & 1.7 ± 1.2 & 0.8 ± 1.3 \\
\midrule
E2E NLG & 20.3 ± 0.3 & 19.9 ± 0.5 & 19.9 ± 0.7 & 20.9 ± 0.9 & 19.7 ± 0.7 & 20.4 ± 0.6 & 19.1 ± 0.8 & 14.2 ± 0.7 \\
XSUM & 3.0 ± 0.1 & 2.9 ± 0.0 & 2.9 ± 0.3 & 2.9 ± 0.2 & 2.9 ± 0.1 & 2.8 ± 0.3 & 2.6 ± 0.2 & 1.8 ± 0.1 \\
WebNLG EN & 8.8 ± 0.4 & 8.3 ± 0.6 & 8.5 ± 0.3 & 8.4 ± 0.6 & 8.1 ± 0.2 & 8.2 ± 0.2 & 7.2 ± 0.3 & 3.3 ± 0.3 \\
WikiLingua EN & 2.9 ± 0.1 & 3.1 ± 0.2 & 3.1 ± 0.1 & 3.0 ± 0.1 & 3.1 ± 0.1 & 3.2 ± 0.3 & 2.7 ± 0.2 & 1.7 ± 0.2 \\
\midrule
bAbI & 15.5 ± 1.0 & 15.7 ± 1.1 & 15.3 ± 0.8 & 15.1 ± 1.5 & 15.9 ± 1.1 & 16.2 ± 0.9 & 14.3 ± 0.6 & 6.6 ± 0.6 \\
\midrule
Average & 19.4 ± 0.5 & 18.5 ± 0.2 & 18.4 ± 0.4 & 18.2 ± 0.4 & 18.2 ± 0.4 & 18.1 ± 0.4 & 16.8 ± 0.5 & 12.7 ± 0.7 \\
\bottomrule
  \end{tabular}
  }
  \label{tab:2b8oscarrep}
\end{table}

\begin{table}[htbp]
  \centering
  \caption{\textbf{Results for 4.2B parameter models trained on repeated data on C4 for 84B total tokens.} Scores are normalized averages of 0-5 few-shots and reported as percentages. We report mean/std. err. across five different models, each trained with a different random seed.}
  \resizebox{\textwidth}{!}{
    \begin{tabular}{l|llllllll}
\toprule    
Unique Tokens & 84B & 42B & 28B & 21B & 17B & 12B & 6B & 1.9B \\
\midrule
Epochs & 1 & 2 & 3 & 4 & 5 & 7 & 14 & 44 \\
\midrule
ANLI R1 & -1.0 ± 0.3 & -0.7 ± 1.1 & -0.7 ± 1.0 & -0.4 ± 1.1 & 0.4 ± 0.8 & 0.5 ± 1.1 & 0.1 ± 0.9 & 0.2 ± 0.9 \\
ANLI R2 & 0.8 ± 0.5 & 0.8 ± 0.8 & 0.0 ± 1.4 & 0.5 ± 0.7 & 0.5 ± 0.9 & 0.3 ± 1.0 & 0.7 ± 0.7 & 2.5 ± 1.0 \\
ANLI R3 & 1.1 ± 0.7 & 0.8 ± 0.9 & 0.3 ± 0.8 & 1.4 ± 1.1 & 1.3 ± 0.9 & 2.3 ± 0.2 & 1.3 ± 0.2 & 1.6 ± 1.2 \\
ARC-Challenge & 5.3 ± 0.6 & 5.1 ± 0.9 & 5.2 ± 2.0 & 6.0 ± 0.8 & 4.7 ± 0.8 & 3.1 ± 0.4 & 2.9 ± 1.0 & -1.3 ± 1.0 \\
ARC-Easy & 49.2 ± 0.9 & 50.4 ± 1.2 & 47.4 ± 4.5 & 49.4 ± 0.7 & 48.7 ± 1.5 & 44.9 ± 0.7 & 45.0 ± 1.2 & 31.9 ± 0.9 \\
BoolQ & 18.2 ± 4.0 & 19.6 ± 5.1 & 22.1 ± 1.0 & 20.4 ± 3.6 & 18.4 ± 6.0 & 18.4 ± 3.9 & 18.9 ± 2.6 & -3.3 ± 7.1 \\
CB & 12.0 ± 7.2 & 8.5 ± 9.2 & 7.9 ± 10.4 & 19.6 ± 7.3 & 17.8 ± 7.3 & 15.1 ± 5.8 & 17.5 ± 3.5 & 19.5 ± 6.6 \\
COPA & 59.1 ± 5.4 & 57.7 ± 3.5 & 56.7 ± 2.0 & 55.5 ± 2.4 & 56.8 ± 1.8 & 58.9 ± 1.7 & 48.7 ± 3.3 & 34.9 ± 3.4 \\
HellaSwag & 27.8 ± 4.8 & 30.2 ± 0.5 & 29.8 ± 0.9 & 29.9 ± 0.7 & 28.5 ± 1.1 & 29.0 ± 0.5 & 27.0 ± 1.2 & 19.7 ± 0.5 \\
PiQA & 50.6 ± 0.5 & 50.8 ± 0.5 & 48.6 ± 3.4 & 50.9 ± 0.7 & 50.3 ± 1.3 & 49.5 ± 0.4 & 47.6 ± 1.2 & 39.5 ± 1.3 \\
RTE & 5.6 ± 3.1 & 2.6 ± 3.9 & 7.2 ± 2.7 & 7.0 ± 3.2 & 8.8 ± 5.3 & 9.3 ± 3.6 & 3.0 ± 4.3 & 2.6 ± 4.2 \\
SciQ & 84.6 ± 3.9 & 86.1 ± 1.3 & 84.4 ± 3.7 & 85.9 ± 0.7 & 86.2 ± 0.8 & 79.0 ± 0.7 & 81.1 ± 1.4 & 65.3 ± 1.1 \\
StoryCloze 2016 & 61.1 ± 3.7 & 62.6 ± 0.2 & 61.9 ± 2.2 & 62.6 ± 0.4 & 61.8 ± 0.8 & 61.5 ± 0.8 & 60.1 ± 0.7 & 53.9 ± 0.5 \\
WinoGrande XL & 17.0 ± 2.6 & 17.8 ± 1.4 & 16.5 ± 1.8 & 17.1 ± 1.8 & 14.9 ± 1.5 & 15.9 ± 1.2 & 11.8 ± 1.5 & 3.9 ± 0.8 \\
\midrule
E2E NLG & 18.2 ± 1.2 & 18.8 ± 0.8 & 17.8 ± 1.5 & 16.0 ± 2.2 & 15.9 ± 2.5 & 13.8 ± 1.3 & 15.7 ± 0.9 & 11.2 ± 1.4 \\
XSUM & 2.9 ± 0.2 & 3.0 ± 0.2 & 2.8 ± 0.3 & 2.9 ± 0.2 & 2.9 ± 0.2 & 1.0 ± 0.4 & 2.4 ± 0.1 & 1.8 ± 0.1 \\
WebNLG EN & 4.8 ± 2.0 & 5.7 ± 0.2 & 5.4 ± 0.3 & 5.6 ± 0.2 & 5.4 ± 0.5 & 5.5 ± 0.1 & 5.4 ± 0.2 & 3.4 ± 0.3 \\
WikiLingua EN & 3.3 ± 0.5 & 3.6 ± 0.1 & 3.4 ± 0.1 & 3.4 ± 0.1 & 3.3 ± 0.1 & 1.4 ± 0.6 & 3.0 ± 0.1 & 2.2 ± 0.1 \\
\midrule
bAbI & 0.0 ± 0.0 & 0.0 ± 0.0 & 0.0 ± 0.0 & 0.0 ± 0.0 & 0.0 ± 0.0 & 0.0 ± 0.0 & 0.0 ± 0.0 & 0.0 ± 0.0 \\
\midrule
Average & 22.1 ± 1.7 & 22.3 ± 0.9 & 21.9 ± 1.2 & 22.8 ± 0.5 & 22.5 ± 0.6 & 21.6 ± 0.5 & 20.6 ± 0.6 & 15.2 ± 1.0 \\
\bottomrule
  \end{tabular}
  }
  \label{tab:4b2rep}
\end{table}

\begin{table}
  \centering
  \caption{\textbf{Results for 4.2B parameter models trained on repeated data on OSCAR for 84B total tokens.} Scores are normalized averages of 0-5 few-shots and reported as percentages. We report mean/std. err. across five different models, each trained with a different random seed.}
  \resizebox{\textwidth}{!}{
    \begin{tabular}{l|llllllll}
\toprule    
Unique Tokens & 84B & 42B & 28B & 21B & 17B & 12B & 6B & 1.9B \\
\midrule
Epochs & 1 & 2 & 3 & 4 & 5 & 7 & 14 & 44 \\
\midrule
ANLI R1 & -0.9 ± 0.5 & -0.8 ± 1.1 & -0.9 ± 1.4 & -0.4 ± 0.4 & -0.1 ± 1.2 & 0.3 ± 1.1 & -0.5 ± 0.8 & 1.1 ± 1.3 \\
ANLI R2 & 0.7 ± 0.9 & 0.7 ± 1.1 & 1.3 ± 1.0 & 1.5 ± 1.1 & 1.7 ± 1.3 & 0.9 ± 0.8 & 0.9 ± 1.0 & 1.7 ± 1.3 \\
ANLI R3 & 0.4 ± 0.6 & 0.6 ± 0.4 & 0.7 ± 0.3 & 0.4 ± 0.8 & 0.7 ± 1.2 & 0.6 ± 1.2 & 0.7 ± 0.5 & 0.8 ± 1.2 \\
ARC-Challenge & 1.3 ± 0.5 & 1.8 ± 0.5 & 1.6 ± 0.7 & 2.4 ± 1.1 & 1.6 ± 0.7 & 2.0 ± 0.7 & 1.6 ± 0.5 & -2.1 ± 0.5 \\
ARC-Easy & 45.5 ± 0.8 & 45.1 ± 1.2 & 44.8 ± 0.9 & 44.8 ± 0.6 & 45.0 ± 1.0 & 43.9 ± 0.7 & 40.7 ± 0.7 & 28.0 ± 0.9 \\
BoolQ & 14.5 ± 1.9 & 15.1 ± 4.6 & 10.8 ± 5.1 & 12.5 ± 1.9 & 6.7 ± 4.0 & 10.1 ± 4.2 & -0.0 ± 6.9 & -4.3 ± 7.2 \\
CB & 21.3 ± 2.3 & 19.2 ± 3.8 & 12.9 ± 6.4 & 16.9 ± 3.4 & 15.1 ± 9.4 & 17.8 ± 3.6 & 15.0 ± 8.1 & 11.2 ± 4.1 \\
COPA & 43.1 ± 3.0 & 42.5 ± 3.7 & 44.4 ± 1.1 & 43.0 ± 3.4 & 41.8 ± 2.3 & 44.6 ± 2.7 & 40.3 ± 3.0 & 34.9 ± 4.9 \\
HellaSwag & 21.1 ± 0.2 & 21.0 ± 0.2 & 20.9 ± 0.1 & 20.7 ± 0.2 & 20.5 ± 0.3 & 20.3 ± 0.1 & 19.3 ± 0.1 & 14.5 ± 0.2 \\
PiQA & 45.3 ± 0.9 & 44.8 ± 0.7 & 44.8 ± 0.9 & 44.4 ± 0.6 & 44.3 ± 0.6 & 43.9 ± 0.5 & 42.2 ± 0.9 & 34.0 ± 0.8 \\
RTE & 4.2 ± 2.8 & 1.5 ± 2.4 & -1.1 ± 3.9 & -2.5 ± 3.9 & 5.3 ± 1.8 & 4.4 ± 1.9 & 1.6 ± 2.2 & -1.0 ± 2.4 \\
SciQ & 86.6 ± 0.7 & 86.5 ± 0.5 & 86.0 ± 0.2 & 86.3 ± 1.0 & 85.4 ± 0.8 & 84.7 ± 0.4 & 82.0 ± 1.4 & 62.9 ± 2.5 \\
StoryCloze 2016 & 56.5 ± 0.6 & 56.8 ± 0.6 & 56.5 ± 0.7 & 55.8 ± 0.3 & 55.9 ± 0.2 & 56.0 ± 0.3 & 54.5 ± 0.7 & 49.3 ± 0.2 \\
WinoGrande XL & 9.7 ± 1.4 & 9.0 ± 1.8 & 9.5 ± 0.7 & 8.9 ± 1.0 & 7.8 ± 1.2 & 7.4 ± 1.4 & 6.8 ± 1.4 & 2.1 ± 1.0 \\
\midrule
E2E NLG & 21.4 ± 1.3 & 21.9 ± 0.4 & 21.2 ± 1.0 & 21.8 ± 0.6 & 21.0 ± 0.9 & 20.5 ± 0.7 & 20.9 ± 1.0 & 16.0 ± 0.6 \\
XSUM & 3.6 ± 0.2 & 3.5 ± 0.2 & 3.5 ± 0.2 & 3.5 ± 0.2 & 3.5 ± 0.3 & 3.2 ± 0.5 & 3.0 ± 0.2 & 1.9 ± 0.1 \\
WebNLG EN & 9.9 ± 0.4 & 9.7 ± 0.8 & 9.3 ± 0.6 & 9.7 ± 0.5 & 9.3 ± 0.7 & 9.4 ± 0.3 & 8.9 ± 0.5 & 3.8 ± 0.4 \\
WikiLingua EN & 3.9 ± 0.1 & 3.8 ± 0.2 & 3.6 ± 0.3 & 3.7 ± 0.2 & 3.6 ± 0.2 & 3.7 ± 0.1 & 3.3 ± 0.2 & 2.1 ± 0.2 \\
\midrule
bAbI & 15.0 ± 7.5 & 19.0 ± 1.2 & 18.8 ± 1.4 & 18.5 ± 1.4 & 19.2 ± 0.6 & 18.1 ± 1.4 & 14.5 ± 1.5 & 9.6 ± 1.7 \\
\midrule
Average & 21.2 ± 0.2 & 21.1 ± 0.4 & 20.4 ± 0.3 & 20.6 ± 0.5 & 20.4 ± 0.5 & 20.6 ± 0.2 & 18.7 ± 0.5 & 14.0 ± 0.6 \\
\bottomrule
  \end{tabular}
  }
  \label{tab:4b2oscarrep}
\end{table}

\begin{table}
  \centering
  \caption{\textbf{Results for 8.7B parameter models trained on repeated data on C4 for 178B total tokens and a data-constrained compute-optimal 6.3B model.} Scores are normalized averages of 0-5 few-shots and reported as percentages. The two models with 25 billion unique tokens are the ones depicted in \autoref{fig:returnalloc} (right). The data-constrained compute-optimal variant (6.3 billion parameters) performs better by using fewer parameters and repeating more data.}
  \resizebox{0.8\textwidth}{!}{
    \begin{tabular}{l|llllllll|l}
\toprule
Parameters & \multicolumn{8}{c|}{8.7B} & 6.3B \\
\midrule
Unique Tokens & 178B & 88B & 58B & 44B & 35B & 25B & 13B & 4B & 25B \\
\midrule
Epochs & 1 & 2 & 3 & 4 & 5 & 7 & 14 & 44 & 9.7 \\
\midrule
ANLI R1 & -0.9 & -1.2 & -4.2 & 0.7 & -1.3 & 0.1 & 1.2 & 2.1 & -0.9 \\
ANLI R2 & -0.4 & -1.2 & -0.2 & 0.2 & -0.4 & -0.1 & 0.4 & 2.2 & 1.0\\
ANLI R3 & 0.7 & 0.5 & 0.7 & 1.8 & 0.4 & 1.6 & 2.0 & 4.0 & 2.6 \\
ARC-Challenge & 12.2 & 11.9 & 10.5 & 12.2 & 10.6 & 11.8 & 8.3 & 2.2 & 12.7\\
ARC-Easy & 58.5 & 58.0 & 56.9 & 57.4 & 56.7 & 58.5 & 52.9 & 37.4 & 57.2\\
BoolQ & 26.1 & 31.8 & 31.3 & 30.3 & 28.8 & 28.5 & 27.9 & 4.1 & 30.6\\
CB & 7.6 & 12.9 & -15.2 & 17.9 & 14.3 & -22.8 & -12.1 & 17.4 & 6.2\\
COPA & 68.0 & 64.7 & 62.3 & 66.3 & 63.3 & 70.0 & 57.0 & 45.0 & 66.0 \\
HellaSwag & 37.8 & 37.8 & 37.3 & 37.4 & 37.1 & 37.5 & 36.1 & 27.5 & 38.1\\
PiQA & 55.9 & 55.6 & 54.7 & 56.5 & 55.8 & 53.9 & 52.4 & 45.7 & 54.3\\
RTE & 14.1 & 11.4 & 11.0 & 8.7 & 15.9 & -2.6 & -1.8 & -3.2 & 7.7\\
SciQ & 90.4 & 91.1 & 90.7 & 90.0 & 89.8 & 89.8 & 87.9 & 72.9 & 90.3\\
StoryCloze 2016 & 68.3 & 67.3 & 67.2 & 67.6 & 67.8 & 66.8 & 66.2 & 58.9 & 68.4\\
WinoGrande XL & 26.3 & 27.7 & 26.5 & 29.0 & 26.1 & 23.5 & 18.1 & 10.0 & 27.0\\
\midrule
E2E NLG & 20.5 & 17.9 & 18.7 & 20.0 & 17.2 & 17.7 & 17.4 & 11.2 & 16.9\\
XSUM & 3.6 & 3.3 & 3.8 & 3.8 & 3.5 & 3.0 & 3.3 & 2.0 & 3.8\\
WebNLG EN & 5.3 & 5.8 & 5.9 & 5.6 & 5.8 & 5.2 & 5.7 & 4.9 & 5.3\\
WikiLingua EN & 4.1 & 4.2 & 4.2 & 4.1 & 4.2 & 4.0 & 3.5 & 2.7 & 4.0 \\
\midrule
bAbI & 0.0 & 0.0 & 0.0 & 0.0 & 0.0 & 0.0 & 0.0 & 0.0 & 0.0\\
\midrule
Average & 26.2 & 26.3 & 24.3 & 26.8 & 26.1 & 23.5 & 22.4 & 18.3 & 25.9\\
\bottomrule
  \end{tabular}
  }
  \label{tab:8b7rep}
\end{table}

\begin{table}
  \centering
  \caption{\textbf{Results for 8.7B parameter models trained on repeated data on OSCAR for 178B total tokens.} Scores are normalized averages of 0-5 few-shots and reported as percentages.}
  \resizebox{0.8\textwidth}{!}{
    \begin{tabular}{l|llllllll}
\toprule
Unique Tokens & 178B & 88B & 58B & 44B & 35B & 25B & 13B & 4B \\
\midrule
Epochs & 1 & 2 & 3 & 4 & 5 & 7 & 14 & 44 \\
\midrule
ANLI R1 & -1.3 & -2.3 & -0.5 & -1.8 & 0.1 & -0.3 & 2.6 & -0.4 \\
ANLI R2 & 0.8 & 3.2 & -0.2 & -1.3 & 1.0 & 0.2 & 1.5 & 0.5 \\
ANLI R3 & 1.1 & 1.2 & 1.3 & 0.9 & 2.8 & -0.4 & 1.1 & -0.1 \\
ARC-Challenge & 6.9 & 6.7 & 6.9 & 3.8 & 6.6 & 4.8 & 4.0 & -0.9 \\
ARC-Easy & 50.2 & 51.6 & 51.2 & 51.0 & 51.9 & 50.8 & 47.0 & 33.0 \\
BoolQ & 18.4 & 11.7 & 19.4 & 22.4 & 17.5 & 20.8 & 7.6 & 4.1 \\
CB & 11.2 & 13.4 & 16.1 & 19.6 & 21.4 & 25.0 & 9.8 & 20.1 \\
COPA & 46.7 & 53.0 & 52.0 & 53.7 & 51.0 & 53.3 & 48.7 & 41.7 \\
HellaSwag & 27.4 & 27.2 & 26.8 & 26.8 & 27.3 & 26.7 & 25.5 & 19.6 \\
PiQA & 49.2 & 49.3 & 50.1 & 48.7 & 48.1 & 47.2 & 45.6 & 37.0 \\
RTE & -0.5 & 1.1 & 0.2 & 1.2 & 10.2 & 3.2 & -3.0 & -7.8 \\
SciQ & 88.1 & 88.0 & 88.4 & 87.9 & 87.9 & 87.4 & 86.3 & 64.6 \\
StoryCloze 2016 & 61.6 & 61.1 & 60.2 & 60.6 & 61.3 & 59.0 & 58.8 & 52.7 \\
WinoGrande XL & 17.6 & 16.3 & 15.4 & 13.7 & 13.9 & 12.8 & 10.8 & -0.6 \\
\midrule
E2E NLG & 23.3 & 24.2 & 22.2 & 22.9 & 23.1 & 22.1 & 22.9 & 16.8 \\
XSUM & 4.2 & 3.8 & 3.9 & 3.8 & 4.3 & 4.0 & 3.2 & 2.4 \\
WebNLG EN & 9.9 & 10.1 & 10.0 & 10.5 & 9.5 & 9.9 & 10.7 & 5.2 \\
WikiLingua EN & 4.3 & 4.0 & 4.1 & 3.7 & 4.3 & 4.2 & 4.0 & 2.7 \\
\midrule
bAbI & 20.4 & 20.6 & 21.7 & 21.4 & 21.1 & 21.3 & 19.4 & 10.7 \\
\midrule
Average & 23.1 & 23.4 & 23.6 & 23.7 & 24.4 & 23.8 & 21.4 & 15.9 \\
\bottomrule
  \end{tabular}
  }
  \label{tab:8b7oscarrep}
\end{table}

\FloatBarrier

\section{Detailed Code Augmentation Results}
\label{sec:code}

We report tabular results for replacing part of C4 or OSCAR with code for 4.2 billion parameter and 2.8 billion parameter models in Tables~\ref{tab:4b2py}-\ref{tab:2b8py}. We find that training on up to 50\% of Python data maintains performance on all natural language tasks while enabling huge performance gains on state-tracking (bAbI) for C4. For OSCAR gains are less clear, which is likely due to OSCAR containing code~\cite{ortiz-suarez-etal-2020-monolingual}, while code data was explicitly filtered out for C4~\cite{raffel2020exploring}.

\begin{table}[htbp]
  \centering
  \caption{\textbf{Results for code-augmentation for 4.2B parameter models.} Models trained on a mix of natural language (C4) and Python (The Stack). Scores are normalized averages of 0-5 few-shots and reported as percentages. We report mean/std. err. across five different models, each trained with a different random seed.}
  \resizebox{\textwidth}{!}{
    \begin{tabular}{l|l|lllllllllll}
\toprule    
 & \multicolumn{10}{c}{\% of Python pre-training data (remainder is C4)} \\
\midrule
Dataset ($\downarrow$) & 0 & 10 & 20 & 30 & 40 & 50 & 60 & 70 & 80 & 90 \\
\midrule
ANLI R1 & -1.0 ± 0.3 & -0.7 ± 0.6 & 0.5 ± 0.6 & -0.2 ± 1.1 & -1.5 ± 0.7 & -0.7 ± 1.1 & -1.1 ± 0.9 & -1.2 ± 1.1 & -1.4 ± 0.7 & -1.4 ± 0.6 \\
ANLI R2 & 0.8 ± 0.5 & 0.4 ± 0.8 & 0.3 ± 1.0 & 0.6 ± 0.5 & 0.5 ± 0.6 & 0.3 ± 0.6 & 0.8 ± 0.7 & 0.4 ± 0.5 & 0.1 ± 1.2 & 1.0 ± 0.3 \\
ANLI R3 & 1.1 ± 0.7 & 0.6 ± 0.5 & 0.5 ± 0.6 & 0.2 ± 0.4 & 0.3 ± 0.5 & 0.2 ± 0.5 & 0.3 ± 0.2 & -0.1 ± 0.3 & -0.0 ± 0.2 & -0.1 ± 0.2 \\
ARC-Challenge & 5.3 ± 0.6 & 6.4 ± 1.0 & 5.2 ± 2.4 & 4.3 ± 1.4 & 5.2 ± 0.8 & 5.2 ± 0.5 & 2.6 ± 0.4 & 1.7 ± 0.5 & -0.4 ± 0.4 & -3.0 ± 0.4 \\
ARC-Easy & 49.2 ± 0.9 & 52.4 ± 1.1 & 49.6 ± 3.7 & 48.1 ± 4.1 & 50.1 ± 1.0 & 49.7 ± 0.3 & 48.0 ± 0.5 & 45.6 ± 0.5 & 43.3 ± 0.4 & 37.7 ± 0.7 \\
BoolQ & 18.2 ± 4.0 & 10.5 ± 12.0 & 16.3 ± 5.2 & 17.8 ± 3.3 & 13.4 ± 3.4 & 14.8 ± 2.1 & 12.5 ± 8.9 & 12.1 ± 6.6 & 7.2 ± 6.7 & 10.7 ± 7.3 \\
CB & 12.0 ± 7.2 & 20.3 ± 2.7 & 14.4 ± 7.1 & 16.5 ± 1.2 & 22.3 ± 3.1 & 22.1 ± 4.8 & 19.4 ± 4.8 & 23.8 ± 3.3 & 23.8 ± 4.1 & 23.4 ± 2.4 \\
COPA & 59.1 ± 5.4 & 56.4 ± 4.9 & 46.7 ± 8.7 & 50.2 ± 3.7 & 52.7 ± 2.5 & 50.1 ± 4.5 & 46.5 ± 1.9 & 43.1 ± 4.2 & 39.2 ± 3.7 & 35.9 ± 4.9 \\
HellaSwag & 27.8 ± 4.8 & 29.4 ± 0.4 & 25.7 ± 4.8 & 27.0 ± 1.7 & 26.3 ± 2.4 & 26.3 ± 0.6 & 25.0 ± 0.1 & 22.6 ± 0.1 & 19.5 ± 0.2 & 14.7 ± 0.1 \\
PiQA & 50.6 ± 0.5 & 50.8 ± 0.6 & 48.6 ± 3.0 & 48.2 ± 2.8 & 48.7 ± 0.7 & 48.4 ± 1.0 & 47.1 ± 0.7 & 45.6 ± 0.3 & 43.4 ± 0.9 & 39.0 ± 0.8 \\
RTE & 5.6 ± 3.1 & 7.3 ± 3.4 & 4.4 ± 4.7 & 6.1 ± 2.6 & 9.1 ± 4.0 & 8.1 ± 5.9 & 7.7 ± 5.3 & 4.0 ± 2.1 & 6.2 ± 2.1 & 4.6 ± 2.5 \\
SciQ & 84.6 ± 3.9 & 87.1 ± 0.2 & 84.6 ± 4.8 & 86.9 ± 1.2 & 86.9 ± 1.2 & 87.9 ± 0.9 & 87.6 ± 0.6 & 87.0 ± 0.2 & 86.0 ± 0.2 & 84.5 ± 0.6 \\
StoryCloze 2016 & 61.1 ± 3.7 & 62.0 ± 0.6 & 59.0 ± 4.8 & 60.8 ± 1.5 & 59.9 ± 1.9 & 60.0 ± 0.7 & 59.0 ± 0.4 & 57.2 ± 0.5 & 54.9 ± 0.4 & 51.0 ± 0.3 \\
WinoGrande XL & 17.0 ± 2.6 & 17.4 ± 2.1 & 14.9 ± 4.4 & 15.2 ± 2.0 & 15.7 ± 1.2 & 14.2 ± 1.0 & 13.5 ± 1.3 & 10.7 ± 1.3 & 9.1 ± 0.6 & 5.3 ± 1.3 \\
\midrule
E2E NLG & 18.2 ± 1.2 & 21.8 ± 1.6 & 15.9 ± 8.6 & 23.3 ± 0.6 & 21.5 ± 3.8 & 23.9 ± 0.6 & 23.7 ± 0.6 & 23.7 ± 0.5 & 24.3 ± 0.7 & 24.0 ± 0.9 \\
XSUM & 2.9 ± 0.2 & 3.2 ± 0.5 & 3.4 ± 0.3 & 3.3 ± 0.3 & 3.6 ± 0.6 & 3.4 ± 0.2 & 3.5 ± 0.2 & 2.9 ± 0.3 & 2.8 ± 0.4 & 2.7 ± 0.2 \\
WebNLG EN & 4.8 ± 2.0 & 9.5 ± 0.7 & 10.2 ± 1.1 & 10.5 ± 0.7 & 10.4 ± 0.8 & 10.4 ± 0.6 & 9.9 ± 0.4 & 10.0 ± 0.5 & 9.3 ± 0.6 & 9.2 ± 0.2 \\
WikiLingua EN & 3.3 ± 0.5 & 4.0 ± 0.1 & 4.0 ± 0.2 & 4.2 ± 0.1 & 4.3 ± 0.3 & 4.2 ± 0.2 & 4.4 ± 0.3 & 4.1 ± 0.2 & 3.9 ± 0.2 & 3.6 ± 0.3 \\
\midrule
bAbI & 0.0 ± 0.0 & 12.5 ± 6.7 & 13.8 ± 7.2 & 15.8 ± 8.2 & 17.4 ± 9.2 & 23.2 ± 1.2 & 23.4 ± 2.0 & 24.3 ± 1.4 & 23.2 ± 1.0 & 24.6 ± 1.8 \\
\midrule
Average & 22.1 ± 1.7 & 23.7 ± 0.7 & 22.0 ± 3.0 & 23.1 ± 1.1 & 23.5 ± 1.0 & 23.8 ± 0.5 & 22.8 ± 1.0 & 22.0 ± 0.6 & 20.8 ± 0.5 & 19.3 ± 0.3 \\
Average (no bAbI) & 23.4 ± 1.8 & 24.4 ± 0.7 & 22.5 ± 2.8 & 23.5 ± 0.8 & 23.9 ± 0.9 & 23.8 ± 0.5 & 22.8 ± 1.0 & 21.8 ± 0.6 & 20.6 ± 0.6 & 19.1 ± 0.4 \\
\bottomrule
  \end{tabular}
  }
  \label{tab:4b2py}
\end{table}

\begin{table}[htbp]
  \centering
  \caption{\textbf{Results for code-augmentation for 2.8B parameter models.} Models trained on a mix of natural language (C4) and Python (The Stack). Scores are normalized averages of 0-5 few-shots and reported as percentages. We report mean/std. err. across five different models, each trained with a different random seed.}
  \resizebox{\textwidth}{!}{
    \begin{tabular}{l|l|lllll|l|lllll}
    \toprule
 & \multicolumn{6}{c|}{\% of Python pre-training data (rest is C4)} & \multicolumn{6}{c}{\% of Python pre-training data (rest is OSCAR)} \\
 \midrule
Dataset ($\downarrow$) & 0 & 10 & 20 & 30 & 40 & 50 & 0 & 10 & 20 & 30 & 40 & 50 \\
\midrule
ANLI R1 & 0.4 ± 1.6 & -1.5 & -0.9 & -1.0 & -0.7 & -2.4 & -0.3 ± 0.5 & 0.0 & -0.6 & -1.6 & -2.4 & -1.7 \\
ANLI R2 & 0.9 ± 0.4 & 0.7 & 0.0 & 0.1 & -0.1 & 0.1 & 1.0 ± 1.0 & 1.2 & -0.1 & -0.0 & 0.0 & 0.8 \\
ANLI R3 & 1.7 ± 0.5 & 0.6 & -0.7 & -0.2 & 0.4 & 0.0 & 0.4 ± 0.8 & -0.4 & -0.2 & -1.7 & -0.8 & -0.5 \\
ARC-Challenge & 1.6 ± 1.0 & 4.2 & 1.7 & 1.5 & 0.2 & -0.2 & -1.4 ± 0.8 & -0.7 & -1.4 & -3.4 & -2.3 & -3.1 \\
ARC-Easy & 44.5 ± 0.5 & 46.4 & 46.5 & 45.4 & 43.6 & 42.7 & 39.7 ± 0.3 & 39.8 & 38.7 & 39.1 & 37.3 & 37.6 \\
BoolQ & 18.8 ± 3.4 & 15.7 & 19.0 & 13.4 & 16.0 & 4.4 & 12.8 ± 4.4 & 3.3 & 12.5 & 10.6 & 5.8 & 8.5 \\
CB & 20.0 ± 4.7 & 22.8 & 10.7 & 20.5 & 17.4 & 15.2 & 19.7 ± 5.1 & 14.7 & 15.6 & 19.6 & 22.8 & 17.0 \\
COPA & 49.7 ± 3.5 & 46.3 & 49.3 & 46.3 & 42.7 & 40.0 & 42.7 ± 2.2 & 42.7 & 41.0 & 42.7 & 35.7 & 38.0 \\
HellaSwag & 24.7 ± 0.3 & 24.1 & 23.3 & 22.3 & 21.9 & 20.9 & 16.3 ± 0.1 & 15.7 & 15.9 & 15.5 & 15.1 & 13.7 \\
PiQA & 47.9 ± 0.6 & 46.9 & 47.7 & 45.1 & 46.2 & 45.5 & 41.2 ± 0.7 & 41.6 & 39.9 & 40.5 & 38.8 & 38.6 \\
RTE & 5.1 ± 4.0 & 8.8 & 7.7 & 5.1 & 7.8 & 10.8 & 3.9 ± 1.1 & 2.2 & 4.3 & 1.1 & 3.7 & -1.7 \\
SciQ & 83.2 ± 0.6 & 83.3 & 85.3 & 84.8 & 83.2 & 83.7 & 83.2 ± 0.6 & 82.4 & 83.5 & 82.8 & 83.3 & 83.2 \\
StoryCloze 2016 & 58.7 ± 0.2 & 59.3 & 57.9 & 56.9 & 56.5 & 56.0 & 52.8 ± 0.3 & 52.0 & 52.2 & 52.0 & 51.8 & 50.9 \\
WinoGrande XL & 11.6 ± 0.8 & 13.0 & 10.7 & 9.3 & 8.2 & 9.6 & 5.8 ± 0.9 & 3.2 & 5.6 & 5.8 & 4.6 & 3.9 \\
\midrule
E2E NLG & 17.0 ± 1.4 & 19.8 & 21.1 & 20.2 & 22.1 & 21.0 & 20.3 ± 0.3 & 21.9 & 20.7 & 20.5 & 20.7 & 21.1 \\
XSUM & 2.4 ± 0.1 & 2.7 & 2.0 & 2.2 & 2.0 & 2.3 & 3.0 ± 0.1 & 2.8 & 3.1 & 3.4 & 3.1 & 2.9 \\
WebNLG EN & 5.3 ± 0.1 & 9.1 & 8.0 & 8.5 & 8.5 & 9.1 & 8.8 ± 0.4 & 8.7 & 9.6 & 9.1 & 8.7 & 9.4 \\
WikiLingua EN & 3.0 ± 0.1 & 3.2 & 3.2 & 3.6 & 3.3 & 3.7 & 2.9 ± 0.1 & 3.3 & 3.6 & 3.5 & 3.4 & 3.5 \\
\midrule
bAbI & 0.0 ± 0.0 & 4.6 & 14.2 & 14.2 & 14.8 & 15.1 & 15.5 ± 1.0 & 16.6 & 17.2 & 17.2 & 17.7 & 15.9 \\
\midrule
Average & 20.9 ± 0.4 & 21.6 & 21.4 & 21.0 & 20.7 & 19.9 & 19.4 ± 0.5 & 18.5 & 19.0 & 18.8 & 18.3 & 17.8 \\
Average (without bAbI) & 22.0 ± 0.5 & 22.5 & 21.8 & 21.3 & 21.1 & 20.1 & 19.6 ± 0.5 & 18.6 & 19.1 & 18.9 & 18.3 & 17.9 \\
\bottomrule
  \end{tabular}
  }
  \label{tab:2b8py}
\end{table}

\FloatBarrier

\section{Filtering Procedure}
\label{sec:filtering}

\begin{figure*}[htbp]
    \centering
    \begin{center}
        \includegraphics[width=\textwidth]{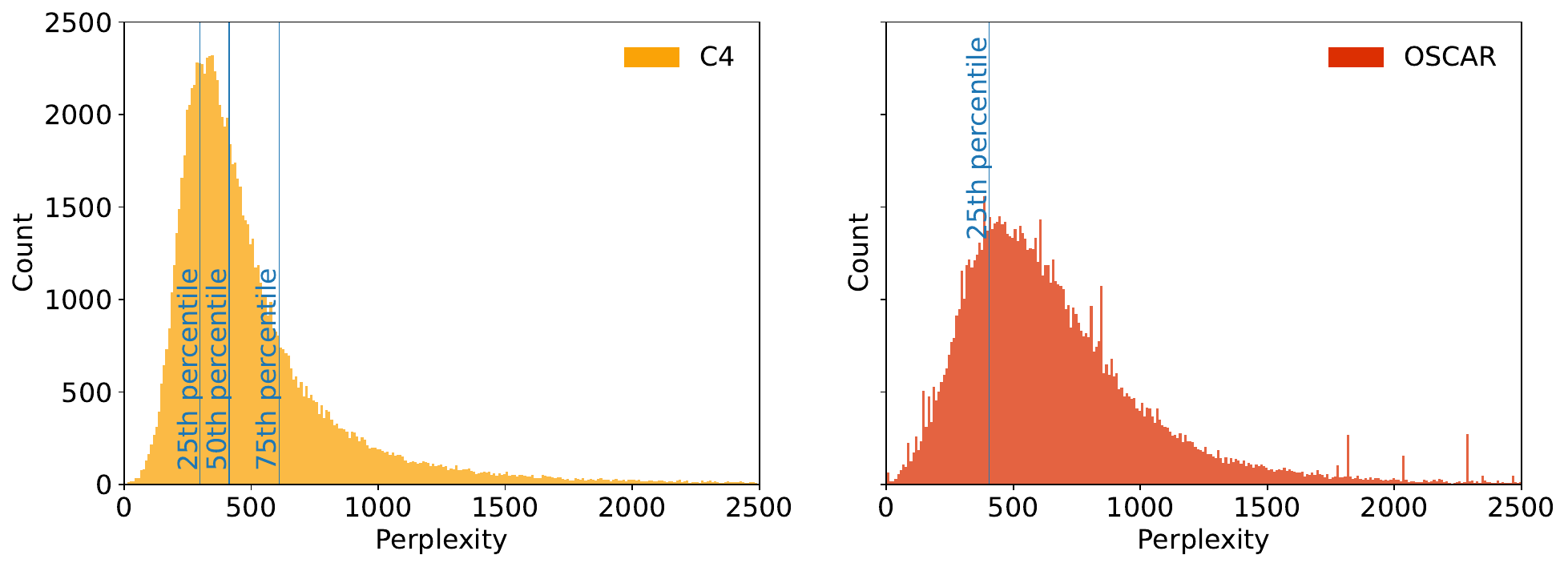}
        \caption{Perplexity histograms for respective datasets. For demonstration purposes, we use 100,000 random samples of each dataset.}
        \label{fig:perplexity_hist}
    \end{center}
\end{figure*}

\paragraph{Perplexity filtering} We follow the approach of~\cite{laurencconbigscience} to perform perplexity filtering and reuse their artifacts - a SentencePiece tokenizer~\cite{kudo-richardson-2018-sentencepiece} and a KenLM 5-gram language model~\cite{heafield-2011-kenlm} trained on Wikipedia introductions and available to download from their repository.\footnote{\url{https://github.com/bigscience-workshop/data-preparation/tree/main/preprocessing/training/01b_oscar_cleaning_and_filtering}} We compute the model's perplexity on all OSCAR and C4 samples and only select samples that fall within a certain percentile threshold. For example, to select the top 25\%, we only select samples with perplexity lower than the 25th percentile. \autoref{fig:perplexity_hist} provides a visual representation of perplexity distribution for respective datasets, highlighting the relevant percentile thresholds.

\paragraph{Deduplication} We perform deduplication leveraging the suffix array-based approach proposed by \citet{deduplicatinglee2021}. We remove any document with at least a 100-character span overlapping with any other document in the corpus. We deduplicate the full C4 dataset. In the case of OSCAR, the memory requirements of the deduplication procedure make performing the full dataset deduplication infeasible. Instead, we select a 25\% subset of the full OSCAR and build a suffix array for this subset. We experiment with leveraging the 25\% OSCAR suffix array in two ways. First, we deduplicate the selected subset. This is very strict and preserves less than 5\% of the full OSCAR. Subsequently, we use the 25\% suffix array to deduplicate the full OSCAR, i.e. we remove any document which has at least a 100-character span overlapping with the 25\% subset we selected. This is more permissive and allows us to preserve 31\% of the original dataset. We refer to the latter as \emph{expanded} in \autoref{tab:filt} and it is used for the training of the 4.2 billion parameter model in \autoref{tab:dedup}, while the smaller deduplicated version of OSCAR is used for the 2.8 billion parameter model.

\paragraph{ROOTS filter} In addition, we benchmark with the filtering procedure from the ROOTS corpus~\cite{laurencconbigscience}. It applies the following set of filters:
\begin{itemize}
    \item Discarding documents with too few words
    \item Discarding documents with overly repeated character- and word-n-grams
    \item Discarding documents with too many special characters
    \item Discarding documents with too few grammatical function words (e.g. ``of'', ``and'')
    \item Discarding documents with too many flagged words
    \item Discarding documents with a low \texttt{fasttext} language identification score
    \item Perplexity filtering
\end{itemize}

\begin{table*}[htbp]
    \caption{\textbf{Sizes of filtered datasets.}}
    \label{tab:filt}
\centering
\begin{tabular}{l r c}
\toprule
Base Dataset & Filter & Tokens after filtering \\
\midrule
C4 & Deduplication & 21 billion \\
C4 & Perplexity Top 25\% & 44 billion \\
C4 & Perplexity Top 50\% & 89 billion \\
C4 & Perplexity 25-75\% & 89 billion \\
OSCAR & Deduplication & 9 billion \\
OSCAR & Deduplication-expanded & 94 billion \\
OSCAR & Perplexity Top 25\% & 80 billion \\
OSCAR & ROOTS & 99 billion \\
\bottomrule
\end{tabular}
\end{table*}
\FloatBarrier

\section{Detailed Filtering Results}
\label{sec:addfilter}

In \autoref{tab:pplx}, we report detailed perplexity filtering results on C4 and OSCAR. For C4, perplexity filtering is only effective at 4.2B parameters. Meanwhile, for OSCAR, which is noisier than C4, perplexity filtering seems effective both for 2.8B and 4.2B parameters. \autoref{tab:dedup} contains deduplication results and results for the ROOTS filter. Deduplication does not improve downstream performance for C4 while being effective for OSCAR which has significantly more noise. Applying the ROOTS filter on OSCAR is not better than the unfiltered OSCAR on our benchmark, but might have other beneficial effects, such as reducing obscenity, templated messages, or repetition, depending on the final use case.

\begin{table}[htbp]
  \centering
  \caption{\textbf{Results for perplexity-filtering.} The training data is perplexity filtered according to the given percentile, e.g. 25\% corresponds to training on the top 25\% percent of examples with the lowest perplexity. The resulting dataset sizes are in \autoref{tab:filt}. The data is repeated until it matches 55B tokens for 2.8B parameter and 84B tokens for 4.2B parameter models. Scores are normalized averages of 0-5 few-shots and reported as percentages. For unfiltered models, we report mean/std. err. across five different models, each trained with a different random seed.}
  \resizebox{\textwidth}{!}{
    \begin{tabular}{l|lll|llll|ll|ll}
    \toprule    
Training Data & \multicolumn{7}{c|}{C4} & \multicolumn{4}{c}{OSCAR} \\
\midrule
Parameters & \multicolumn{3}{c|}{2.8B} & \multicolumn{4}{c|}{4.2B} & \multicolumn{2}{c|}{2.8B} & \multicolumn{2}{c|}{4.2B} \\
\midrule
Percentile & All & 25\% & 50\% & All & 25\% & 50\% & 25-75\% & All & 25\% & All & 25\% \\
\midrule
ANLI R1 & 0.4 ± 1.6 & -0.1 & 0.9 & -0.5 ± 1.4 & -0.0 & -0.7 & -0.8 & -0.3 ± 0.5 & -0.4 & -0.4 ± 1.2 & -2.2 \\
ANLI R2 & 0.9 ± 0.4 & -0.2 & -0.7 & 0.0 ± 1.3 & -0.4 & -0.0 & 1.1 & 1.0 ± 1.0 & 1.7 & 1.0 ± 0.9 & 0.7 \\
ANLI R3 & 1.7 ± 0.5 & 0.5 & 1.4 & 0.7 ± 0.5 & 0.7 & 2.9 & 0.4 & 0.4 ± 0.8 & 1.7 & 1.2 ± 0.5 & 2.1 \\
ARC-Challenge & 1.6 ± 1.0 & 3.3 & 2.9 & 4.2 ± 1.6 & 10.2 & 9.3 & 7.9 & -1.4 ± 0.8 & 3.3 & 1.8 ± 0.8 & 6.3 \\
ARC-Easy & 44.5 ± 0.5 & 47.3 & 47.7 & 48.1 ± 4.8 & 55.8 & 53.7 & 51.0 & 39.7 ± 0.3 & 46.8 & 45.7 ± 0.6 & 51.8 \\
BoolQ & 18.8 ± 3.4 & 17.1 & 17.7 & 22.4 ± 3.3 & 27.7 & 23.5 & 24.5 & 12.8 ± 4.4 & 11.8 & 12.4 ± 5.9 & 22.2 \\
CB & 20.0 ± 4.7 & 16.1 & 13.8 & 9.3 ± 16.6 & 24.6 & 22.3 & 12.5 & 19.7 ± 5.1 & 17.0 & 23.9 ± 3.8 & 20.1 \\
COPA & 49.7 ± 3.5 & 55.7 & 56.0 & 55.3 ± 3.8 & 60.7 & 66.0 & 61.0 & 42.7 ± 2.2 & 44.0 & 41.1 ± 3.0 & 49.3 \\
HellaSwag & 24.7 ± 0.3 & 24.7 & 26.0 & 29.4 ± 1.3 & 30.7 & 32.7 & 33.1 & 16.3 ± 0.1 & 19.0 & 21.0 ± 0.2 & 23.3 \\
PiQA & 47.9 ± 0.6 & 43.4 & 45.8 & 48.8 ± 3.8 & 47.9 & 52.2 & 52.1 & 41.2 ± 0.7 & 38.3 & 45.0 ± 0.6 & 44.4 \\
RTE & 5.1 ± 4.0 & 5.7 & 7.3 & 6.9 ± 3.1 & 11.9 & 2.2 & 10.3 & 3.9 ± 1.1 & -1.2 & 2.2 ± 4.3 & 7.0 \\
SciQ & 83.2 ± 0.6 & 82.4 & 82.8 & 86.3 ± 1.1 & 88.6 & 87.4 & 88.4 & 83.2 ± 0.6 & 84.0 & 86.3 ± 0.6 & 86.5 \\
StoryCloze 2016 & 58.7 ± 0.2 & 61.1 & 61.2 & 62.8 ± 0.5 & 65.5 & 65.6 & 65.1 & 52.8 ± 0.3 & 57.9 & 57.2 ± 0.6 & 60.2 \\
WinoGrande XL & 11.6 ± 0.8 & 15.3 & 14.3 & 18.7 ± 1.0 & 24.9 & 22.3 & 18.7 & 5.8 ± 0.9 & 9.7 & 10.1 ± 1.0 & 14.8 \\
\midrule
E2E NLG & 17.0 ± 1.4 & 16.1 & 16.8 & 17.9 ± 0.7 & 18.8 & 17.8 & 19.2 & 20.3 ± 0.3 & 19.5 & 21.6 ± 0.7 & 22.6 \\
XSUM & 2.4 ± 0.1 & 2.6 & 3.0 & 3.0 ± 0.3 & 3.9 & 3.2 & 3.0 & 3.0 ± 0.1 & 3.2 & 3.7 ± 0.2 & 2.7 \\
WebNLG EN & 5.3 ± 0.1 & 4.8 & 5.1 & 5.6 ± 0.3 & 5.4 & 5.7 & 5.2 & 8.8 ± 0.4 & 6.9 & 9.3 ± 0.5 & 10.6 \\
WikiLingua EN & 3.0 ± 0.1 & 3.2 & 3.3 & 3.6 ± 0.2 & 3.4 & 3.5 & 3.4 & 2.9 ± 0.1 & 3.4 & 4.0 ± 0.1 & 3.8 \\
\midrule
bAbI & 0.0 ± 0.0 & 0.0 & 0.0 & 0.0 ± 0.0 & 0.0 & 0.0 & 0.0 & 15.5 ± 1.0 & 14.5 & 19.3 ± 1.0 & 17.2 \\
\midrule
Average & 20.9 ± 0.4 & 21.0 & 21.3 & 22.2 ± 1.4 & 25.3 & 24.7 & 24.0 & 19.4 ± 0.5 & 20.1 & 21.4 ± 0.5 & 23.3 \\
\bottomrule
  \end{tabular}
  }
  \label{tab:pplx}
\end{table}

\begin{table}
  \centering
  \caption{\textbf{Results for filtering with deduplication and the ROOTS filters.} The resulting dataset sizes are in \autoref{tab:filt}. The data is repeated until it matches 55B tokens for 2.8B parameter and 84B tokens for 4.2B parameter models. Scores are normalized averages of 0-5 few-shots and reported as percentages. For unfiltered models we report mean/std. err. across five different models, each trained with a different random seed.}
  \resizebox{\textwidth}{!}{
    \begin{tabular}{l|llll|llllll}
    \toprule    
Training Data & \multicolumn{4}{c|}{C4} & \multicolumn{6}{c}{OSCAR} \\
\midrule
Parameters & \multicolumn{2}{c|}{2.8B parameters} & \multicolumn{2}{c|}{4.2B parameters} & \multicolumn{3}{c|}{2.8B parameters} & \multicolumn{3}{c|}{4.2B parameters} \\
\midrule
Method & All & Dedup. & All & Dedup. & All & Dedup. & ROOTS & All & Dedup.-exp. & ROOTS \\
\midrule
ANLI R1 & 0.4 ± 1.6 & -0.2 & -0.5 ± 1.4 & -0.8 & -0.3 ± 0.5 & -2.1 & -1.7 & -0.4 ± 1.2 & -1.8 & 1.2 \\
ANLI R2 & 0.9 ± 0.4 & 1.1 & 0.0 ± 1.3 & -0.1 & 1.0 ± 1.0 & 2.0 & 0.7 & 1.0 ± 0.9 & -0.5 & -0.3 \\
ANLI R3 & 1.7 ± 0.5 & 1.8 & 0.7 ± 0.5 & 0.4 & 0.4 ± 0.8 & 0.4 & 0.2 & 1.2 ± 0.5 & 0.8 & -0.3 \\
ARC-Challenge & 1.6 ± 1.0 & 0.6 & 4.2 ± 1.6 & 3.9 & -1.4 ± 0.8 & 2.6 & -0.9 & 1.8 ± 0.8 & 6.8 & 0.6 \\
ARC-Easy & 44.5 ± 0.5 & 43.0 & 48.1 ± 4.8 & 46.8 & 39.7 ± 0.3 & 44.6 & 42.3 & 45.7 ± 0.6 & 51.0 & 47.1 \\
BoolQ & 18.8 ± 3.4 & 1.5 & 22.4 ± 3.3 & 2.2 & 12.8 ± 4.4 & 3.4 & 13.4 & 12.4 ± 5.9 & 13.0 & 7.0 \\
CB & 20.0 ± 4.7 & 0.4 & 9.3 ± 16.6 & 0.9 & 19.7 ± 5.1 & 25.4 & 14.3 & 23.9 ± 3.8 & 25.0 & 28.1 \\
COPA & 49.7 ± 3.5 & 57.0 & 55.3 ± 3.8 & 60.0 & 42.7 ± 2.2 & 47.3 & 37.7 & 41.1 ± 3.0 & 55.3 & 43.0 \\
HellaSwag & 24.7 ± 0.3 & 25.1 & 29.4 ± 1.3 & 30.7 & 16.3 ± 0.1 & 22.8 & 17.6 & 21.0 ± 0.2 & 26.3 & 22.4 \\
PiQA & 47.9 ± 0.6 & 49.1 & 48.8 ± 3.8 & 53.4 & 41.2 ± 0.7 & 45.1 & 41.9 & 45.0 ± 0.6 & 48.5 & 46.3 \\
RTE & 5.1 ± 4.0 & 3.2 & 6.9 ± 3.1 & 0.1 & 3.9 ± 1.1 & 6.1 & 5.8 & 2.2 ± 4.3 & 1.1 & 8.9 \\
SciQ & 83.2 ± 0.6 & 80.4 & 86.3 ± 1.1 & 82.2 & 83.2 ± 0.6 & 82.6 & 83.1 & 86.3 ± 0.6 & 88.5 & 86.4 \\
StoryCloze 2016 & 58.7 ± 0.2 & 61.8 & 62.8 ± 0.5 & 65.2 & 52.8 ± 0.3 & 58.1 & 54.3 & 57.2 ± 0.6 & 61.6 & 58.6 \\
WinoGrande XL & 11.6 ± 0.8 & 13.3 & 18.7 ± 1.0 & 19.7 & 5.8 ± 0.9 & 12.7 & 5.6 & 10.1 ± 1.0 & 16.2 & 11.0 \\
\midrule
E2E NLG & 17.0 ± 1.4 & 15.6 & 17.9 ± 0.7 & 14.2 & 20.3 ± 0.3 & 20.5 & 20.5 & 21.6 ± 0.7 & 2.4 & 22.6 \\
XSUM & 2.4 ± 0.1 & 2.1 & 3.0 ± 0.3 & 2.5 & 3.0 ± 0.1 & 3.2 & 3.1 & 3.7 ± 0.2 & 4.6 & 3.8 \\
WebNLG EN & 5.3 ± 0.1 & 4.3 & 5.6 ± 0.3 & 4.4 & 8.8 ± 0.4 & 7.4 & 7.4 & 9.3 ± 0.5 & 9.7 & 9.4 \\
WikiLingua EN & 3.0 ± 0.1 & 3.2 & 3.6 ± 0.2 & 3.2 & 2.9 ± 0.1 & 3.0 & 3.1 & 4.0 ± 0.1 & 4.3 & 4.0 \\
\midrule
bAbI & 0.0 ± 0.0 & 0.0 & 0.0 ± 0.0 & 0.0 & 15.5 ± 1.0 & 17.2 & 14.3 & 19.3 ± 1.0 & 21.1 & 18.0 \\
\midrule
Average & 20.9 ± 0.4 & 19.1 & 22.2 ± 1.4 & 20.5 & 19.4 ± 0.5 & 21.2 & 19.1 & 21.4 ± 0.5 & 22.8 & 22.0 \\
\bottomrule
\end{tabular}
  }
  \label{tab:dedup}
\end{table}

\FloatBarrier

\section{Loss Curves for Complementary Strategies}
\label{sec:losscomp}

\begin{figure*}[htbp]
    \centering
    \begin{center}
        \includegraphics[width=\textwidth]{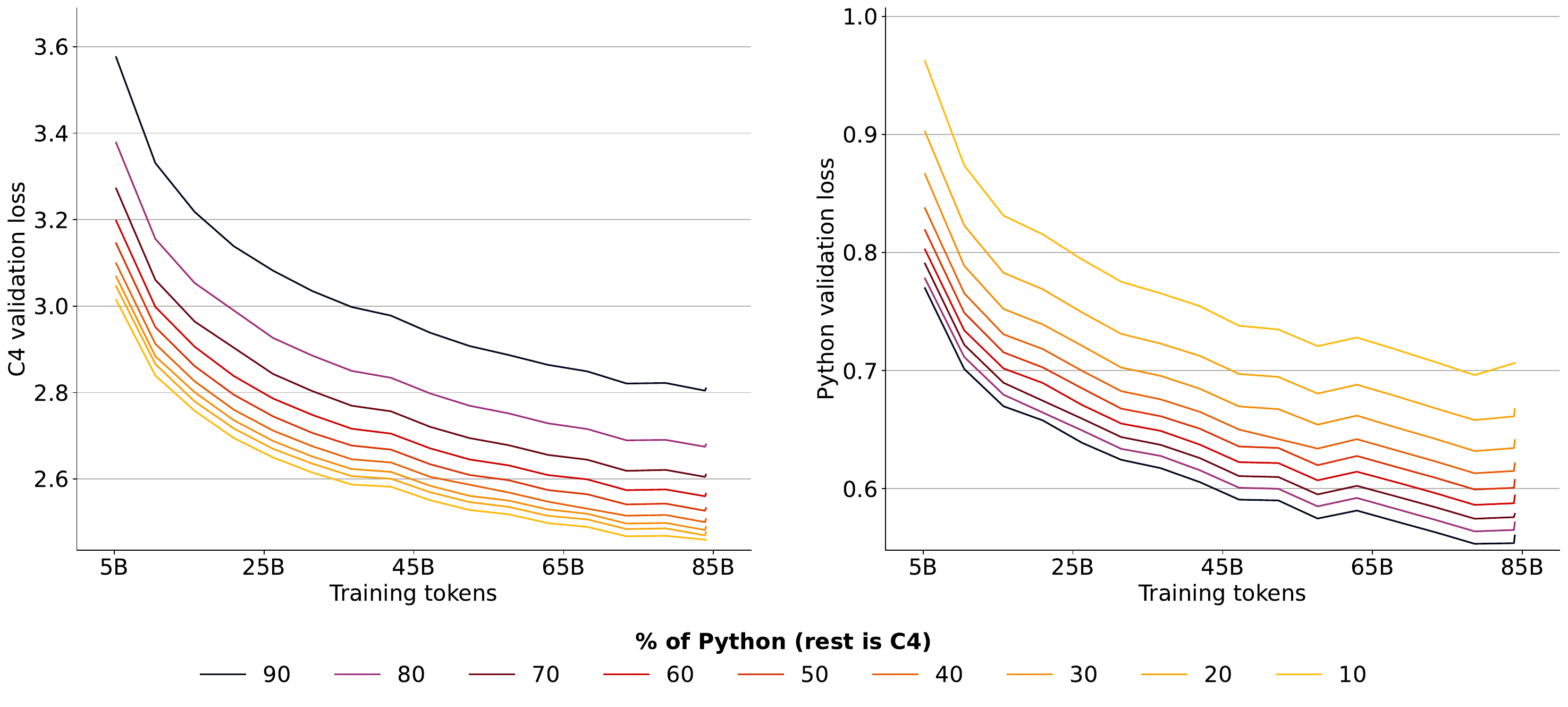}
        \caption{\textbf{Validation loss of models trained on a mix of natural language (C4) and Python data.}}
        \label{fig:valc4py}
    \end{center}
\end{figure*}

\begin{figure*}[htbp]
    \centering
    \begin{center}
        \includegraphics[width=\textwidth]{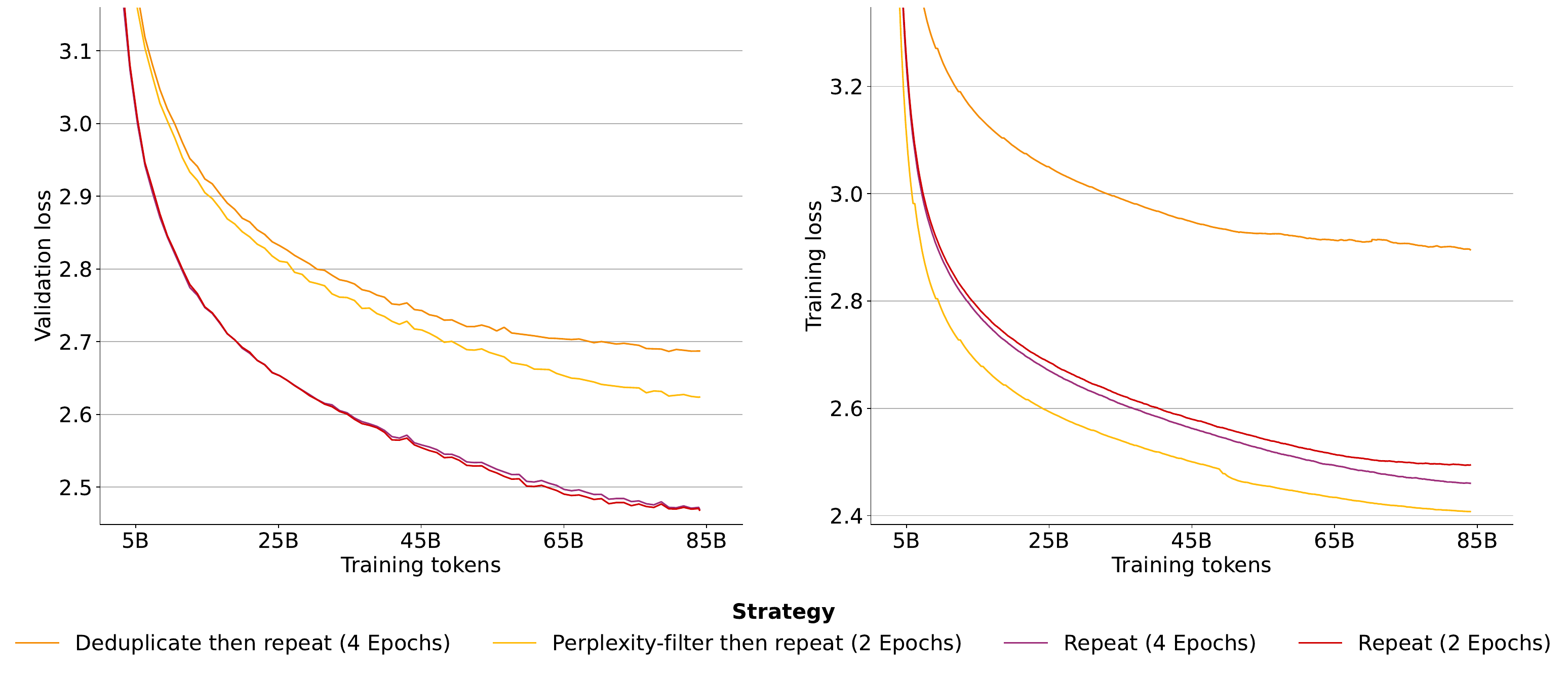}
        \caption{\textbf{Validation and training loss of models trained with different data strategies.} Training loss is smoothed with exponential moving average smoothing and a weight of 0.999. Downstream performance of the models is in \autoref{fig:beyond}.}
        \label{fig:beyondloss}
    \end{center}
\end{figure*}

To compare complementary data strategies in \autoref{sec:beyond}, we have used downstream performance on natural language tasks detailed in \autoref{sec:eval} instead of loss. This is because validation loss gives an unfair advantage to models trained on a larger fraction of data from the same distribution. For example, when making up for missing natural language data with code, models that are trained on more code will have better validation loss on code data while having worse loss on the natural language data as seen in \autoref{fig:valc4py}: The model pre-trained on 90\% of Python code data and 10\% of C4 has the highest C4 validation loss, but the lowest Python validation loss.

Models trained on deduplicated or perplexity-filtered data have higher validation loss as the held-out validation data has not gone through the same filtering steps. Thus, its distribution more closely resembles the training data of models trained on the unfiltered data resulting in worse validation loss for the two filtering strategies in \autoref{fig:beyondloss} (left). Meanwhile, for training loss in \autoref{fig:beyondloss} (right) the model trained on perplexity-filtered data has the lowest loss. Its training data has been filtered to the top 25\% of examples with the lowest perplexity (\autoref{sec:filtering}) thus high loss examples have been explicitly filtered out from the training data resulting in low training loss. The model trained on deduplicated data has the highest validation and training loss. This is because commonly repeated sequences have been filtered out from its training data. Thus, when encountering these common sequences in the unfiltered validation set, its loss is comparatively high as other models have likely simply memorized them. Similarly, fewer repeated sequences during training results in higher training loss as unseen sequences are harder to predict.

\FloatBarrier

\section{Limitations and Future Work}
\label{sec:limits}

\paragraph{Repeating fractions of the data} In this work we focus on repeating the entire unique dataset for several epochs. Alternatively, one can repeat only a fraction of the dataset. For example, repeating 10\% of the dataset for 10 epochs while repeating the rest only for a single epoch as done by \citet{hernandez2022scaling}. To predict loss in that scenario, one may need to adapt our scaling laws with an additional parameter to account for the fraction that is repeated and possibly a parameter that captures at what point in training the data is repeated. Repeating earlier in training when most model weights are still randomly initialized is likely to cause less damage than later in training. Adapting our parametric fit to make concrete scaling predictions for such scenarios is an exciting future research direction.

\paragraph{Sensitivity to hyperparameters} The returns from additional epochs may heavily depend on hyperparameters such as learning rate, dropout, or the optimizer choice. It is likely that increasing the learning rate, for example, would lead to diminishing returns from additional epochs kicking in earlier. In this work, we have fixed most hyperparameters to commonly used values for the training of LLMs and leave such explorations to future work.

\paragraph{Other datasets} The optimal data strategy is dependent on the dataset at hand and we cannot give universally applicable filtering recommendations. By looking into C4 and OSCAR, we have covered two of the most commonly used English text datasets. Our findings on both datasets were overall in agreement with each other. We have highlighted some of the differences, such as deduplication being more effective on OSCAR due to it being more noisy than C4. Further, we have focused on large-scale pre-training datasets. There is a lot of research on the optimal fine-tuning dataset and methodology for LLMs~\cite{sanh2022multitask,longpre2023flan,yong2022bloom+,ouyang2022training,wei2021finetuned,min2021metaicl,wang2022self,zhou2023lima,wang2023far,gupta2023instruction,xu2023wizardlm,muennighoff2023octopack,longpre2023data}. More investigations of resolving data-constraints when fine-tuning LLMs may be of interest for future work.

\paragraph{Other modalities or architectures} Our work focuses on text datasets and uses the GPT transformer architecture~\cite{radford2019language}. Prior work has experimented with many variations to the GPT or transformer architecture~\cite{dehghani2018universal,tay2022scaling,scao2022language}, as well as scaling laws for non-text datasets~\cite{aghajanyan2023scaling}. Overall, variations of the GPT or transformer architecture have proven very robust and generalizable to other domains~\cite{huang2018music,chen2020generative,muennighoff2020vilio,madani2020progen,muennighoff2022sgpt,tay2022scaling,dehghani2023scaling}. Nonetheless, it may be of interest for future work to test the applicability of our findings in this work to different data modalities or model architectures.

\paragraph{Other strategies} There are numerous strategies to solve data constraints not covered in this work that are worth exploring. Like we have shown for Python, future research may consider to what extent augmenting with a natural language (e.g. Chinese) improves performance in another language (e.g. English) and what is the best language to choose~\cite{lin2019choosing,xia2021metaxl}. Similarly, while we have looked at deduplication and perplexity filtering, other filtering strategies, such as popularity-based filters~\cite{allal2023santacoder,zhao2023survey} and toxicity filters~\cite{gehman2020realtoxicityprompts,henderson2022pile,longpre2023pretrainers,prabhumoye2023adding,penedo2023refinedweb} are worth exploring.


\FloatBarrier

\section{Contributions}
\label{sec:contributions}

\textbf{Niklas Muennighoff} led experiments, analysis, writing, and the overall project. He implemented, trained and evaluated all models.

\textbf{Alexander M. Rush} contributed to framing, results analysis, and paper writing.

\textbf{Boaz Barak} contributed to formal and experimental analysis as well as paper writing.

\textbf{Teven Le Scao} provided guidance, led data choices and preprocessing, and contributed to framing and writing.

\textbf{Aleksandra Piktus} created perplexity and deduplication datasets and contributed to writing.

\textbf{Nouamane Tazi} contributed to enabling high-performance training on AMD hardware.

\textbf{Sampo Pyysalo} contributed to enabling high-performance training and early repetition experiments.

\textbf{Thomas Wolf} provided guidance on experimental design and contributed to paper writing.

\textbf{Colin Raffel} provided guidance on experimental design and contributed to paper writing.

\FloatBarrier

\section{Hyperparameters and Setup}
\label{sec:arch}

For all training runs we use 1\% of tokens for linear warm-up of the learning rate to a maximum learning rate of 2e-4 that is decayed to 2e-5 following a cosine schedule. We use a batch size of 256 for models with fewer than 2 billion parameters, 512 for models with 2 - 5 billion parameters and 1024 for models with more than 5 billion parameters. All models are trained in bfloat16 precision using the Adam optimizer~\cite{kingma2014adam} with $eps=1e-8$, $beta1=0.9$. For $beta2$, we found a value of $0.95$ to result in slightly lower final loss and fewer loss spikes than the default value of $0.999$ in implementations such as PyTorch. However, except for models with FLOP budgets of $C=9.3 \times 10^{20}$ and $2.1 \times 10^{21}$, we always use $beta2=0.999$. We use a dropout rate of $0.1$, a weight decay rate of $0.1$ and clip gradients at $1.0$. These hyperparameter choices are largely based on prior work~\cite{hoffmann2022training,touvron2023llama} and performance on test runs. As none of our hyperparameter choices is particularly exotic, we expect our setup to generalize to many other setups. In \autoref{tab:all_models} we list the model architectures we use. They are an extended version of the architectures from \cite{hoffmann2022training}. We calculate model parameters following~\cite{narayanan2021efficient}, which includes embedding parameters:

\begin{equation}
  P = 12lh^2\left(1 + \dfrac{13}{12h}+\dfrac{V+s}{12lh}\right)
  \label{eq:num-params}
\end{equation}

where $P$ is the final parameter count, $l$ are layers, $h$ is the hidden dimension, $V=50257$ the vocabulary size and $s=2048$ the sequence length. We find the parameter counts reported in Chinchilla~\cite{hoffmann2022training} to be significantly different than our calculations, especially at larger scales. Part of this difference comes from their smaller vocabulary size of $V=32000$.  We report both our and the parameter counts of Chinchilla in \autoref{tab:all_models}, but we use our 
parameter estimates everywhere in this work. Further, we have corrected the number of heads of the 3,530 and 4,084 million parameter models from~\cite{hoffmann2022training} to obey the relationship $d\_model=kv\_size \cdot n\_heads$.


To train our models, we have forked the Megatron-DeepSpeed~\cite{rasley2020deepspeed,smith2022using} framework and adapted it for ROCm to enable training on AMD GPUs. We have made our training code publicly available at \url{https://github.com/TurkuNLP/Megatron-DeepSpeed}. Models are trained using data, tensor and pipeline parallelism on up to 256 AMD Instinct MI250X GPUs distributed across up to 64 nodes on the LUMI supercomputer located in Finland. As of June 2023, LUMI is the largest supercomputer in Europe and ranks third worldwide with a performance of around 310 PFLOPs.\footnote{https://www.top500.org/lists/top500/2023/06/} We trained models in parallel using up to 2,200 nodes at a single point in time (equivalent to around 8,800 GPUs or 17,600 GCDs or 86\% of all GPUs on LUMI). We have used a total of around 3 million GPU hours. The cluster is powered 100\% by renewable energy (hydroelectricity) and its waste heat is used for heating the nearby city reducing the city's carbon emissions by up to 20\%. Thanks to the low temperatures in Finland, relatively little cooling for the cluster is required further reducing its impact on the environment. As of June 2023, it ranks as the seventh greenest supercomputer.\footnote{https://www.top500.org/lists/green500/2023/06/}

\clearpage
\footnotesize
\begin{longtable}[h!]
{rr | rrrrr }
\toprule
\multicolumn{2}{c|}{Parameters (millions)} & d\_model & ffw\_size & kv\_size & n\_heads &  n\_layers \\
This work & Chinchilla & & &  &  & \\
\midrule
7 & - & 128 & 512 & 32 & 4 & 3 \\
14 & - & 224 & 896 & 32 & 7 & 4 \\
20 & - & 288 & 1152 & 32 & 7 & 5 \\
38 & - & 448 & 1792 & 32 & 7 & 6 \\
52 & 44 & 512 & 2048 & 64 & 8 & 8 \\
66 & 57 & 576 & 2304 & 64 & 9 & 9 \\
83 & 74 & 640 & 2560 & 64 & 10 & 10 \\
97 & 90 & 640 & 2560 & 64 & 10 & 13 \\
112 & 106 & 640 & 2560 & 64 & 10 & 16 \\
125 & 117 & 768 & 3072 & 64 & 12 & 12 \\
146 & 140 & 768 & 3072 & 64 & 12 & 15 \\
168 & 163 & 768 & 3072 & 64 & 12 & 18 \\
182 & 175 & 896 & 3584 & 64 & 14 & 14 \\
201 & 196 & 896 & 3584 & 64 & 14 & 16 \\
220 & 217 & 896 & 3584 & 64 & 14 & 18 \\
255 & 251 & 1024 & 4096 & 64 & 16 & 16 \\
280 & 278 & 1024 & 4096 & 64 & 16 & 18 \\
305 & 306 & 1024 & 4096 & 64 & 16 & 20 \\
421 & 425 & 1280 & 5120 & 128 & 10 & 18 \\
480 & 489 & 1280 & 5120 & 128 & 10 & 21 \\
502 & 509 & 1408 & 5632 & 128 & 11 & 18 \\
539 & 552 & 1280 & 5120 & 128 & 10 & 24 \\
574 & 587 & 1408 & 5632 & 128 & 11 & 21 \\
619 & 632 & 1536 & 6144 & 128 & 12 & 19 \\
645 & 664 & 1408 & 5632 & 128 & 11 & 24 \\
704 & 724 & 1536 & 6144 & 128 & 12 & 22 \\
789 & 816 & 1536 & 6144 & 128 & 12 & 25 \\
865 & 893 & 1792 & 7168 & 128 & 14 & 20 \\
981 & 1018 & 1792 & 7168 & 128 & 14 & 23 \\
1096 & 1143 & 1792 & 7168 & 128 & 14 & 26 \\
1215 & 1266 & 2048 & 8192 & 128 & 16 & 22 \\
1364 & 1424 & 2176 & 8704 & 128 & 17 & 22 \\
1366 & 1429 & 2048 & 8192 & 128 & 16 & 25 \\
1517 & 1593 & 2048 & 8192 & 128 & 16 & 28 \\
1535 & 1609 & 2176 & 8704 & 128 & 17 & 25 \\
1650 & 1731 & 2304 & 9216 & 128 & 18 & 24 \\
1706 & 1794 & 2176 & 8704 & 128 & 17 & 28 \\
1905 & 2007 & 2304 & 9216 & 128 & 18 & 28 \\
2160 & 2283 & 2304 & 9216 & 128 & 18 & 32 \\
2179 & 2298 & 2560 & 10240 & 128 & 20 & 26 \\
2494 & 2639 & 2560 & 10240 & 128 & 20 & 30 \\
2809 & 2980 & 2560 & 10240 & 128 & 20 & 34 \\
3090 & - & 2688 & 10752 & 128 & 22 & 34 \\
3263 & 3530 & 2688 & 10752 & 128 & 21 & 36 \\
3574 & 3802 & 2816 & 11264 & 128 & 22 & 36 \\
3900 & 4084 & 2944 & 11776 & 128 & 23 & 36 \\
4239 & 4516 & 3072 & 12288 & 128 & 24 & 36 \\
6355 & 6796 & 3584 & 14336 & 128 & 28 & 40 \\
8672 & 9293 & 4096 & 16384 & 128 & 32 & 42 \\
10912 & 11452 & 4352 & 17408 & 128 & 32 & 47 \\
11455 & 12295 & 4608 & 18432 & 128 & 36 & 44 \\
12220 & 12569 & 4608 & 18432 & 128 & 32 & 47 \\
13601 & 13735 & 4864 & 19456 & 128 & 32 & 47 \\
14917 & 14940 & 4992 & 19968 & 128 & 32 & 49 \\
15056 & 16183 & 5120 & 20480 & 128 & 40 & 47 \\
\bottomrule
\caption{\textbf{Model architectures.}
We list the architectures of all models trained as part of this work. Many shown models have been trained multiple times on different amounts of unique data and for varying epochs.}
\label{tab:all_models}
\end{longtable}

\FloatBarrier

\section{Prompts and Samples}
\label{sec:samples}

The following figures illustrate the prompts with samples from each evaluation dataset. Prompts stem from PromptSource~\cite{promptsource} or GPT-3~\cite{brown2020language}. All data comes from the ground truth datasets in this section, and no generations are shown here.

\begin{figure}[!h] { \tt \footnotesize \begin{tabularx}{\linewidth}{r X} \toprule Context $\to$ & Edmond (or Edmund) Halley, FRS (pronounced ; 8 November [O.S. 29 October] 1656 – 25 January 1742 [O.S. 14 January 1741] ) was an English astronomer, geophysicist, mathematician, meteorologist, and physicist who is best known for computing the orbit of Halley's Comet. He was the second Astronomer Royal in Britain, succeeding John Flamsteed.

Question: Edmond Halley was born outside of the United Kingdom. True, False, or Neither?

Answer:\\
 \midrule 
 Correct Answer $\to$ & ~Neither \\ 
 Incorrect Answer $\to$ & ~True \\ 
 Incorrect Answer $\to$ & ~False \\
 \bottomrule \end{tabularx} }
 \caption{Formatted dataset example from ANLI R1 evaluated using accuracy as described in \autoref{sec:eval}.}
 \label{eval:anlir1}
 \end{figure}

\begin{figure}[!h] { \tt \footnotesize \begin{tabularx}{\linewidth}{r X} \toprule Context $\to$ & The 1970 Swedish Open was a combined men's and women's tennis tournament played on outdoor clay courts held in Båstad, Sweden and was part of the Grand Prix circuit of the 1970 Tour. It was the 23rd edition of the tournament and was held from 2 July through 12 July 1970. Dick Crealy and Peaches Bartkowicz won the singles titles.

Question: Dick Crealy and Peaches Bartkowicz beat eachother in the 1970 Swedish Open. True, False, or Neither?

Answer:\\
\midrule 
Correct Answer $\to$ & ~False \\ 
Incorrect Answer $\to$ & ~True \\ 
Incorrect Answer $\to$ & ~Neither \\
\bottomrule \end{tabularx} }
\caption{Formatted dataset example from ANLI R2 evaluated using accuracy as described in \autoref{sec:eval}.}  
\label{eval:anlir2}  
\end{figure}

\begin{figure}[!h] { \tt \footnotesize \begin{tabularx}{\linewidth}{r X} \toprule Context $\to$ & Tokyo - Food group Nestle is seeking to lure Japanese holiday shoppers with a taste for fine snacking with a gold-wrapped Kit Kat chocolate bar. The single finger Kit Kat is wrapped in a thin layer of gold leaf. Only 500 of the bars go on sale from Dec. 29 with a price tag of around 2,016 yen (\$16). The Kit Kat chocolate bar made its debut in Japan in 1973 and since then a variety of flavors -- from green tea to wasabi -- have been produced.

Question: Japanese like kit kat. True, False, or Neither?

Answer:\\
\midrule 
Correct Answer $\to$ & ~True \\ 
Incorrect Answer $\to$ & ~False \\ 
Incorrect Answer $\to$ & ~Neither \\
\bottomrule \end{tabularx} }
\caption{Formatted dataset example from ANLI R3 evaluated using accuracy as described in \autoref{sec:eval}.}  
\label{eval:anlir3}  
\end{figure}
 
\begin{figure}[!h] { \tt \footnotesize \begin{tabularx}{\linewidth}{r X} \toprule Context $\to$ & An astronomer observes that a planet rotates faster after a meteorite impact. Which is the most likely effect of this increase in rotation? \\
\midrule 
Correct Answer $\to$ & ~Planetary days will become shorter. \\ 
Incorrect Answer $\to$ & ~Planetary years will become longer. \\ 
Incorrect Answer $\to$ & ~Planetary gravity will become stronger. \\
\bottomrule \end{tabularx} }
\caption{Formatted dataset example from ARC-Challenge evaluated using accuracy as described in \autoref{sec:eval}.}  
\label{eval:arcc}  
\end{figure}

\begin{figure}[!h] { \tt \footnotesize \begin{tabularx}{\linewidth}{r X} \toprule Context $\to$ & To express the distance between the Milky Way galaxy and other galaxies, the most appropriate unit of measurement is the \\
\midrule 
Correct Answer $\to$ & ~light-year. \\ 
Incorrect Answer $\to$ & ~meter. \\ 
Incorrect Answer $\to$ & ~kilometer. \\
Incorrect Answer $\to$ & ~astronomical unit. \\
\bottomrule \end{tabularx} }
\caption{Formatted dataset example from ARC-Easy evaluated using accuracy as described in \autoref{sec:eval}.}  
\label{eval:arce}  
\end{figure}

\begin{figure}[!h] { \tt \footnotesize \begin{tabularx}{\linewidth}{r X} \toprule Context $\to$ & Radio wave -- Radio waves are a type of electromagnetic radiation with wavelengths in the electromagnetic spectrum longer than infrared light. Radio waves have frequencies as high as 300 gigahertz (GHz) to as low as 30 hertz (Hz). At 300 GHz, the corresponding wavelength is 1 mm, and at 30 Hz is 10,000 km. Like all other electromagnetic waves, radio waves travel at the speed of light. They are generated by electric charges undergoing acceleration, such as time varying electric currents. Naturally occurring radio waves are emitted by lightning and astronomical objects.

Question: do radio waves travel at the speed of light?

Answer:\\
\midrule 
Correct Answer $\to$ & ~yes \\ 
Incorrect Answer $\to$ & ~no \\
\bottomrule \end{tabularx} }
\caption{Formatted dataset example from BoolQ evaluated using accuracy as described in \autoref{sec:eval}.}  
\label{eval:boolq}  
\end{figure}

\begin{figure}[!h] { \tt \footnotesize \begin{tabularx}{\linewidth}{r X} \toprule Context $\to$ & A: Okay. So Frank, what, uh, type of, uh, budget do you or your family have? B: Well, uh I don't know that we really have a budget.

Question: he and his family really have a budget. True, False or Neither?

Answer:\\
\midrule 
Correct Answer $\to$ & ~False \\ 
Incorrect Answer $\to$ & ~True \\
Incorrect Answer $\to$ & ~Neither \\
\bottomrule \end{tabularx} }
\caption{Formatted dataset example from CB evaluated using accuracy as described in \autoref{sec:eval}.}  
\label{eval:cb}  
\end{figure}

\begin{figure}[!h] { \tt \footnotesize \begin{tabularx}{\linewidth}{r X} \toprule Context $\to$ & The computer was expensive to fix therefore \\
\midrule 
Correct Answer $\to$ & ~I bought a new one. \\ 
Incorrect Answer $\to$ & ~I got it repaired. \\
\bottomrule \end{tabularx} }
\caption{Formatted dataset example from COPA evaluated using accuracy as described in \autoref{sec:eval}.}  
\label{eval:copa}  
\end{figure}

\begin{figure}[!h] { \tt \footnotesize \begin{tabularx}{\linewidth}{r X} \toprule Context $\to$ & Canoeing: Two women in a child are shown in a canoe while a man pulls the canoe while standing in the water, with other individuals visible in the background. The child and a different man\\
\midrule 
Correct Answer $\to$ & ~sit in a canoe while the man paddles. \\ 
Incorrect Answer $\to$ & ~are then shown paddling down a river in a boat while a woman talks. \\
Incorrect Answer $\to$ & ~are driving the canoe, they go down the river flowing side to side. \\
Incorrect Answer $\to$ & ~walking go down the rapids, while the man in his helicopter almost falls and goes out of canoehood. \\
\bottomrule \end{tabularx} }
\caption{Formatted dataset example from HellaSwag evaluated using accuracy as described in \autoref{sec:eval}.}  
\label{eval:hellaswag}  
\end{figure}

\begin{figure}[!h] { \tt \footnotesize \begin{tabularx}{\linewidth}{r X} \toprule Context $\to$ & Question: How to sleep in proper posture?

Answer:\\
\midrule 
Correct Answer $\to$ & ~Sleep straight with a pillow under your head. \\ 
Incorrect Answer $\to$ & ~Sleep straight with a pillow over your head. \\
\bottomrule \end{tabularx} }
\caption{Formatted dataset example from PiQA evaluated using accuracy as described in \autoref{sec:eval}.}  
\label{eval:piqa}  
\end{figure}

\begin{figure}[!h] { \tt \footnotesize \begin{tabularx}{\linewidth}{r X} \toprule Context $\to$ & As spacecraft commander for Apollo XI, the first manned lunar landing mission, Armstrong was the first man to walk on the Moon. "That's one small step for a man, one giant leap for mankind." With these historic words, man's dream of the ages was fulfilled. 

Question: Neil Armstrong was the first man who landed on the Moon. True or False?

Answer:\\
\midrule 
Correct Answer $\to$ & ~True. \\ 
Incorrect Answer $\to$ & ~False. \\
\bottomrule \end{tabularx} }
\caption{Formatted dataset example from RTE evaluated using accuracy as described in \autoref{sec:eval}.}  
\label{eval:rte}  
\end{figure}

\begin{figure}[!h] { \tt \footnotesize \begin{tabularx}{\linewidth}{r X} \toprule Context $\to$ & The electromagnetic spectrum encompasses a very wide range of wavelengths and frequencies. Visible light is only a very small portion of the spectrum with wavelengths from 400-700 nm.

Question: With wavelengths from 400-700 nm, what kind of light represents only a very small portion of the spectrum?

Answer:\\
\midrule 
Correct Answer $\to$ & ~visible light.\\
Incorrect Answer $\to$ & ~ultraviolet light. \\ 
Incorrect Answer $\to$ & ~invisible light. \\
Incorrect Answer $\to$ & ~sunlight. \\
\bottomrule \end{tabularx} }
\caption{Formatted dataset example from SciQ evaluated using accuracy as described in \autoref{sec:eval}.}  
\label{eval:sciq}  
\end{figure}

\begin{figure}[!h] { \tt \footnotesize \begin{tabularx}{\linewidth}{r X} \toprule Context $\to$ & Bob went to the gas station to fill up his car. His tank was completely empty and so was his wallet. The cashier offered to pay for his gas if he came back later to pay. Bob felt grateful as he drove home.

Answer:\\
\midrule 
Correct Answer $\to$ & ~Bob believed that there were good people in the world.\\
Incorrect Answer $\to$ & ~Bob contemplated how unfriendly the world was.\\
\bottomrule \end{tabularx} }
\caption{Formatted dataset example from StoryCloze evaluated using accuracy as described in \autoref{sec:eval}.}
\label{eval:storycloze}  
\end{figure}

\begin{figure}[!h] { \tt \footnotesize \begin{tabularx}{\linewidth}{r X} \toprule Correct Context $\to$ & Johnny likes fruits more than vegetables in his new keto diet because the fruits:\\
Incorrect Context $\to$ & Johnny likes fruits more than vegetables in his new keto diet because the
vegetables:\\
\midrule 
Target Completion $\to$ & ~are saccharine.\\
\bottomrule \end{tabularx} }
\caption{Formatted dataset example from WinoGrande evaluated using accuracy as described in \autoref{sec:eval}.}  
\label{eval:winogrande}  
\end{figure}

\begin{figure}[!h] { \tt \footnotesize \begin{tabularx}{\linewidth}{r X} \toprule Context $\to$ & Given the following data about a restaurant:

name : The Wrestlers

eatType : pub

food : Japanese

priceRange : cheap

area : riverside

near : Raja Indian Cuisine\\
& \\
& Generate some text about this restaurant.\\
\midrule 
Target $\to$ & ~The Wrestlers offers Japanese food and pub with cheap price near Raja Indian Cuisine in riverside.\\
\bottomrule \end{tabularx} }
\caption{Formatted dataset example from E2E NLG evaluated using ROUGE as described in \autoref{sec:eval}.}
\label{eval:e2enlg}  
\end{figure}

\begin{figure}[!h] { \tt \footnotesize \begin{tabularx}{\linewidth}{r X} \toprule Context $\to$ & Article: The artificial intelligence system - LipNet - watches video of a person speaking and matches the text to the movement of their mouths with 93\% accuracy, the researchers said.

Automating the process could help millions, they suggested.

But experts said the system needed to be tested in real-life situations.

Lip-reading is a notoriously tricky business with professionals only able to decipher what someone is saying up to 60\% of the time.

"Machine lip-readers have enormous potential, with applications in improved hearing aids, silent dictation in public spaces, covert conversations, speech recognition in noisy environments, biometric identification and silent-movie processing," wrote the researchers.

They said that the AI system was provided with whole sentences so that it could teach itself which letter corresponded to which lip movement.

To train the AI, the team - from Oxford University's AI lab - fed it nearly 29,000 videos, labelled with the correct text. Each video was three seconds long and followed a similar grammatical pattern.

While human testers given similar videos had an error rate of 47.7\%, the AI had one of just 6.6\%.

The fact that the AI learned from specialist training videos led some on Twitter to criticise the research.

Writing in OpenReview, Neil Lawrence pointed out that the videos had "limited vocabulary and a single syntax grammar".

"While it's promising to perform well on this data, it's not really groundbreaking. While the model may be able to read my lips better than a human, it can only do so when I say a meaningless list of words from a highly constrained vocabulary in a specific order," he writes.

The project was partially funded by Google's artificial intelligence firm DeepMind. \\
& \\
& Summary:\\
\midrule 
Target $\to$ & ~Scientists at Oxford University have developed a machine that can lip-read better than humans.\\
\bottomrule \end{tabularx} }
\caption{Formatted dataset example from XSUM evaluated using ROUGE as described in \autoref{sec:eval}.}  
\label{eval:xsum}  
\end{figure}

\begin{figure}[!h] { \tt \footnotesize \begin{tabularx}{\linewidth}{r X} \toprule Context $\to$ & I will verbalize an abstract representation of a sentence in natural language. To do so, I will first show the representation and then the natural language. The text needs to include all of the information in the representation.

Brandon\_Carter | almaMater | University\_of\_Cambridge, University\_of\_Cambridge | chancellor | David\_Sainsbury,\_Baron\_Sainsbury\_of\_Turville, Brandon\_Carter | birthPlace | England, University\_of\_Cambridge | viceChancellor | Leszek\_Borysiewicz\\
\midrule 
Target $\to$ & ~The University of Cambridge is the alma mater of Brandon Carter, who was born in England. David Sainsbury, also known as the Baron Sainsbury of Turville, and Leszek Borysiewicz are respectively the chancellor and vice chancellor of the University of Cambridge.\\
\bottomrule \end{tabularx} }
\caption{Formatted dataset example from WebNLG evaluated using ROUGE as described in \autoref{sec:eval}.}  
\label{eval:webnlg}  
\end{figure}

\begin{figure}[!h] { \tt \footnotesize \begin{tabularx}{\linewidth}{r X} \toprule Context $\to$ & Attributes are placed within the tag itself, making additional alterations to the \"element content\" between the start and end tag. They never stand alone. They are written in the format name=\"value\", where name is the name of the attribute (for instance \"color\"), and value describes this specific instance (for instance \"red\"). You've actually seen attributes before, if you followed the tutorial in the basic HTML section. <img> tags use the src attribute, anchors use the name attribute, and links use the href attribute. See how those all follow the \_\_\_=\"\_\_\_\" format? Making a table, or chart, requires several different tags. Play with these tags, or learn about HTML tables in more detail.  Start with table tags around the entire table:<table></table>  Row tags around the contents of each row: <tr>  Column headers in the first row: <th>  Cells in subsequent rows: <td>  Here's an example of how it all fits together:<table><tr><th>Column 1: Month</th><th>Column 2: Money Saved</th></tr><tr><td>January</td><td>\$100</td></tr></table> You've already learned the <head> tag, which shows up at the start of each document. Besides the <title> tag, it can include the following types of tags:   Meta tags, which are used to provide metadata about a web page. This data can be used by search engines when the robot scours the internet to locate and list websites. To make your website more visible on search engines, use one or more <meta> start tags (no end tags necessary), each with exactly one name attribute and one content attribute, for example: <meta name=\"description\" content=\"write a description here\">; or <meta name=\"keywords\" content=\"write a list of keywords, each separated by a comma\"> <link> tags are used to associate other files with the page. This is mainly used to link to CSS stylesheets, which are made using a different type of coding to alter your HTML page by adding color, aligning your text, and many other things. <script> tags are used to link the page to JavaScript files, which can cause the page to change as... \\
& \\
& TL;DR in English:\\
\midrule 
Target $\to$ & ~Learn about attributes. Experiment with HTML tables. Learn the miscellaneous head tags. Play around with HTML found on websites. Learn more advanced web design from comprehensive guides.\\
\bottomrule \end{tabularx} }
\caption{Formatted dataset example from WikiLingua evaluated using ROUGE as described in \autoref{sec:eval}.}  
\label{eval:wikilingua}  
\end{figure}


\begin{figure}[ht] { \tt \footnotesize \begin{tabularx}{\linewidth}{r X} \toprule Context $\to$ & John travelled to the kitchen. Sandra moved to the kitchen. Daniel went to the kitchen. John journeyed to the hallway. Mary journeyed to the bedroom. Mary journeyed to the kitchen. Mary travelled to the bedroom. Sandra travelled to the bedroom. John went to the office. John went back to the kitchen. Where is Mary?\\
\midrule 
Target $\to$ & ~bedroom\\
\bottomrule \end{tabularx} }
\caption{Formatted dataset example from bAbI evaluated using exact match as described in \autoref{sec:eval}.}  
\label{eval:babi}  
\end{figure}


\clearpage

\section{Other Experiments}
\label{sec:other}
\paragraph{UL2} We experimented with the UL2 objective~\cite{tay2022unifying,tay2022transcending} for a causal model but did not find it to outperform regular causal language modeling on our evaluation tasks. This may stem from UL2 being better suited as an Encoder-Decoder model or from mistakes in our UL2 implementation.

\paragraph{The Pile} We have also trained several models on The Pile~\cite{gao2020pile} and found similar trends as for OSCAR and C4. We make these models publicly available.

\section{Release of Artifacts}
\label{sec:release}


We open-source all of our models and code under Apache 2.0 licenses. Our filtered datasets are released with the same licenses as the datasets they stem from. All material can be found at: \url{https://github.com/huggingface/datablations}.

\section{Version Control}
\label{sec:vc}

\textbf{V4 → V5:}
\begin{itemize}
    \item Small writing improvements
    \item Fixed $R^2$ scores in \autoref{tab:fits}
\end{itemize}

\textbf{V3 → V4:}
\begin{itemize}
    \item Added comparison of different fits in terms of loss and $R^2$ in \autoref{tab:fits}
    \item Small writing improvements
\end{itemize}

\textbf{V2 → V3:}
\begin{itemize}
    \item Added loss curves of complementary strategies in \autoref{sec:losscomp}
    \item Fixed OSCAR validation plot in \autoref{sec:fixoscar}
    \item Clarified the usage of smoothing in training and validation plots in \autoref{sec:eval}
    \item Added more references
\end{itemize}

\textbf{V1 → V2:}
\begin{itemize}
    \item Added experiments decaying alpha and beta to allow excess epochs or parameters to hurt in \autoref{sec:toomany}  
    \item Added Galactica case study in \autoref{sec:galactica}    
    \item Added more details on the calculation of $U_N$ given $U_D$ in \autoref{sec:scalinglaws}
    \item Added hyperparameter sensitivity limitation in \autoref{sec:limits}
    \item Added more detail on how score normalization is done in \autoref{sec:eval}
    \item Mentioned modification of number of heads in \autoref{sec:arch}    
\end{itemize}

\section{Broader Impacts}
\label{sec:broad}

Large language models carry potential risks such as outputting offensive language, propagating social biases, and leaking private information \cite{weidinger2021ethical,bender2021dangers}. By publicly releasing all of our models and providing new insights to improve the scaling of LLMs we may contribute to the further proliferation of these harms. However, we note that there are already much larger and more capable models freely available~\cite{black2021gpt,black2022gpt,scao2022bloom,biderman2023pythia} that can be used in such harmful ways. Thus, we consider the open-source release of our models and research to significantly outweigh its downsides.

\end{document}